\documentclass[10pt,twocolumn,letterpaper]{article}

\usepackage[pagenumbers]{cvpr}

\usepackage[table]{xcolor}

\usepackage{booktabs}
\usepackage{multirow}
\usepackage{array}
\usepackage{capt-of}

\newcommand{\plip}{\mathcal{P}_{\mathrm{lip}}}
\newcommand{\mdyn}{M_{\mathrm{dyn}}}

\usepackage{dsfont}
\definecolor{cvprblue}{rgb}{0.21,0.49,0.74}
\usepackage[pagebackref,breaklinks,colorlinks,allcolors=cvprblue]{hyperref}

\def\confName{CVPR}
\def\confYear{2027}

\title{When Stillness Looks Fake: Mouth Dynamics Bias in Deepfake Detection}

\author{
Catello Cascone$^{1,2}$ \quad Lucia Cascone$^{1}$ \quad Carlo Sorrentino$^{3}$ \quad Enzo Tartaglione$^{4}$\\
$^{1}$Department of Computer Science, University of Salerno, Italy\\
$^{2}$IMT School for Advanced Studies Lucca, Italy\\
$^{3}$Department of Management \& Innovation Systems, University of Salerno, Italy\\
$^{4}$LTCI, T\'el\'ecom Paris, Institut Polytechnique de Paris, France\\
{\tt\small \{cacascone, lcascone, carsorrentino\}@unisa.it, enzo.tartaglione@telecom-paris.fr}
}

\begin{document}
\twocolumn[{%
\renewcommand\twocolumn[1][]{#1}%
\maketitle
\centering
\captionsetup{type=figure} 
\includegraphics[width=\textwidth]{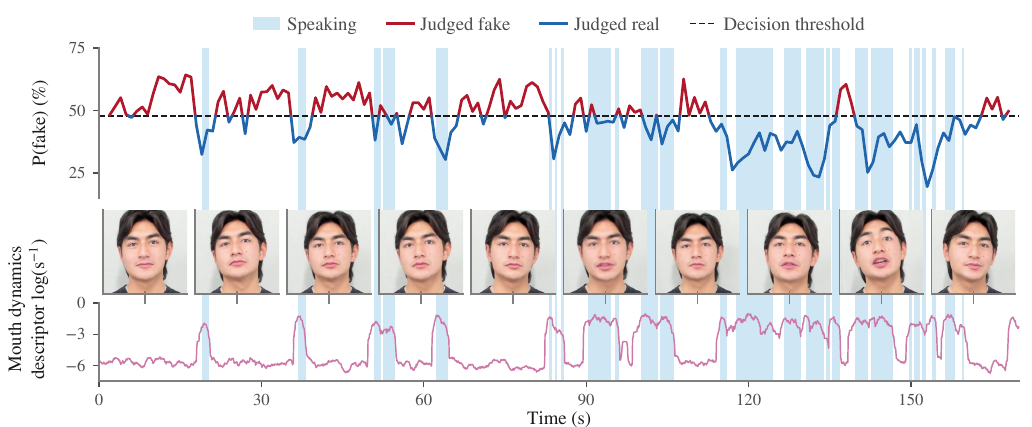}
\caption{\textbf{The same real person looks fake when they stop talking.} We run DFD-FCG~\cite{han2025dfdfcg}, a state-of-the-art detector (95.0\% AUC on Celeb-DF v2~\cite{li2020celebdf}), on a real video from Seamless Interaction~\cite{seamless2025}. Top: $P(\text{fake})$ over time against a threshold set at 5\% false positives on FaceForensics++~\cite{rossler2019faceforensics} real videos, so every point above it (red) is a false positive. Bottom: how much the mouth moves (log scale), with sampled frames. False positives cluster in silent stretches, where the mouth barely moves, and largely vanish during speech (blue bands).} 

\label{fig:teaser}
\vspace{1.2em}
}]

\begin{abstract}

Deepfake detectors perform well on standard benchmarks, but accuracy alone reveals little about the cues they rely on.
We identify a mouth dynamics bias: in real videos, the same person receives higher fake scores when their mouth moves less.
This bias affects nearly all state-of-the-art detectors we evaluate and is strongest in temporal models. Controlled lip animations show that this dependence can be exploited: increasing mouth motion of manipulated videos can halve their detection rate. 
We trace a plausible source of the bias to FaceForensics++, where reduced mouth motion correlates with manipulation. Because its test split shares this correlation, standard benchmark evaluation can overlook the bias. Finally, a mouth-dynamics-aware score correction substantially reduces the bias for unseen individuals from the calibration dataset but does not reliably transfer across datasets. The code will be released upon acceptance of the article.

\end{abstract}

\section{Introduction}
\label{sec:intro}

Deepfake detectors achieve strong performance on established benchmarks~\cite{yan2024df40}, yet high accuracy does not imply that they rely on evidence of facial manipulation. They may instead exploit spurious correlations in training data, learning features that predict manipulation labels without reflecting manipulation itself~\cite{hermann2024shortcut}.

Prior work has leveraged mouth dynamics for deepfake detection, particularly through temporal irregularities in lip movements and audio-visual inconsistencies~\cite{haliassos2021lips,haliassos2022realforensics,peng2026havic}. Related studies have examined shortcuts arising from audio silence~\cite{smeu2025shortcuts} and the detection of manipulated listeners~\cite{liu2026listening}. Closer to our setting, attribute-based analyses have reported detection disparities across facial attributes, including mouth closure~\cite{xu2024madd}. We study instead how natural variations in mouth \emph{motion}, rather than static appearance, affect detector scores within the same authentic person. We also ask how training data may encourage this dependence and why standard benchmarks may not reveal it.

We identify a \emph{mouth dynamics bias}: detector scores depend on how much the mouth moves, even when this is unrelated to manipulation. We observe this bias as a \emph{non-speaking penalty}: detectors tend to assign higher fake scores to the same authentic person during non-speaking periods, when mouth dynamics are typically lower, sometimes misclassifying authentic videos as fake (Real~$\rightarrow$~Fake; \cref{fig:teaser}). This dependence also raises an adversarial concern: increasing mouth motion in manipulated videos may shift predictions toward real (Fake~$\rightarrow$~Real). We trace a plausible origin of the bias to FaceForensics++ (FF++)~\cite{rossler2019faceforensics}, a widely used training benchmark in which several manipulation methods attenuate mouth dynamics while authentic low-motion behavior is poorly represented.

Our main contributions are summarized as follows:

\begin{itemize}
\item \textbf{Mouth dynamics bias.} We uncover a widespread tendency of deepfake detectors to assign higher fake scores to real videos with limited mouth movement (\cref{fig:teaser}), particularly in temporal models. Within the same person, detector scores decrease as mouth dynamics increase, even within each conversational state (\cref{sec:bias}).

\item \textbf{Vulnerability to mouth-motion manipulation.} Increasing lip motion in manipulated videos substantially reduces their detection. In contrast, lips held still but open do not: scores follow mouth motion, not opening, exposing a potential Fake~$\rightarrow$~Real evasion (\cref{sec:changing-mouth}).

\item \textbf{Training-data origin and underspecification.} Controlled retraining links the non-speaking penalty to FF++ training composition. We further show that classifiers sharing the same representations can exhibit different penalties despite nearly identical FF++ performance, and FF++-only selection procedures fail to reliably identify the less biased classifier (\cref{sec:origin}).

\item \textbf{Diagnostic score correction.} Using within-person mouth dynamics, we substantially reduce the non-speaking penalty on unseen participants from the calibration dataset without retraining detector representations (\cref{fig:mdyn-geometry}). Limited cross-dataset transfer, residual false-positive gaps, and reduced detection of certain manipulations reveal the limits of score-level correction (\cref{sec:score-edit}).

\end{itemize}

\section{Related Work}

\paragraph{Shortcuts and dataset bias.}
Deep neural networks may exploit spurious correlations that predict training labels but fail to generalize under distribution shifts~\cite{geirhos2020shortcut,hermann2024shortcut}. In deepfake detection, detectors may learn who appears in a video rather than whether the face is manipulated~\cite{dong2023identity}, exhibit different error rates across demographic groups~\cite{trinh2021fairness,lamsaf2025bias}, and overfit to artifacts associated with particular manipulation methods~\cite{yan2023ucf,shiohara2022sbi}. Audio can provide another predictive cue: a brief silence at the beginning of fake recordings may reveal their label~\cite{smeu2025shortcuts}. More closely related to facial behavior, attribute-based studies have examined detection disparities across facial attributes, including mouth closure~\cite{xu2024madd}. Unlike attribute-based comparisons, we examine how natural variations in mouth dynamics affect detector scores within the same authentic person, independently of static mouth appearance.

\paragraph{Mouth and audio-visual cues.}
Mouth motion is already an important source of evidence in deepfake detection: existing methods analyze visual speech representations and talking-face dynamics~\cite{haliassos2021lips,haliassos2022realforensics}, temporal facial-landmark trajectories and phase-based lip motion~\cite{sun2021lrnet,prashnani2025phaseforensics,abate2027labial}, or inconsistencies between speech and visible articulation~\cite{agarwal2020phoneme,feng2023avad}, with much of this work developed and evaluated on videos of people speaking. Recent work has also investigated detecting manipulated faces of people who are listening~\cite{liu2026listening}, extending attention to facial behavior during periods without speech. We investigate the complementary false-positive scenario: whether detectors assign higher manipulation scores to authentic individuals during non-speaking periods because their mouths naturally move less.

\paragraph{Underspecification and bias correction.}
Models with similar benchmark performance may behave differently outside the evaluation distribution, as highlighted by work on underspecification~\cite{damour2022underspecification} and predictive multiplicity~\cite{marx2020multiplicity}. The final classifier alone can make the difference: retraining only the last layer on group-balanced data can reduce reliance on spurious correlations while leaving the learned representations unchanged~\cite{kirichenko2023dfr}. In our setting, we show that classifiers sharing the same representations can achieve nearly identical FF++ performance while exhibiting substantially different non-speaking penalties, and that the FF++-only selection procedures we evaluate fail to reliably distinguish them. Existing approaches to addressing spurious correlations include worst-group optimization~\cite{sagawa2020groupdro}, upweighting misclassified examples~\cite{liu2021jtt}, automatic feature reweighting~\cite{qiu2023afr}, feature-distribution alignment~\cite{sun2016coral}, linear concept erasure~\cite{belrose2023leace}, and fairness-aware learning~\cite{donini2018erm,zafar2017fairness,lin2024preserving}. Rather than retraining detector representations, we use within-person variation in authentic mouth dynamics to correct detector scores without requiring speaking labels for fitting, and examine the resulting trade-offs in false-positive disparities, cross-dataset transfer, and detection of manipulated videos.
\definecolor{penalty}{RGB}{230,120,40}

\section{Experimental Setup}
\label{sec:setup}

\subsection{Datasets}
\label{sec:data}

\paragraph{Conversational datasets.}
We use three corpora of real conversations in which each participant alternates between speaking and not, and which differ in recording conditions and speech annotation: Seamless Interaction~\cite{seamless2025} (dyadic conversations in controlled settings: 210 single-participant recordings of 99 participants from the naturalistic test split; speech from each participant's own audio track), RealTalk~\cite{geng2023realtalk} (470 two-person YouTube conversations, 937 participants; TalkNet active-speaker detection~\cite{tao2021talknet}), and AMI~\cite{carletta2005ami} (121 four-person meetings filmed with close-up cameras, 124 participants; manual transcripts). Given that TalkNet also uses lip motion, Seamless and AMI, labeled from audio and transcripts, serve as controls for this choice. 

From each recording, we sample non-overlapping 4\,s windows: a window is \emph{speaking} when the visible participant speaks for at least 85\% of it, and \emph{non-speaking} when the participant is silent throughout it and for 1\,s on either side. These rules were fixed before any detector was scored and applied unchanged to all corpora. Non-speaking thus means sustained silence rather than brief pauses, not stillness: participants may smile, laugh, nod, or move their lips. We evaluate only participants observed in both states; window counts, the split of RealTalk into development and held-out participants, and checks of the labeling rules are detailed in the supplementary material.

\paragraph{Deepfake datasets.}
FF++~\cite{rossler2019faceforensics} (c23) contains 1{,}000 real YouTube videos and 4{,}000 fakes generated by DeepFakes (DF), Face2Face (F2F), FaceSwap (FS), and NeuralTextures (NT). For cross-dataset evaluation, we use the Celeb-DF++ test split~\cite{li2025celebdfpp}, with 178 real and 5{,}240 fake videos from 22 face-swapping, face-reenactment, and talking-face generators, in \cref{sec:changing-mouth,sec:score-edit}. It includes the Celeb-DF v2 test split~\cite{li2020celebdf} (the same 178 real and 340 face-swapped videos), which we also report on its own as the standard benchmark.
Subsets and preprocessing are detailed in the supplementary material.

\subsection{Deepfake Detectors}
\label{sec:detectors}

\begin{table}[t]
\centering
\footnotesize
\setlength{\tabcolsep}{4pt}
\renewcommand{\arraystretch}{1.05}
\newcommand{\pv}[1]{\cellcolor{penalty!\fpeval{max(15,min(55,(#1-50)*1.4))}}#1}
\newcommand{\tfam}[2]{\addlinespace[3pt]\multicolumn{6}{@{}l@{}}{\textit{#1}\ \ {\footnotesize\color{black!55}#2}}\\}
\newcommand{\xd}{$^{\star}$}
\caption{\textbf{The non-speaking penalty is widespread across released detectors.}
$A_{\mathrm{within}}$ (\cref{eq:awithin}) on Seamless (Seam.), RealTalk (RT) and AMI, with detectors grouped into temporal (T) and frame-based (F) families.
Shaded cells mark a statistically supported penalty, darker for larger values.
Detectors marked $^{\star}$ use training data beyond FF++.}
\label{tab:detectors}
\begin{tabular}{@{}lllccc}
\toprule
 & & & \multicolumn{3}{c}{$A_{\mathrm{within}}$ (\%)}\\
\cmidrule(l){4-6}
Detector & Venue & Pretrain. & Seam. & RT & AMI\\
\midrule
\tfam{T-mouth}{mouth region, about 1\,s}
RealForensics~\cite{haliassos2022realforensics}\xd & CVPR'22 & LRW & \pv{87.9} & \pv{86.2} & \pv{77.2}\\
LipForensics~\cite{haliassos2021lips} & CVPR'21 & LRW & \pv{69.5} & \pv{67.5} & \pv{70.2}\\
\tfam{T-3D}{whole face, 3D convolutions}
I3D-DF40-FR~\cite{yan2024df40}\xd & NeurIPS'24 & K400 & \pv{88.4} & \pv{83.4} & \pv{75.1}\\
I3D-DF40-FS~\cite{yan2024df40}\xd & NeurIPS'24 & K400 & \pv{83.9} & \pv{82.7} & \pv{68.9}\\
I3D-DF40-FSFR~\cite{yan2024df40}\xd & NeurIPS'24 & K400 & \pv{82.5} & \pv{77.4} & \pv{68.3}\\
FTCN~\cite{zheng2021ftcn} & ICCV'21 & -- & \pv{58.8} & \pv{52.3} & \pv{60.4}\\
\tfam{T-transformer}{whole face, spatio-temporal attention}
TALL~\cite{xu2023tall} & ICCV'23 & IN-22k & \pv{75.6} & \pv{77.8} & \pv{71.8}\\
FakeSTormer~\cite{nguyen2025fakestormer}\xd & ICCV'25 & IN-1k & \pv{66.4} & \pv{63.2} & 45.4\\
TFCU~\cite{guo2025tfcu} & CVPR'25 & BEiT v2 & 52.9 & \pv{58.4} & \pv{58.0}\\
\tfam{T-foundation}{frozen backbone, temporal aggregator}
DFD-FCG~\cite{han2025dfdfcg} & CVPR'25 & CLIP & \pv{74.6} & \pv{76.3} & \pv{62.5}\\
\tfam{F-foundation}{frozen backbone, per-frame head}
DFD-HR~\cite{sun2026dfdhr} & CVPR'26 & CLIP & \pv{63.6} & \pv{66.9} & \pv{55.2}\\
VLAForge~\cite{zhu2026vlaforge}\xd & CVPR'26 & CLIP & 53.8 & \pv{60.0} & \pv{55.8}\\
GenD-CLIP~\cite{yermakov2026gend} & WACV'26 & CLIP & \pv{58.1} & \pv{65.0} & \pv{53.5}\\
GenD-PE~\cite{yermakov2026gend} & WACV'26 & PE & \pv{58.2} & \pv{63.0} & \pv{61.5}\\
GenD-DINOv3~\cite{yermakov2026gend} & WACV'26 & DINOv3 & \pv{57.0} & \pv{65.6} & 49.9\\
\tfam{F-CNN}{CNN trained end to end, per frame}
SBI~\cite{shiohara2022sbi}\xd & CVPR'22 & IN-1k & \pv{64.7} & \pv{66.3} & \pv{60.9}\\
\bottomrule
\end{tabular}
\end{table}

We evaluate 16 publicly released deepfake detectors using their pretrained checkpoints, without retraining. We select detectors that perform well on standard cross-dataset benchmarks such as Celeb-DF v2 and that span architectures, pretraining strategies, and manipulation training data, to test whether the non-speaking penalty is specific to particular approaches or extends across detector families. 

All detectors receive the same 4\,s windows and apply their own released face preprocessing and frame sampling within them. On full Celeb-DF v2 test videos, no checkpoint falls more than 1.2 points below its reported video-level AUC (checkpoints and preprocessing in supp. mat.). \Cref{tab:detectors} groups the detectors by temporal processing: we call a detector temporal when it processes several frames jointly, whatever its architecture, and frame-based when it processes each frame alone and averages the decisions. All use FF++ real videos during training, while their training fakes and pretraining differ.

\subsection{Mouth Dynamics Descriptor}
\label{sec:measure}

To characterize mouth movement independently of speech annotations, we measure temporal variations in lip aperture, following geometric approaches to visual speech analysis~\cite{sodoyer2009}. We extract facial landmarks using MediaPipe Face Mesh~\cite{kartynnik2019facemesh} independently in each frame and align each clip to a common scale using a single similarity transform, fixing the interocular distance at 64 pixels. At each frame $t$, we compute the area $A_t$ enclosed by the 20 inner-lip landmarks and normalize it as $a_t=A_t/64^2$.

We define the \emph{mouth dynamics score} as:
\begin{equation}
\mdyn =
\log\left(
\frac{\mathrm{fps}}{T-1}
\sum_{t=1}^{T-1}|a_{t+1}-a_t|
\right),
\label{eq:mdyn}
\end{equation}
where $T$ is the number of frames and $\mathrm{fps}$ is the frame rate (alignment, missing frames, and frame-rate effects in supp. mat.).
Higher scores indicate greater temporal variation in mouth opening.

\subsection{Evaluation Metrics}
\label{sec:metric}

\paragraph{Non-speaking penalty.}
We define the \emph{non-speaking penalty} as the tendency of a detector to assign higher fake scores to non-speaking than to speaking windows of the same participant. Let $\mathcal{L}_p$ and $\mathcal{T}_p$ denote the non-speaking and speaking windows of participant $p$, and $\mathcal{P}$ the set of participants with windows in both states. We measure the penalty with the participant-balanced within-participant AUC
\begin{equation}
A_{\mathrm{within}} =
\frac{100}{|\mathcal{P}|}
\sum_{p\in\mathcal{P}}
\frac{1}{|\mathcal{L}_p|\,|\mathcal{T}_p|}
\sum_{i\in\mathcal{L}_p}
\sum_{j\in\mathcal{T}_p} h_{ij},
\label{eq:awithin}
\end{equation}
where $h_{ij}=\mathds{1}(s_i>s_j)+\frac{1}{2}\mathds{1}(s_i=s_j)$ and $s_i$ is the fake score of window $i$, oriented so that higher means more likely fake for every detector. Each participant contributes equally, regardless of the number of windows; the supplementary material reports the pooled-pair alternative and the effect of editing cuts. A value of 50 corresponds to chance-level ranking, and we consider the penalty statistically supported when the lower bound of the 95\% confidence interval of $A_{\mathrm{within}}$ exceeds 50. Confidence intervals come from a bootstrap over the connected components of the person--session graph, which accounts for dependencies among windows of the same participant and among participants recorded in the same conversation.

\paragraph{Size of the penalty.}
Since $A_{\mathrm{within}}$, a rank statistic, cannot be decomposed into contributions, we also measure the \emph{score gap} $\Delta$ in $y_{i}=\alpha_{p(i)}+\Delta L_i+\varepsilon_i$, where $y_i$ is the normal score of the rank of the detector score within each detector and corpus, $\alpha_p$ a participant fixed effect and $L_i$ the non-speaking indicator. Adding $\delta\,x_i$, with $x_i$ the normal score of $\mdyn$, gives the adjusted gap $\Delta_{\mathrm{adj}}$ and the share $1-\Delta_{\mathrm{adj}}/\Delta$ that the covariate accounts for linearly.

\paragraph{Decisions.}
For threshold-based detection analyses, we use \emph{FF95}, the detector-specific threshold yielding a 5\% false-positive rate on the video-level scores of the 140 real FF++ test videos, unless otherwise specified; $\Delta$FPR is the difference in false-positive rate at FF95 between non-speaking and speaking real windows. Since FF95 is transferred from FF++, we also recalibrate it on each corpus: non-speaking windows remain flagged more often in all 16 detectors on RealTalk (7.4\% against 2.8\%, medians; results provided in the supp. mat.).

\section{Mouth Dynamics Bias in Deepfake Detectors}
\label{sec:bias}

A real video does not become less authentic when the person stops speaking. Yet, if deepfake detectors rely on conversational behavior rather than manipulation evidence, they may systematically assign higher fake scores to real participants when they are not speaking. First, we establish how widespread this bias is across detector architectures (\cref{sec:widespread}), then we relate it to mouth dynamics (\cref{sec:follows}), and finally, we edit the mouth directly to test whether the score follows it (\cref{sec:changing-mouth}).

\subsection{The Non-Speaking Penalty Across Detectors}
\label{sec:widespread}

\paragraph{The penalty is widespread.}
\Cref{tab:detectors} reports $A_{\mathrm{within}}$ across 16 deepfake detectors; 95\% confidence intervals are given in the supplementary material. On RealTalk, all detectors exhibit a statistically supported tendency to assign higher fake scores to real videos when the person is not speaking. The effect persists across datasets with different recording conditions and speech annotations, with $A_{\mathrm{within}}$ significantly above 50 for 14 of 16 detectors on Seamless and AMI. The penalty is thus widespread, but its magnitude varies across detectors, motivating a closer examination of their architectures.

\paragraph{The largest penalties occur when decisions span multiple frames.}

On RealTalk, the highest $A_{\mathrm{within}}$ values occur in detectors that combine information across frames. This tendency extends beyond conventional temporal architectures: DFD-FCG aggregates independently extracted frame representations, while TALL processes multiple frames arranged into a single image, yet both exhibit pronounced penalties. Among detectors sharing a frozen CLIP backbone, DFD-FCG consistently yields higher $A_{\mathrm{within}}$ than the three frame-based ones (GenD-CLIP, DFD-HR and VLAForge) across all three datasets. The two temporal detectors with the smallest penalties are exceptions by design: TFCU processes groups of four consecutive frames jointly but classifies each frame and averages the decisions, and FTCN processes each spatial location separately over time ($1\times1$ spatial kernels), so it does not see the shape of the mouth. Since detectors also differ in input crop, pretraining, and training data (\cref{tab:detectors}), we isolate the role of temporal aggregation within a single model.

On Seamless, the temporal output of DFD-FCG reaches $A_{\mathrm{within}}=74.6$, whereas its spatial output alone shows no statistically supported penalty. Frame-based detectors also show a penalty (\cref{tab:detectors}), so temporal processing is not the only route to it, but within the same model it carries most of it. A plausible reason lies in what distinguishes the two states: they differ far more in how much the mouth moves than in how open it is in a single frame (supp. mat.). A detector that sees one frame can only respond to how a closed mouth looks; one that sees several can also respond to the mouth not moving.

\subsection{The Penalty Tracks Mouth Dynamics}
\label{sec:follows}

\paragraph{The excess false positives track mouth-related behavior.}

Speaking and non-speaking windows differ in more than lip articulation, so we describe each window with 43 features in six groups (mouth, head, eyes and gaze, upper-face expression, other facial regions, image quality) and measure, within participants, the share of $\Delta$FPR that each combination of groups explains. In the 16 detector-corpus pairs on RealTalk and Seamless that meet a prespecified criterion, adding all other groups to the mouth raises this share by at most one percentage point in 14 pairs, whereas adding the mouth to all other groups raises it by 18-64 points, with confidence intervals excluding zero, in 11 (supp. mat.). We therefore focus on mouth motion, quantified by $M_{\mathrm{dyn}}$ (Sec.~\ref{sec:measure}), which ranks speaking above non-speaking windows within participants in 96.2\% of pairs on RealTalk, 99.0\% on Seamless, and 95.2\% on AMI. This separation also appears on corpora whose speech annotations do not rely on lip motion, and with alternative mouth-motion descriptors (supp. mat.).

\paragraph{Scores track mouth motion within conversational states.}
The association extends beyond the speaking/non-speaking contrast. Regressing detector scores on $\mdyn$ separately within each state, with participant fixed effects, reveals negative slopes among non-speaking RealTalk windows for 11 of 15 evaluable detectors (95\% bootstrap confidence intervals excluding zero). DFD-FCG and TALL exhibit this association on all three corpora. For DFD-FCG, it persists after jointly controlling for head motion and landmark instability, under separate controls for non-lip facial motion, and with independently extracted mouth descriptors. For TALL, the association partly attenuates when controlling for landmark instability, while several I3D models are also sensitive to head motion (supp. mat.). Thus, the relationship is not merely a between-state contrast, although its specificity varies across detectors.

\paragraph{Mouth dynamics account statistically for much of the score gap.}

Adding $\mdyn$ to the participant fixed-effects regression of \cref{sec:metric} attenuates the non-speaking score gap $\Delta$ by a median of 67\% on RealTalk, 61\% on Seamless, and 49\% on AMI across evaluable detectors. On RealTalk development and Seamless, adding $\mdyn$ after head speed yields a further statistically supported attenuation in 28 of 30 detector--dataset comparisons, whereas adding head speed after $\mdyn$ yields none. Since speaking and non-speaking windows overlap little in $\mdyn$, these shares are linear projections; cubic, decile, and state-specific adjustments leave an equal or smaller gap for most detectors (supp. mat.). These reductions reflect statistical associations, not causal effects or changes in false-positive rates.

\subsection{Changing the Mouth Changes the Score}
\label{sec:changing-mouth}

Using LivePortrait~\cite{guo2024liveportrait}, we animate only the lips of each fake video with the mouth movements of a real video of the same person, scaled by a factor $\alpha$: still mouth ($\alpha=0$), natural amplitude ($\alpha=1$), and doubled motion ($\alpha=2$), which remains within the range of real speech. Head pose, driving clip, and rendering are identical across $\alpha$, so only the amount of lip motion changes. 

We focus on DFD-FCG and TALL, the two detectors whose within-state association holds on all three corpora. Changing lip motion alone can halve the detection rate: at FF95, going from a still mouth to doubled motion lowers the detection rate of the Celeb-DF++ test fakes~\cite{li2025celebdfpp} from 93.5\% to 50.3\% for DFD-FCG and from 72.3\% to 29.0\% for TALL, in every manipulation family. Re-animating the fakes at natural amplitude brings detection close to the original level (77.8\% and 49.8\%), suggesting that rendering artifacts alone do not explain the observed changes. Keeping the mouth still but as widely open as under doubled motion keeps detection high: the score follows mouth motion, not mouth appearance.

On authentic videos that the detectors flag as fake, the same edit works in reverse: doubling the lip motion turns 73\% of the false positives of DFD-FCG and 32\% of those of TALL back to real, against 24\% and 2.5\% of the unedited clips and almost none with a still mouth. These controls are detailed in the supplementary material, which also shows that the effect keeps its sign with a second lip-sync generator, MuseTalk~\cite{musetalk}, and with a non-neural warp of the lips.

\section{Mouth Dynamics Bias: FF++ Training Effects and Evaluation Limits}
\label{sec:origin}

The 16 detectors differ in their fake training data and pretraining, but all use FF++ real videos (\cref{sec:detectors}). We investigate whether FF++ could make mouth dynamics a shortcut (\cref{sec:ff_regularities}), how the composition of the training fakes affects the non-speaking penalty (\cref{sec:ffpp-training}), and whether FF++ evaluation can reliably identify less-penalizing decisions (\cref{sec:ffpp-selection}).

\subsection{Mouth-Dynamics Regularities in FF++}
\label{sec:ff_regularities}

\paragraph{Manipulation alters mouth dynamics.}

We compute $\mdyn$ on 10,709 three-second clips from the FF++ training split with facial landmarks detected in at least 90\% of frames (2,142 real, 8,567 fake). The median $\mdyn$ is 0.234 lower for fake clips (95\% CI $[0.215, 0.252]$), corresponding to a lip-aperture change rate approximately 79\% of that in real clips. The difference is largest for DF ($-0.394$), followed by NT ($-0.284$) and F2F ($-0.256$), and smallest for FS ($-0.021$), which does not resynthesize the mouth; the ordering is the same at 25 and 30\,fps. To assess whether this pattern depends on $\mdyn$, we repeat the comparison using three alternative mouth descriptors, two independent of the inner-lip area (supp. mat.). The reduction persists for DF, F2F and NT across all descriptors, whereas the FS result varies with the descriptor. With equal total weight assigned to real and fake clips, fakes account for 67\% (95\% CI $[59\%, 78\%]$) of clips in the lowest 2\% of $\mdyn$, compared with 21\% ($[17\%, 25\%]$) in the highest 2\%. In FF++, a still mouth thus predicts the label, although stillness alone is not evidence of
manipulation.

\paragraph{Authentic low mouth dynamics are rare in FF++.}

\begin{figure*}[t]
  \centering
  \includegraphics[width=\textwidth]{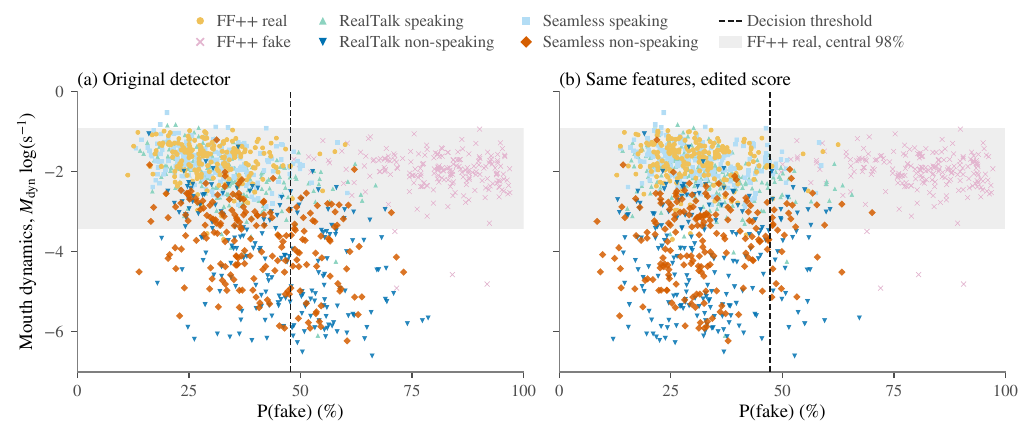}
 \caption{\textbf{Still mouths fall outside the range of FF++ real clips, and the score edit acts mainly there.}
 DFD-FCG scores for FF++ training clips and for windows of Seamless and RealTalk participants (200 random points per group). (a) Original detector, (b)~Score corrected with the mouth-dynamics correction, calibrated on a disjoint set of RealTalk participants (\cref{sec:edit-method}).}
  \label{fig:mdyn-geometry}
\end{figure*}

The median non-speaking window of RealTalk falls at the 0.47th percentile of FF++ real training clips (95\% CI $[0.09, 0.93]$): only about ten of the 2,142 real clips have comparably low measured mouth dynamics. The median speaking window falls at the 23.8th percentile, while non-speaking medians on Seamless and AMI fall at the 0.79th and 0.65th percentiles, respectively. As illustrated in \cref{fig:mdyn-geometry}(a), authentic non-speaking windows extend well below the range typically covered by FF++ real clips. These results persist when FF++ real clips are restricted to the frame rate of each corpus and when conversational windows are re-encoded and cropped as in FF++ (supp. mat.). As landmark noise contributes substantially at very low motion levels, these results reflect measured mouth dynamics rather than physical stillness. Thus, FF++ contains few real examples with low mouth dynamics, while such dynamics are associated with manipulation labels. This creates a plausible training shortcut that evaluation on the similarly distributed FF++ test set may fail to reveal.

\subsection{Training Manipulations Influence the Non-Speaking Penalty}
\label{sec:ffpp-training}

If the FF++ imbalance of \cref{sec:ff_regularities} shapes the penalty, changing the training fakes should change it. We test this with two retraining experiments on a single case study, DFD-FCG, chosen for its strong cross-dataset performance (95.0 video-level AUC on Celeb-DF v2), pronounced non-speaking penalty (\cref{tab:detectors}), and publicly available training recipe, which we reproduce (supp. mat.). Its frozen CLIP backbone keeps the pretrained weights fixed across conditions. In both experiments, we keep the real videos, architecture, and training protocol fixed, train three seeds per condition, and compare with training on all four methods with the same seeds.

\paragraph{Changing the fake generator changes the penalty.}
We first retrain the detector on one FF++ manipulation method at a time (DF, F2F, NT, or FS), with three seeds per method and the same real videos, architecture, and training protocol. On the 647 RealTalk development participants, training on DF, F2F, or NT yields penalties comparable to training on all four methods with the same seeds: $A_{\mathrm{within}}=78.3$, $78.4$, and $76.6$, versus $75.9$. In contrast, training on FS, whose mouth dynamics are closest to those of real videos (\cref{sec:ff_regularities}), yields $A_{\mathrm{within}}=51.6$, 26.7 points below DF (95\% CI $[21.6, 31.8]$, including variability across seeds). The DF-FS difference persists on the 97 Seamless test participants scored by DFD-FCG (30.6 points, $[20.5, 40.8]$; 80.5 versus 49.8).

\paragraph{The penalty grows with the mouth-dynamics cue in training.}

To separate mouth dynamics from other differences between methods, such as where their artifacts lie, we then vary mouth motion within each method. Since a fake inherits part of its motion from its source and target videos, we regress its $\mdyn$ on theirs within each method and keep the 250 fakes per method with the highest residuals (faithful) or the lowest (attenuated), about 35\% of the FF++ fakes in each arm (supp. mat.). This sets the strength of the cue: a real training clip has higher $\mdyn$ than a fake with probability 0.47 with faithful fakes and 0.82 with attenuated ones, against 0.66 with all fakes. The penalty follows the cue: attenuated fakes yield higher $A_{\mathrm{within}}$ than faithful ones in all three paired seeds, by 9.3 points on RealTalk (95\% CI $[3.6, 15.0]$; 79.5 versus 70.2) and 14.9 on Seamless test ($[9.8, 19.9]$; 81.6 versus 66.7), including variability across seeds, while training on all fakes falls in between (75.9 and 70.7). The penalty that remains with faithful fakes is consistent with the other side of the imbalance in \cref{sec:ff_regularities}: FF++ real clips rarely move as little as a listener's mouth. Both arms share methods and training size, but rendering differences that covary with mouth motion are not controlled.

\subsection{FF++ Is Blind to the Penalty}
\label{sec:ffpp-selection}

We now keep the learned representations fixed and ask whether FF++ can distinguish classification heads with different penalties. 

\paragraph{Heads without the penalty exist.}
Let $x$ denote the feature vector entering the detector's final classifier, and let $s_0(x)$ be the corresponding pre-sigmoid score produced by the released model. To test whether the penalty is inherent to the learned representations, we modify the classification head along a single feature direction $w$,
\begin{equation}
\label{eq:score-edit}
s_w(x) = s_0(x) - w^\top(x - \mu_{\mathrm{FF}}),
\end{equation}
where $\mu_{\mathrm{FF}}$ is the mean of the FF++ training features, real and fake pooled. For each of eight detectors (DFD-FCG, RealForensics, LipForensics, TALL, FakeSTormer, DFD-HR, GenD-PE, SBI; selection in supp. mat.), we obtain a head $s_1$ with $w=c\,(\Sigma_{\mathrm{FF}}+\lambda I)^{-1}d$, where $\Sigma_{\mathrm{FF}}$ is the covariance of the same FF++ features and $d$ the mean within-person feature difference between non-speaking and speaking windows of 323 RealTalk participants; $c$ cancels their mean within-person score gap, and $\lambda$, chosen on held-out FF++ identities, keeps changes to FF++ scores small (supp. mat.). On the other 324 development participants, $s_1$ brings $A_{\mathrm{within}}$ into $[45, 55]$ for seven of the eight detectors; for RealForensics, it drops from $86.2$ to $47.1$. The penalty is therefore not inherent to the frozen representations, but $s_1$ serves only as a reference, since it requires conversational data and speaking labels.

\paragraph{FF++ cannot reliably distinguish biased and less-biased heads.}
We compare $s_0$ and $s_1$ on the FF++ test set. Their AUCs differ by $0.01$-$0.12$ points in seven detectors, comparable to the difference between two trainings of the same DFD-FCG recipe (up to $0.115$), and by $0.48$ in FakeSTormer. The test set is also too small to show that the two heads make the same decisions: with 140 real videos, even observing no disagreement only shows that they disagree on fewer than 2.1\% of real videos (one-sided, $\alpha = 0.05$). To test whether this only reflects directions that FF++ barely varies, we decompose the features of the eight detectors into the subspace that holds 99\% of the FF++ feature variance and its complement. The penalty of $s_0$ comes almost entirely from this subspace, whereas 65--100\% of the correction in $s_1$ lies outside it. The correction does not need the complement, however: built within the subspace alone, it still brings $A_{\mathrm{within}}$ into $[45,55]$ in all eight detectors (FakeSTormer only at a large FF++ cost), and in six of them FF++ test AUC changes by at most $0.12$ points, as for $s_1$ (supp. mat.). In most detectors, FF++ thus cannot separate heads with and without the penalty even along the directions it covers.

\paragraph{Training on FF++ does not find them.}
We then retrain the head of each detector on FF++ alone, varying the loss, regularization, feature normalization, head architecture, and optimization, for a total of 285 heads per detector. In all detectors, the head ranked first by FF++ test AUC keeps $A_{\mathrm{within}}$ above 55 on the same 324 participants (training variants and per-head results in supp. mat.). The procedures we tested thus do not find the penalty-free head, although other FF++-only approaches cannot be ruled out.

\section{Mouth-Dynamics-Aware Score Correction}
\label{sec:score-edit}
FF++ performance does not reveal the non-speaking penalty (\cref{sec:ffpp-selection}), while the less-biased head $s_1$ requires speaking labels. We ask whether within-person mouth dynamics alone can guide a score correction without these labels or retraining detector representations. If mouth dynamics account for much of the penalty, removing their linear association with detector scores should substantially reduce it. We test this prediction on unseen participants and examine its limits in decision parity, cross-dataset transfer, and manipulation detection.

\subsection{A Within-Person Score Correction}
\label{sec:edit-method}

We retain the score edit of \cref{eq:score-edit}, leaving detector representations unchanged, but estimate $w$ from mouth dynamics rather than speaking labels: the correction minimizes changes to FF++ training scores while approximately cancelling the within-participant covariance between the edited score and $\mdyn$ on calibration data. We fit one correction per detector using 1,848 authentic RealTalk windows from 323 calibration participants. Fitting requires neither real/fake nor speaking labels, although speaking annotations were used to sample windows and select the descriptor during development. Because the correction targets a linear covariance on calibration participants rather than the ranking itself, its effect on $A_{\mathrm{within}}$ for unseen participants is not guaranteed by construction. \cref{fig:mdyn-geometry}(b) illustrates the edit for DFD-FCG. We apply the correction to 14 of the 16 detectors: it requires deterministic features and a linear final layer, which FTCN and VLAForge, respectively, lack. Details of the optimization, the development choices, and the exclusions are in the supplementary material.

\begin{table}[t]
\centering
\footnotesize
\setlength{\tabcolsep}{2.5pt}
\renewcommand{\arraystretch}{1.05}
\caption{\textbf{Mouth-dynamics-aware score correction.} 
$A_{\mathrm{within}}$ before (Orig.) and after (Corr.) the correction of \cref{sec:edit-method}, calibrated on RealTalk and applied without refitting.
Bold: within $[45,55]$. Both RealTalk columns use only the unseen participants.}

\label{tab:mouth-edit}
\begin{tabular}{@{}c@{\hspace{2pt}}lcccccc@{}}
\toprule
 & & \multicolumn{6}{c}{$A_{\mathrm{within}}$ (\%)}\\
\cmidrule(l){3-8}
 & & \multicolumn{2}{c}{RealTalk (unseen)} & \multicolumn{2}{c}{Seamless test} & \multicolumn{2}{c}{AMI}\\
\cmidrule(lr){3-4}\cmidrule(lr){5-6}\cmidrule(l){7-8}
 & Detector & Orig. & Corr. & Orig. & Corr. & Orig. & Corr.\\
\midrule
\multirow{9}{*}{\rotatebox[origin=c]{90}{Temporal}}
& RealForensics & 85.4 & \textbf{51.6} & 87.9 & 65.7 & 77.2 & \textbf{47.0}\\
& LipForensics  & 66.5 & \textbf{53.3} & 69.5 & 56.1 & 70.2 & 61.6\\
& I3D-DF40-FR   & 82.2 & \textbf{54.3} & 88.4 & 78.1 & 75.1 & \textbf{52.3}\\
& I3D-DF40-FS   & 80.6 & 40.5 & 83.9 & 56.6 & 68.9 & 29.5\\
& I3D-DF40-FSFR & 76.3 & \textbf{49.5} & 82.5 & 66.8 & 68.3 & 36.4\\
& TALL          & 76.6 & \textbf{51.0} & 75.6 & \textbf{51.0} & 71.8 & 60.9\\
& FakeSTormer   & 63.8 & \textbf{53.5} & 66.4 & 65.2 & \textbf{45.4} & 44.6\\
& TFCU          & 56.8 & 43.9 & \textbf{52.9} & 37.6 & 58.0 & \textbf{51.9}\\
& DFD-FCG       & 75.2 & \textbf{49.0} & 74.6 & \textbf{49.4} & 62.5 & 43.0\\
\midrule
\multirow{5}{*}{\rotatebox[origin=c]{90}{Frame-based}}
& DFD-HR        & 64.8 & \textbf{47.0} & 63.6 & \textbf{45.7} & 55.2 & \textbf{48.1}\\
& GenD-CLIP     & 64.1 & \textbf{48.4} & 58.1 & 42.3 & \textbf{53.5} & \textbf{46.8}\\
& GenD-PE       & 60.9 & \textbf{45.2} & 58.2 & 39.9 & 61.5 & \textbf{53.8}\\
& GenD-DINOv3   & 64.3 & \textbf{48.9} & 57.0 & 31.7 & \textbf{49.9} & 37.9\\
& SBI           & 63.8 & \textbf{50.3} & 64.7 & 58.1 & 60.9 & 56.9\\
\midrule
\multicolumn{2}{@{}l}{Median} & 65.7 & 49.3 & 68.0 & 53.5 & 62.0 & 47.5\\
\bottomrule
\end{tabular}
\end{table}
\subsection{The Correction Reduces the Penalty on Unseen Participants}
\label{sec:edit-results}

\paragraph{The ranking penalty approaches chance.}
We apply each correction, without refitting, to 290 unseen RealTalk participants, drawn from conversations that share no participants with those used for calibration and development. Across the 14 detectors, the median $A_{\mathrm{within}}$ drops from $65.7$ to $49.3$, and 12 of the 14 corrected point estimates fall within $[45,55]$ (\cref{tab:mouth-edit}). Only six of the 14 bootstrap confidence intervals lie entirely within the band, so it describes the point estimates rather than testing equivalence. As a reference, the head $s_1$ of \cref{sec:ffpp-selection}, cross-fitted on the same participants and built from speaking labels instead of $\mdyn$, reaches the band for 11 of the 14 detectors. Without any speaking label, mouth dynamics thus match what the labels themselves achieve.

Other signals do not produce the same reduction. On the 324 development participants not used for calibration, replacing $\mdyn$ with its within-participant shuffle brings a median of 3 of 14 detectors into the band across 20 shuffles, and replacing it with head motion residualized on $\mdyn$ brings only one. Random feature directions reach the band only with corrections about 20 times larger, which lower Celeb-DF v2 AUC by about 16 points, against $-0.20$ for the mouth-based correction. Confidence intervals and these controls are detailed in the supplementary material.

\subsection{Scope of the Correction}
\label{sec:limits}

The correction balances rankings, which does not by itself guarantee balanced decisions, transfer to other corpora, or unchanged fake detection. At each detector's FF95 threshold, recomputed for the corrected score, the median false-positive-rate gap between non-speaking and speaking windows of unseen RealTalk participants shrinks from $+10.6$ to $+3.9$ percentage points, and on AMI from $+9.8$ to $-1.4$. A positive gap remains statistically supported for 8 of 14 detectors on RealTalk, because non-speaking windows can keep a heavier upper tail even when their within-person ranking approaches chance. Applied without refitting, the RealTalk correction lowers the median $A_{\mathrm{within}}$ from $68.0$ to $53.5$ on Seamless test and from $62.0$ to $47.5$ on AMI, bringing 3 and 6 of 14 detectors into $[45,55]$ while over- or undercorrecting the others (\cref{tab:mouth-edit}); the cross-fitted $s_1$ fares no better on AMI (5 of 14), so a RealTalk calibration does not carry over unchanged to other corpora. Finally, median AUC changes are small on FF++ test ($-0.05$ points) and Celeb-DF v2 ($-0.20$), but motion-transfer fakes driven by non-speaking clips, which share the low-motion cue that the correction stops penalizing, lose detection at FF95 in all 14 detectors (median $-8.9$ points with TPSMM~\cite{zhao2022tpsmm}, the generator with the larger loss, and $-7.7$ with FOMM~\cite{siarohin2019fomm}; exploratory). These limits belong to a score-level correction on frozen features, which still encode mouth motion (supp. mat.).

\section{Conclusion}
\label{sec:conclusion}

Deepfake detectors can achieve high benchmark accuracy while partly relying on natural mouth motion: they assign higher fake scores to the same authentic person when the mouth moves less, and this dependence can be exploited to evade detection. Such a shortcut can arise when reduced mouth motion is associated with manipulation in training data but is rare in authentic examples, and a test set sharing this imbalance does not reveal it. Indeed, classifiers on the same representations can match in benchmark performance yet differ substantially in this bias.
Our score correction shows that this dependence can be reduced without retraining detector representations, but reducing it does not eliminate the bias. More balanced scores can coexist with unequal decisions, limited cross-dataset transfer, and weaker detection of certain manipulations. Removing the bias at its source will likely require real training videos to cover the full range of mouth motion. Ultimately, deepfake detectors should be evaluated not only for their ability to recognize manipulations but also for their reliability under natural variations in authentic facial behavior.

\section*{LLM Usage Statement}

OpenAI ChatGPT and Anthropic Claude were used to assist with language editing and consistency checks. All AI-assisted suggestions were reviewed by the authors, who remain responsible for the scientific content, experimental results, and references.

{
    \small
    \bibliographystyle{ieeenat_fullname}
    \bibliography{main}
}

\clearpage
\setcounter{page}{1}
\maketitlesupplementary
\appendix

\section{Datasets and Windows}
\label{supp:data}

\subsection{Windows and Speech Labels}
\label{supp:windows}

\Cref{sec:data} introduces the three conversational datasets and the speaking and non-speaking windows used throughout our evaluation. We detail how those windows were selected, which participants contribute to each analysis, and how the speech annotations were validated. These checks are particularly important for RealTalk, whose active-speaker labels rely partly on visual mouth motion.

\paragraph{Window counts and participant splits.}
\Cref{tab:supp-corpora} reports the recordings, participants, and speaking
and non-speaking windows used in the evaluation. Only participants with windows in both states contribute to $A_{\mathrm{within}}$. Unless otherwise specified, Seamless refers to its naturalistic test split throughout the paper. Of the 103 Seamless test participants meeting the minimum requirement of 30\,s of speech and 30\,s of non-speech, four yield eligible windows in only one state, leaving 99 participants for evaluation.

RealTalk is partitioned by conversation into a development set of 325 conversations (647 participants) and a held-out set of 145 conversations (290 participants). Each participant count here represents an eligible side of a conversation: the development set has 647 such observations out of 650 possible sides, and the held-out set has 290 out of 290, yielding the 937 observations reported in \cref{sec:data}. A RealTalk participant is one side of one conversation; since some individuals recur across conversations, conversations sharing a matched face (ArcFace cosine $\geq 0.35$) form connected components (228 in development, 135 held-out), which are the bootstrap units. The development participants are split by component into two disjoint groups (163 and 162 conversations): 323 are used to construct the classification heads of \cref{sec:ffpp-selection} and calibrate the score correction of \cref{sec:score-edit}, while the remaining 324 are used for their development evaluation. The retrained detectors of \cref{sec:ffpp-training} are instead evaluated on all 647 development participants.

The held-out set comes from 327 further conversations, from which we removed every conversation linked to a development participant: comparing ArcFace embeddings with a cosine threshold of 0.35, we discarded conversations matched to a development participant directly or through other held-out conversations (182 of 327). No held-out participant thus matches a development participant at this threshold. The resulting 145 conversations were reserved for evaluating the score
correction on participants excluded from development.

\Cref{tab:detectors} combines RealTalk development and held-out participants. \Cref{tab:supp-sets} specifies the dataset splits and participant populations used in each experimental analysis, distinguishing detector evaluation, model development, and held-out testing.

\begin{table}[h]
\centering\footnotesize
\setlength{\tabcolsep}{4pt}
\caption{\textbf{Conversational datasets used in main \cref{tab:detectors}.}
Only participants with eligible windows in both states are counted. Recordings correspond to individual recordings (Seamless), conversations (RealTalk), or meetings (AMI). Speech labels are obtained from participant audio, TalkNet active-speaker detection, and manual transcripts, respectively.}
\label{tab:supp-corpora}
\begin{tabular}{@{}lrrr@{}}
\toprule
 & Seamless & RealTalk & AMI \\
\midrule
Recordings & 210 & 470 & 121 \\
Participants & 99 & 937 & 124 \\
Speaking windows & 586 & 2{,}757 & 1{,}306 \\
Non-speaking windows & 573 & 2{,}749 & 1{,}419 \\
Speech labels & Audio & TalkNet & Transcripts \\
\bottomrule
\end{tabular}

\end{table}

\begin{table*}[h]
\centering
\small
\setlength{\tabcolsep}{5pt}
\renewcommand{\arraystretch}{1.15}
\caption{\textbf{Participants used in each experimental analysis.}
Counts refer to the analysis populations; the number of evaluable participants may vary across detectors due to preprocessing failures or unavailable mouth-dynamics measurements. For the evaluation in \cref{tab:detectors}, the I3D models, TALL, and FakeSTormer score 929, 934, and 935 of the 937 RealTalk participants, respectively (including 285, 287, and 289 of the 290 held-out participants); on Seamless, the I3D models, TALL, and DFD-FCG score 88, 89, and 97 of the 99
participants, respectively.}
\label{tab:supp-sets}
\begin{tabular}{@{}>{\raggedright\arraybackslash}p{0.25\textwidth}>{\raggedright\arraybackslash}p{0.17\textwidth}c>{\raggedright\arraybackslash}p{0.38\textwidth}@{}}
\toprule
Analysis & Dataset & Participants & Purpose \\
\midrule
Non-speaking penalty
(\cref{tab:detectors}, \cref{sec:bias})
& RealTalk (all) & 937 & Measure the non-speaking penalty across 16 released detectors \\
& Seamless & 99 & Assess the penalty using speech labels derived from audio \\
& AMI & 124 & Assess the penalty using manually annotated speech \\
\addlinespace[5pt]
Mouth dynamics and score gap
(\cref{sec:follows})
& RealTalk (all) & 917 & Quantify the association between mouth dynamics and detector scores \\
& Seamless & 96 & Test whether the association persists with audio-derived speech labels \\
& AMI & 123 & Test whether the association persists with manual speech annotations \\
\addlinespace[5pt]
Training-data experiments
(\cref{sec:ffpp-training})
& RealTalk (development) & 647 & Measure how changing FF++ training manipulations affects the penalty \\
& Seamless & 97 & Test whether the retraining effects persist across datasets \\
\addlinespace[5pt]
Classification heads $s_1$
(\cref{sec:ffpp-selection})
& RealTalk (development) & 323 & Construct alternative heads with reduced non-speaking penalties \\
& RealTalk (development) & 324 & Evaluate their penalties on participants excluded from head construction \\
\addlinespace[5pt]
Mouth-dynamics-aware correction
(\cref{sec:edit-method}, \cref{sec:edit-results})
& RealTalk (development) & 323 & Estimate the correction from within-person mouth dynamics \\
& RealTalk (development) & 324 & Test specificity against shuffled mouth dynamics and alternative signals \\
& RealTalk (held-out) & 290 & Evaluate the correction on participants excluded from development \\
& Seamless & 99 & \multirow{2}{*}{Assess cross-dataset transfer without recalibration} \\
& AMI & 124 & \\
\bottomrule
\end{tabular}

\end{table*}

\paragraph{Window selection.}
The selection rules were fixed before any detector was scored on conversational windows, first applied to 34 recordings of the Seamless development split, and then applied unchanged to the Seamless test split, RealTalk, and AMI. The criteria defining speaking and non-speaking windows are shared across datasets; speech-label sources and face-visibility checks differ by corpus.
\begin{enumerate}
\item \textbf{Candidate windows.} Windows are 4\,s long and sampled on a
0.5\,s grid (12 frames, or 0.48\,s, at RealTalk's 25\,fps). The first
5\,s and last 2\,s of each recording are excluded.
\item \textbf{Speaking windows.} A window is labeled speaking if the
participant speaks for at least 85\% of its duration. In Seamless and
AMI, speech segments separated by less than 0.3\,s are merged before
computing coverage. In RealTalk, coverage is computed from the per-frame current-speaker labels distributed with RealTalk, which designate one active speaker per frame from TalkNet scores.
\item \textbf{Non-speaking windows.} A window is labeled non-speaking if the participant does not speak during the window or within 1\,s before or after it.

\item \textbf{Participant eligibility.} Participants must have at least 30\,s of speaking and 30\,s of non-speaking material. For AMI, this criterion applies separately to each participant--meeting pair.
\item \textbf{Window sampling.} For each state, up to three
non-overlapping windows per recording are selected in randomized order
using a fixed seed. Seamless development windows, used only in the measurement checks of \cref{supp:measure,supp:compare}, include all non-overlapping windows of each recording, taken from left to right.
\item \textbf{Face-visibility checks.} In RealTalk, the participant's
face must be visible in all 100 frames, within the corresponding half
of the split-screen video. In AMI, a candidate is accepted only if a
face is detected in all five sampled frames, with a median
bounding-box side of at least 64\,px at 288 lines and a median
absolute yaw of at most $45^\circ$. At most 40 candidates per state
and participant--meeting pair are examined. No additional
sampled-frame visibility criterion is imposed on Seamless;
detector-specific face-crop coverage is examined in
\cref{supp:ci}.
\end{enumerate}
Participants contribute an average of 5.9 windows per state in
Seamless, 2.9 in RealTalk, and 11.0 in AMI, where participants may
appear in multiple meetings.

\paragraph{Dataset-specific exclusions.}
Before computing detector scores, Seamless test archives were selected using a greedy procedure that covered each participant with at least two recordings, or all available ones (50 archives, 113 participants, 333 recordings); evaluated participants contribute one to five recordings. This is the candidate pool, not the final evaluation sample: \cref{tab:supp-corpora} reports 210 recordings from 99 participants with eligible windows in both states. Three recordings with empty video files were subsequently excluded. In the RealTalk development set, 19 of 350 conversations contained no eligible participant, six were excluded because of non-monotonic timestamps, and 62 windows were removed after face-cropping failures. 
In AMI, four meeting series used to set the face-visibility checks and the video stream were excluded, with their participants, before any detector was run.

\paragraph{TalkNet label dependence.}
Because TalkNet uses visual lip information, its labels could bias the selection of RealTalk non-speaking windows toward reduced mouth motion.
To assess this possibility, we compared TalkNet labels with an audio-based speech-state estimate on 1{,}942 windows from 324 RealTalk development participants not used for calibration (162 conversations). The audio-based estimate combines voice activity detection on the mixed conversation audio with a speaker classifier enrolled separately for each conversation using segments labeled by TalkNet at least 10\,s away from the evaluated windows (2\,s for five short conversations).
The two labeling procedures agree on 94.1\% of non-speaking windows and 84.7\% of speaking windows. This comparison is not fully independent of TalkNet: speaker enrollment relies on its labels outside the evaluated windows and the classifier configuration was selected on four conversations based on agreement with TalkNet.

We therefore consider two additional checks. First, Seamless and AMI use speech labels derived from participant audio and manual transcripts, respectively, without visual information. Both exhibit the non-speaking penalty (\cref{tab:detectors}) and a separation of speaking and non-speaking windows by $\mdyn$ (\cref{sec:follows}). Second, we test whether the association between mouth dynamics and detector scores on RealTalk persists under stricter window selections. Among non-speaking windows, the within-participant slope of the detector score on $\mdyn$ has a 95\% interval below zero in 11 of the 15 evaluable detectors ($A_{\mathrm{within}}\geq55$ and at least 30 participants; \cref{supp:protocol}) with all windows, 11 when windows are further restricted by the participant's own TalkNet score (at most 5\% of frames with a positive score for non-speaking and at least 50\% for speaking windows), 10 in the half with the most negative TalkNet scores, and 11 of 14 on windows where TalkNet and the audio-based estimate agree (RealTalk development only). 
With the window-quality covariates of \cref{tab:supp-slopes-controls}, supported negative slopes fall from 9 of 15 with all windows to 5 of 15, 4 of 15,
and 4 of 14 in the three subsets, which retain 87\%, 38\%, and 28\% of the windows (876, 516, and 297 participants). The adjusted model has limited power
in this setting: for a simulated true slope of $-0.15$ it detects 6, 1, and 3 cells in the same subsets. The adjusted point estimates remain negative in 15,
15, and 12 of the cells, and DFD-FCG remains supported in every subset ($-0.57$, $-0.57$, $-0.42$).

\paragraph{Speech-coverage threshold and silence margin.}
We assessed the sensitivity of the non-speaking penalty to the speaking-window criterion by increasing the minimum speech coverage from 85\% to 90\% and 95\%. These thresholds retain subsets of the original speaking windows while leaving non-speaking windows unchanged. Across the 16 detectors, $A_{\mathrm{within}}$ changes by at most 0.5 percentage points on RealTalk and 3.3 on Seamless. Of the 60
detector--dataset--threshold combinations with a supported penalty at
85\%, only one loses support (GenD-DINOv3 on Seamless at 95\%). At the strictest threshold, 935 RealTalk and 92 Seamless participants retain windows in both states. AMI is not included, since its per-window speech coverage is not available.

The 1\,s silence margin around non-speaking windows excludes turn transitions, backchannels and brief pauses, so the non-speaking penalty and the score correction of \cref{sec:score-edit} refer to sustained periods of non-speaking.

\subsection{Deepfake Datasets}
\label{supp:deepfake-data}
\Cref{sec:data} introduces FF++, Celeb-DF++, and Celeb-DF v2 as the
datasets supporting our training, benchmark evaluation, and mouth-intervention experiments. Here we specify the subsets used for each purpose, distinguishing training and calibration populations from videos used for evaluation and controlled interventions. Detector-specific preprocessing is documented in \cref{supp:detectors}.

\paragraph{FF++.}
We use FF++ at compression level c23, following its official training
and test splits. Of the 720 real and 2{,}880 fake videos in the official
training split, 717 real and 2{,}868 fake videos (DF: 716; F2F: 717;
FS: 718; NT: 717) pass the training loader's face-tracking requirement. 
The remaining three real and twelve fake videos are automatically excluded because the face is tracked in fewer than 90\% of their frames.

The retained training videos serve three purposes. First, detector features extracted from the same clips and aggregated consistently with each detector's released scoring procedure are used to construct the classification heads of \cref{sec:ffpp-selection} and define the covariance metric for the score correction of \cref{sec:score-edit}.
In both cases, the ridge parameter is selected using FF++ training identities held out from fitting. Second, 3\,s clips with valid facial landmarks in at least 90\% of their frames are used to characterize the distribution of $\mdyn$ in \cref{sec:ff_regularities} (details in \cref{supp:ffpp}). Third, DFD-FCG is retrained using these training videos in \cref{sec:ffpp-training} (\cref{supp:retraining}).

The official test split contains 140 real and 560 fake videos. For
detectors other than DFD-FCG, FF95 is the linearly interpolated 95th
percentile of their video-level scores on the 140 real FF++ test
videos. Each test video is scored in full using the detector's own
sampling and aggregation (\cref{tab:supp-detectors}), so this threshold
is transferred from full videos to 4\,s windows. DFD-FCG instead uses its
clip-based, video-weighted FF95 calibration: each real FF++ test video
contributes up to three 3\,s clips, and FF95 is the weighted 95th
percentile over their scores, with equal total weight per video (each clip weighted by $1/n$ for a video with $n$ clips), matching
the single 3\,s clip that DFD-FCG scores per window; all clips are
used because first-clip scores are systematically lower. Its FF++
test metrics use the same clips. The decision threshold is recalibrated on other corpora only where explicitly stated (\cref{supp:recal}) and recomputed for the corrected scores of \cref{sec:score-edit}. All FF++ test AUCs (\cref{sec:ffpp-selection,sec:limits}) are computed on the complete set of 700 test videos.
Videos are not modified before scoring: each detector applies its
released face detection, alignment, cropping, and frame sampling procedures (\cref{supp:detectors}).

\paragraph{Celeb-DF++ and Celeb-DF v2.}
\Cref{tab:supp-cdfpp} summarizes the 5{,}240 fake videos in the
Celeb-DF++ test split, grouped by manipulation family and generator.
The Celeb-DF v2 test split is contained within Celeb-DF++: it comprises the same 178 real videos (108 celebrity and 70 YouTube videos) and 340 fake videos, identified by their file names in the Celeb-DF++ test list. Consequently, AUCs on the two datasets are not independent. The pretrained-checkpoint fidelity evaluation in \cref{supp:detectors} and the recipe-reproduction benchmark in \cref{supp:retraining} use video-level AUC on full Celeb-DF v2 test videos. Retrained last-checkpoint analyses that reuse the FF++ clip-based evaluation protocol are reported separately and use video-weighted clip-level AUC. The fidelity check (\cref{tab:supp-fidelity}) uses the 516-video list read by the DFD-FCG loader (\cref{supp:detectors}). AUC changes of retrained detectors,
heads, and corrections (\cref{sec:ffpp-training,sec:ffpp-selection,sec:score-edit}) use the Celeb-DF v2 videos of the Celeb-DF++ test split (178 real, 340 fake), fewer when a detector's preprocessing rejects videos (DFD-FCG: 174 real and 335 fake); restricting three other detectors to these 509 videos changes their AUC by at most 0.23 points.

\begin{table}[t]
\centering\footnotesize
\setlength{\tabcolsep}{2.3pt}
\caption{\textbf{Fake videos in the Celeb-DF++ test split
\cite{li2025celebdfpp}.} The split covers 22 generators: the 21
introduced by Celeb-DF++ and the face-swap method of Celeb-DF v2
\cite{li2020celebdf} (340 fakes). The 178 real test videos are shared
by both datasets.}
\label{tab:supp-cdfpp}
\begin{tabular}{@{}>{\raggedright\arraybackslash}p{0.21\linewidth}>{\raggedright\arraybackslash}p{0.48\linewidth}rr@{}}
\toprule
Family & Source & Fakes/source & Total \\
\midrule
Face swap & BlendFace, GHOST, HifiFace, InSwapper, MobileFaceSwap,
SimSwap, UniFace & 200 & 1{,}400 \\
& Celeb-DF v2 & 340 & 340 \\
Face reenactment & DaGAN, FSRT, HyperReenact, LIA, LivePortrait,
MCNET, TPSMM & 200 & 1{,}400 \\
Talking face & AniTalker, EchoMimic, EDTalk, FLOAT, IP\_LAP,
Real3DPortrait, SadTalker & 300 & 2{,}100 \\
\midrule
Total & 22 generators & -- & 5{,}240 \\
\bottomrule
\end{tabular}

\end{table}

\paragraph{Mouth interventions on Celeb-DF++.}
In \cref{sec:changing-mouth}, we modify the mouth dynamics of existing
Celeb-DF++ test fakes rather than generating new deepfakes. For each
fake, we select the first 3\,s window, comprising
$\mathrm{round}(3\cdot\mathrm{fps})$ frames at the native frame rate
(for nine videos at 29.6--29.9\,fps, 90 frames read at 30\,fps), in
which MediaPipe detects the face in at least 95\% of frames.

Each fake is animated using a real driving clip of the person depicted in the video. For face-swap and face-reenactment fakes, we identify this person using face-recognition embeddings; for talking-face fakes, we use the identity specified in the filename.
The driving clip comes from another real Celeb-DF video of the same person, excluding the real source videos used to generate the fake (\cref{supp:liveportrait}). The original window and all edited variants are re-encoded with x264 at CRF~18, using identical encoding settings across conditions.

Of the 5{,}240 Celeb-DF++ test fakes, 5{,}227 spanning 59 identities are edited. The remaining 13 are face-reenactment fakes with no 3\,s window in which MediaPipe detects the face in at least 95\% of frames.
Requiring valid scores for all three conditions (original, still mouth at $\alpha=0$, and doubled mouth motion at $\alpha=2$) leaves 5{,}193 fakes for DFD-FCG and 5{,}222 for TALL. The controls in \cref{supp:liveportrait} use 400 of these fakes, stratified by manipulation family and generation batch, covering all 59 identities.

\paragraph{Additional deepfake generation.}
In a separate experiment (\cref{sec:limits}), we generate new motion-transfer fakes of the 324 RealTalk development participants not used for calibration, driven by clips of their conversation partners, with two image-animation methods, TPSMM~\cite{zhao2022tpsmm} and FOMM~\cite{siarohin2019fomm}. The generation protocol is detailed in \cref{supp:mdyn-edit}.

\section{Detectors: Evaluation Protocol and Fidelity}
\label{supp:detectors}

\subsection{Detector Evaluation Protocol}
\Cref{sec:detectors} evaluates whether the non-speaking penalty extends across
16 detectors with different architectures and training strategies.
Here we document how their fixed pretrained checkpoints are used to
score the four-second conversational windows. The checkpoint sources,
model inputs, face preprocessing, and window-level aggregation are
summarized in \cref{tab:supp-detectors}. The separate DFD-FCG retraining experiments in \cref{sec:ffpp-training} are described in \cref{supp:retraining}; the frozen features used in \cref{sec:ffpp-selection} are specified in \cref{supp:heads-detectors}.

Each detector receives the same selected window, but applies its own
face processing and frame-sampling procedure. The table distinguishes
the released configuration from the choices made for our evaluation:
TALL and TFCU use the checkpoint sources indicated there; for GenD, DFD-HR, and VLAForge, face localization uses the FAN landmarks~\cite{bulat2017fan} already available in our pipeline. On Celeb-DF v2, DFD-HR reaches 94.9 AUC with its original dlib crops and 94.8 with FAN crops.
Changes to temporal sampling or score aggregation are marked separately. The resulting score is a single value per window, obtained from the frame- or clip-level predictions according to the detector-specific rule in the table.

\begin{table*}[h]
\centering
\caption{Detector checkpoints and evaluation pipelines used in our
experiments. Inputs, face preprocessing, and window scores refer to
the four-second conversational windows. Unless marked, configurations
follow the released implementation. $^{*}$ Checkpoint obtained from a
source other than the detector authors. $^{\dagger}$ Original face
localizer replaced by our FAN-landmark pipeline. $^{\ddagger}$ Window
sampling or score aggregation defined for our evaluation. Fidelity checks in \cref{tab:supp-fidelity} use the
authors' full-video inference
rules rather than these conversational-window aggregations.}
\label{tab:supp-detectors}
\begingroup
\scriptsize
\setlength{\tabcolsep}{3pt}
\renewcommand{\arraystretch}{1.08}
\begin{tabular}{@{}p{0.150\textwidth}p{0.200\textwidth}p{0.190\textwidth}p{0.265\textwidth}p{0.135\textwidth}@{}}
\hline
\textbf{Detector} & \textbf{Checkpoint (source)} & \textbf{Model input} & \textbf{Face preprocessing} & \textbf{Window score} \\
\hline
RealForensics & \texttt{realforensics\_}\newline\texttt{ff.pth} (authors) & 25 frames, RGB 112 & FAN-68, temporal smoothing, full-face crop (center 140 px) & Mean clip logit \\
LipForensics & \texttt{lipforensics\_}\newline\texttt{ff.pth} (authors) & 25 frames, grayscale mouth 88 & FAN-68, 5-point alignment, mouth crop & Mean clip logit \\
I3D-DF40-FR / FS / FSFR & \texttt{i3d.pth} (DF40 release) & 2 clips of 16 frames, RGB 224 & DeepfakeBench~\cite{yan2023deepfakebench}: dlib-81, scale 1.3, 256$\rightarrow$224 & Mean clip logit$^{\ddagger}$ \\
FTCN & \texttt{ftcn\_tt.pth} (authors) & Up to 8 evenly spaced clips of 32 frames (3 per 4 s)$^{\ddagger}$, RGB 224 & RetinaFace + tracking, aligned crop & Mean clip logit \\
TALL & \texttt{tall\_trainFF\_}\newline\texttt{testCDF.pth} (DeepfakeBench)$^{*}$ & 8 clips of 4 frames, 2$\times$2 thumbnail, 224 & DeepfakeBench: dlib-81, scale 1.3 & Mean clip logit$^{\ddagger}$ \\
FakeSTormer & \texttt{model\_best.pth}, C23 (authors) & 8 sub-clips of 4 frames, RGB 224 & RetinaFace, padding 0.25, affine alignment & Mean CLS logit \\
TFCU & Publicly available pretrained checkpoint$^{*}$ & 32 clips of 4 frames, RGB 224 & RetinaFace + tracking, ArcFace 5-point alignment & Mean probability \\
DFD-FCG & \texttt{weights.ckpt} (authors) & 10 frames from first 3 s$^{\ddagger}$, RGB 224 & 5-point alignment to mean face, 150-px crop & Logit margin$^{\ddagger}$ \\
DFD-HR & \texttt{ckpt\_best.pth} (authors) & 32 frames, 448, global + 4 tiles & DeepfakeBench crop from FAN landmarks$^{\dagger}$ & Mean frame logit$^{\ddagger}$ \\
VLAForge & \texttt{VLAForge.pth} (authors) & Per-frame, RGB 256 (224 for CLIP) & FAN-based face crop$^{\dagger}$ & Mean frame logit$^{\ddagger}$ \\
GenD-CLIP / PE / DINOv3 & GenD release (CLIP-L/14, PE-L, DINOv3-L) & 32 frames, RGB 256 & FAN landmarks, 5-point alignment, scale 1.3$^{\dagger}$ & Mean frame logit$^{\ddagger}$ \\
SBI & \texttt{FFc23.tar} (authors) & 32 frames, RGB 380 & RetinaFace, margin 1/8 & Mean frame probability \\
\hline
\end{tabular}
\endgroup
\end{table*}

The four-second window is therefore the common \emph{evaluation unit},
not necessarily the temporal span observed by every detector.
DFD-FCG uses the first three seconds, and LipForensics and TFCU do not
cover the final 0.67 and 0.77 seconds, respectively, at 30 fps. By
contrast, the mouth-dynamics descriptor is measured on the complete
four-second window. The state labels carry over to the first 3\,s
(at least 80\% speech in speaking windows), where $\mdyn$ has a
Spearman correlation of 0.93--0.97 with its full-window value and
separates the states within 1.0 point of it.

\subsection{Fidelity on Celeb-DF v2}
To check that the evaluated checkpoints retain credible benchmark
performance, we compare their video-level AUCs on Celeb-DF v2 with
published results (\cref{tab:supp-fidelity}). This check uses full-video
inference and the detector-specific evaluation rules, rather than the
four-second window scoring described above. It therefore assesses
checkpoint and implementation fidelity without conflating the two
sampling protocols.

\begin{table}[h]
\centering
\caption{Video-level AUC (\%) on Celeb-DF v2 using pretrained
checkpoints and full-video inference. For TALL, the two published
references are the DeepfakeBench checkpoint and the original model,
respectively.}
\label{tab:supp-fidelity}
\small
\setlength{\tabcolsep}{7pt}
\begin{tabular}{@{}lrr@{}}
\hline
\textbf{Detector} & \textbf{Ours} & \textbf{Published} \\
\hline
RealForensics & 89.6 & 86.9 \\
LipForensics & 82.5 & 82.4 \\
I3D-DF40-FSFR & 79.9 & 76.4 \\
I3D-DF40-FR & 67.7 & 68.4 \\
I3D-DF40-FS & 87.6 & 86.3 \\
FTCN & 89.2 & 86.9 \\
TALL & 85.6 & 83.1 / 90.8 \\
FakeSTormer & 92.1 & 92.4 \\
TFCU & 93.0 & 93.2 \\
DFD-FCG & 95.0 & 95.0 \\
DFD-HR & 94.8 & 96.0 \\
VLAForge & 95.6 & 96.8 \\
GenD-CLIP & 95.6 & 96.0 \\
GenD-PE & 95.8 & 95.8 \\
GenD-DINOv3 & 91.4 & 92.2 \\
SBI & 93.0 & 92.9 \\
\hline
\end{tabular}
\end{table}

The comparison uses the 516-video Celeb-DF v2 test list read by the
DFD-FCG evaluation pipeline (176 real and 340 fake), rather than all
518 videos in the official split. The two absent real videos are
\texttt{YouTube-real 00275} and \texttt{Celeb-real id39\_0005}. They
are absent from the test list released with the code, although both
decode correctly; we keep the list because it reproduces the published
DFD-FCG AUC of 95.0.  TALL and the I3D models score 515 and 514 videos, respectively, because their DeepfakeBench preprocessing rejects short videos. FTCN is checked on 50 real and 50 fake videos (AUC 89.2; 95\% bootstrap interval [82.4, 95.0]). Following its released inference rule, DFD-FCG assigns a score of 0.5 to nine videos rejected by preprocessing.

Relative to the appropriate published reference, every detector matches or exceeds the reported AUC or falls below it by at most 1.2 percentage points. For TALL, this comparison is with the DeepfakeBench checkpoint actually evaluated (85.6 versus 83.1), not the original TALL model (90.8). For the I3D variants, the reference values correspond to their respective DF40 training families (FR, FS, and FS+FR). The benchmark check supports the reported performance of these configurations; it does not require that their preprocessing produce identical crops on the conversational datasets.
\section{Mouth Dynamics Descriptor: Computation and Reliability}
\label{supp:measure}

Section~\ref{sec:measure} defines $\mdyn$ as the temporal variation of normalized inner-lip area, measured independently of the speech annotations; \cref{fig:supp-mdyn-steps} illustrates its computation on one speaking and one non-speaking window.
Here we specify how the descriptor and a spectral variant are computed and examine the extent to which image sampling, tracking, and head motion contribute to its value.
These checks are essential when interpreting the association between $\mdyn$ and detector scores in \cref{sec:follows}.

\subsection{Computing Mouth Dynamics}

\begin{figure}[t]
\centering
\includegraphics[width=\linewidth]{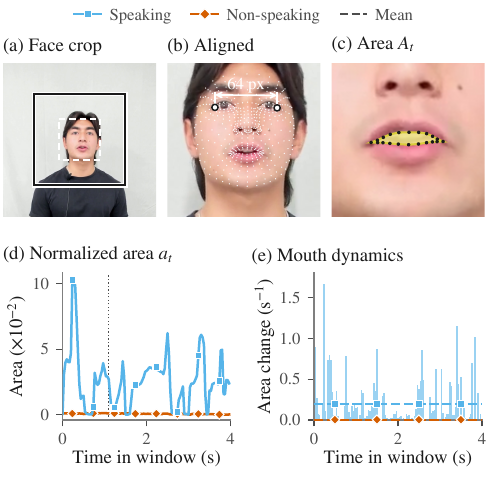}
\caption{\textbf{Mouth dynamics measures how much the inner-lip area changes between frames.}
Two 4-s windows (30\,fps) of the Seamless participant of \cref{fig:teaser}~\cite{seamless2025} (CC BY-NC 4.0), plotted from the start of each window.
(a)~Square crop, 2.2 times the median face box (dashed); (b)~MediaPipe landmarks, one similarity transform per window with the outer eye corners 64\,px apart; (c)~area $A_t$ of the 20 inner-lip landmarks; (d)~$a_t=A_t/64^2$, with the frame shown in (a)--(c) dotted; (e)~its absolute change per second, whose mean gives $\mdyn$ (\cref{eq:mdyn}): $-1.61$ speaking, $-5.63$ non-speaking.}
\label{fig:supp-mdyn-steps}
\end{figure}

\paragraph{Landmarks and alignment.}
For each clip we take a fixed square face crop, with side length 2.2 times the median face-box side, and resize it to 512\,px.
We extract landmarks with MediaPipe Face Mesh~\cite{kartynnik2019facemesh} (v0.10.11), running independently on every frame without temporal smoothing.
A single similarity transform, estimated from the median landmark configuration of the clip, aligns all frames to a common coordinate system with the outer eye corners (landmarks 33 and 263)
64\,px apart. Using a separate transform per frame would introduce additional variation from the reference landmarks.
The mouth area $A_t$ is obtained by the shoelace formula on the 20 inner-lip landmarks (78, 95, 88, 178, 87, 14, 317, 402, 318, 324, 308, 415, 310, 311, 312, 13, 82, 81, 80, 191). 
Dividing by $64^2$ gives the normalized area $a_t$ used in \cref{eq:mdyn} (\cref{fig:supp-mdyn-steps}(a)--(d)).

\paragraph{Missing frames and numerical stability.}
When a face is not detected, the missing $a_t$ values are linearly interpolated; clips with detections in fewer than 90\% of frames are excluded from mouth-dynamics measurement. We compute successive
differences after interpolation, thus using $T-1$ intervals for a $T$-frame clip (\cref{fig:supp-mdyn-steps}(e)). To keep the logarithm finite, the mean absolute frame-to-frame change is floored at $10^{-9}$. This floor is never
activated in our data: the smallest observed mean change in each conversational corpus and FF++ lies between $e^{-10.5}$ and $e^{-8.8}$.

\paragraph{Frame-rate effects.}
Conversational windows last 4\,s: RealTalk and AMI are sampled at 25\,fps, while Seamless is sampled at 30\,fps except for one participant at 43\,fps. FF++ clips last 3\,s at their native frame
rate (25 or 30\,fps for 97\% of real clips). 
Although the fps factor in \cref{eq:mdyn} expresses changes per second, frame-to-frame landmark noise also contributes to the measurement. For FF++ real training clips, median $\mdyn$ is 0.19 higher at 30 than at
25\,fps, close to $\log(30/25)=0.18$ if per-frame noise dominates.
This does not create a frame-rate difference between speaking and non-speaking windows of the same participant, but it matters when comparing FF++ with conversational corpora. The ordering of FF++
manipulation methods is unchanged at the two rates, and frame-rate matched comparisons are provided in \cref{supp:ffpp-robust}.
The score correction of \cref{sec:edit-method} uses within-participant covariances during calibration and then operates only on detector features (\cref{eq:score-edit}); a constant frame-rate offset within a participant therefore does not affect its fit. The limited transfer to AMI cannot be attributed solely to fps, as AMI and RealTalk both use 25\,fps (\cref{sec:limits}).

\paragraph{Syllabic-band power.}
\label{supp:plip}
As a spectral alternative to $\mdyn$, we sum the power of the same normalized area $a_t$ over the syllabic band, 
\begin{equation*}
\plip=\log\!\sum_{f\in\{3,4,5,6\}\,\mathrm{Hz}}\widehat{P}(f).
\end{equation*}
Here, $\widehat{P}$ is the Welch spectral estimate~\cite{welch1967} of $a_t$, computed from 1\,s Hann-windowed segments with approximately 50\% overlap after removing each segment's mean.
The four frequency bins span 2.5--6.5\,Hz. RealTalk and Seamless test are used to develop $\plip$.
In a separate confirmation on 400 Seamless development windows (200 per state), $\plip$ ranks speaking above non-speaking windows within participants in 96.8\% [94.3, 98.6] of pairs.
Applying the same rotation and translation to all frames leaves it nearly unchanged (Spearman correlation 0.976 [0.956, 0.980], lower of the two states).
The head-trajectory intervention of \cref{supp:head} changes $\plip$ by a median 0.039 [0.033, 0.047] of the speaking--non-speaking gap.
The band is not specific: an 8--12\,Hz control band also separates the states in 93--97\% of pairs, and the bands rank non-speaking windows similarly (Spearman 0.93--0.99).
On FF++ training clips (\cref{supp:ffpp}), $\plip$ and $\mdyn$ have a correlation of 0.92; with equal total class weights, the lowest 2\% of each descriptor have fake fractions of 0.68 and 0.67, respectively.
We use $\plip$ in the mouth-feature group and descriptor checks (\cref{supp:regions,tab:supp-compare}) and in the FF++ analyses (\cref{supp:ffpp,supp:ffpp-heads}).

\subsection{Reliability of the Measurement}

\paragraph{Temporal discontinuities.}
With a fixed crop and alignment, editing cuts or tracking failures may create abrupt changes in $a_t$ that inflate $\mdyn$.
A window is flagged if, between consecutive available landmark frames (normalized per frame across missing intervals), the eye midpoint shifts by more than 0.3 times the median interocular
distance (IOD), or the absolute difference in log IOD exceeds 0.15.
We also flag windows in which more than 10\% of frames have a landmark within 2\% of the crop boundary. 
These thresholds were set by inspecting large frame-to-frame discontinuities, without using detector scores. We examine the sensitivity of the results to less restrictive thresholds below.

Among windows with valid $\mdyn$, the flags cover 16.5\% of RealTalk development, 14.5\% of RealTalk held-out, 5.8\% of Seamless test, 3.2\% of Seamless development, and 8.8\% of AMI windows, as well as
1.2\% of FF++ real clips. 
On RealTalk development, flagged non-speaking windows have a higher median $\mdyn$ than unflagged ones ($-3.66$ versus $-4.22$), consistent with inflated estimates at discontinuities.

Windows flagged for temporal discontinuities are excluded from the mouth-dynamics comparisons with FF++, score-gap regressions, within-state slopes, and associated quality checks on RealTalk
development and both Seamless splits (\cref{supp:gap,supp:within,supp:head}). 
For the main Seamless comparison with FF++, 23 of the 24 flagged non-speaking windows are retained by measuring mouth dynamics within each uninterrupted shot (514 rather than 491 windows; \cref{supp:ffpp-robust}); excluding them leaves the percentile unchanged.
The corresponding analyses on RealTalk held-out and AMI retain these windows; applying the same screening to them (below) does not change the conclusions.
The detector evaluation in \cref{tab:detectors} uses all eligible windows, regardless of the discontinuity flags. The controlled head-motion interventions likewise retain their selected windows
without this screening.

Applying the same flags to RealTalk held-out and AMI leaves the pooled RealTalk non-speaking median at the 0.47th percentile of FF++ real clips ($\mdyn=-4.15$ instead of $-4.14$) and moves AMI from the 0.65th to the 0.56th percentile; speaking medians move by less than 0.6 percentile points.
The median share of the score gap explained by $\mdyn$ rises from 67\% to 75\% on pooled RealTalk and from 49\% to 59\% on AMI (53\% to 62\% when I3D-DF40-FR, not part of the main-text summaries, is added).
Supported negative non-speaking slopes remain 11 of 15 on  RealTalk and rise from 5 to 6 of 10 on AMI (6 to 7 of 11 with I3D-DF40-FR).
With more permissive thresholds (1.0\,IOD and 0.3 in log scale), the median shares are 73\% on RealTalk, 68\% on Seamless test and 50\% on AMI.

\paragraph{Head motion and landmark noise.}
\label{supp:head}
We next examine whether variations in $\mdyn$ can arise even when the lips do not articulate. In a controlled intervention we use 392 of 400 Seamless development windows (200 selected per state with a fixed seed), for which both manipulated and reference measurements are valid.
Applying the smoothed in-plane head trajectory (rotation, scale, translation) of a speaking window of another person, rather than a fixed transform, changes $\mdyn$ by a median 0.048 [0.041, 0.070] of the speaking--non-speaking state gap. The effect is 0.025 [0.021, 0.032] for speaking windows but 0.14 [0.12, 0.16] for non-speaking windows, where it increases the measured motion. 
The corresponding relative changes on non-speaking RealTalk development and Seamless test
windows are 0.11 and 0.10.

A second intervention holds the mouth image fixed. Repeating the first frame of 130 non-speaking windows and applying the window's head trajectory produces a mean absolute change of $a_t$ equal to a median 0.70 [0.64, 0.76] of that of the original window (higher in 26\% of cases). Using a speaking person's trajectory gives 0.98 [0.82, 1.07]. An entirely static input is not an informative control here, since MediaPipe in image mode returns the same landmarks for identical frames.

To estimate landmark instability, we repeat the measurement using
four versions of each frame: original, mirrored, shifted (by 2\% of the crop, scale 1.03), and mirrored-and-shifted (opposite shift, scale 0.97). Noise velocity is the mean absolute frame-to-frame change of each version's deviation from the four-version mean, averaged over versions and scaled by $\sqrt{4/3}$; measured velocity is the same change for the original $a_t$.

Estimated noise velocity relative to the measured velocity is 0.83 [0.81, 0.85], 0.87 [0.83, 0.90], and 0.87 [0.86, 0.90] for non-speaking windows of RealTalk development, Seamless test, and Seamless development, respectively.
The corresponding ratios are 0.29--0.35 for speaking windows and 0.27 [0.26, 0.28] for FF++ real clips (0.54 among the stillest 5\%). Within non-speaking windows of a participant, the log estimated noise velocity and $\mdyn$ correlate at 0.94--0.95, partly because the measurement contains noise.
These ratios are not exact noise fractions: shared artifacts such as compression jitter may be missed, and the replicate-based estimate can increase with genuine movement.

As a control for in-plane head motion, \emph{head speed} is the logarithm of the mean inter-frame quantity
\[
\frac{\|\Delta\mathbf c_t\|}{d}
+|\Delta\theta_t|
+|\Delta\log s_t|,
\]
where centroid $\mathbf c_t$, in-plane rotation $\theta_t$, and scale $s_t$ are obtained by fitting a similarity transformation from ten rigid eye/nose landmarks to their median configuration, and $d$ is the interocular distance.
The regional, descriptor, and incremental head-motion comparisons of \cref{supp:follows} use MediaPipe landmarks in video mode; the quality-adjusted slopes use the image-mode landmarks of $\mdyn$ (\emph{framewise} head speed).

Since 2D alignment does not remove out-of-plane head rotation, we also estimate yaw, pitch, and roll from MediaPipe's facial transformation matrix. Among non-speaking windows, within-person
correlations of yaw and pitch variability with $\mdyn$ are 0.30--0.52. 
However, on pooled RealTalk, Seamless test, and AMI, these variables add little beyond in-plane head motion and landmark noise ($\Delta R^2\leq 0.003$, upper limits $\leq 0.012$, versus
$R^2=0.77$--$0.78$ for the existing controls). 
Windows with above-median rotation have $\mdyn$ values 0.45--0.88 log units higher within participants, a descriptive difference that may also reflect mouth motion. Importantly, restricting to windows with low yaw and pitch variability preserves the within-person AUC of $\mdyn$ for distinguishing speaking from non-speaking: 97.2\%, 98.7\%, and 96.2\% on RealTalk, Seamless test, and AMI (449, 62, and 115 participants), versus 96.5\%, 98.8\%, and 95.5\% on all windows of those same participants. 
On the complete participant populations the corresponding values are 96.2\%, 99.0\%, and 95.2\% (\cref{sec:follows}).

These results distinguish the descriptor's two uses. The speaking--non-speaking separation, approximately 2 log units, is 7--10 times larger than the typical head-motion contribution, so $\mdyn$ reliably distinguishes the states and supports the FF++ comparison of \cref{sec:ff_regularities}. Within non-speaking periods, however, landmark instability and head motion account for much of its variation.
We therefore interpret $\mdyn$ as \emph{measured} mouth dynamics, not physical lip velocity, and interpret within-state detector-score associations only after adjustment for these measurement factors (\cref{supp:within}).
\section{Evaluation Metrics: Details}
\label{supp:protocol}

\Cref{sec:metric} defines three complementary views of the non-speaking
penalty: the within-participant ranking $A_{\mathrm{within}}$,
the within-participant score gap $\Delta$, and the false-positive
rate (FPR) at a decision threshold. We clarify how each quantity is
estimated and test whether the main observations depend on participant
weighting, editing cuts, or the threshold transferred from FF++.

\subsection{Ranking-Based Penalty and Sensitivity}

\paragraph{Statistical inference.}
The 95\% confidence intervals (CIs) behind \cref{tab:detectors} are pointwise and are not adjusted for multiple comparisons.
A non-speaking penalty is considered statistically supported when the lower confidence bound of $A_{\mathrm{within}}$ exceeds 50\%.
The weakest supported result is GenD-CLIP on AMI ($A_{\mathrm{within}}=53.5\%$, lower bound 50.03\%).

\paragraph{Participant weighting.}
As defined in \cref{eq:awithin}, each participant contributes equally.
Pooling all within-participant non-speaking/speaking pairs instead
weights each participant by $|\mathcal{L}_p||\mathcal{T}_p|$.
This changes $A_{\mathrm{within}}$ by at most 0.26 percentage points
on RealTalk, 1.7 on AMI and 6.5 on Seamless (DFD-HR: 63.6 to 57.1).
On Seamless, GenD-CLIP, GenD-PE and GenD-DINOv3 no longer show a
supported penalty (54.6, 54.4 and 53.7, respectively); all other
support decisions in \cref{tab:detectors} remain unchanged.

\paragraph{Temporal discontinuities and measurement coverage.}
The main results include windows flagged for temporal discontinuities
(\cref{supp:measure}). Excluding these windows (about 15\%, 4--5\% and 8\% of each detector's RealTalk, Seamless and AMI windows; same scores and bootstrap as \cref{tab:detectors}) changes $A_{\mathrm{within}}$ by at most 2.2 points (median absolute
change 0.4). In RealTalk, the estimate increases for 12 of 16
detectors (DFD-FCG: 76.3 to 77.5); only FTCN on RealTalk (52.1, lower bound 49.72) and GenD-CLIP on AMI (53.4, lower bound 49.95) lose statistical support across the 48 cells. Removing additionally the windows with insufficient landmarks to compute $\mdyn$ (another 1--7\%) changes the estimates by at most 2.4 points and also removes support for GenD-DINOv3 on Seamless (54.8, lower bound 49.99). Thus, the ranking penalty is not generally explained by these exclusions, although the smallest effects can be sensitive to them.

\subsection{Within-Participant Score Gap}

Unlike the rank-based $A_{\mathrm{within}}$, the score gap
$\Delta$ in \Cref{sec:metric} quantifies a linear association between
the detector score and the non-speaking indicator after removing
participant means. Detector scores, originally expressed as
probabilities, logits or margins, and covariates are converted to
rank-based normal scores within each detector--corpus sample:
\[
\Phi^{-1}\!\left(\frac{r-0.375}{n+0.25}\right),
\]
where $r$ is the rank, $n$ the number of windows, and $\Phi^{-1}$
the inverse standard normal cumulative distribution function.
This reduces sensitivity to extreme values and makes comparisons
across detectors more meaningful.

The coefficient $\Delta$ is estimated by least squares after within-participant centering, equivalently with participant fixed effects. Unlike $A_{\mathrm{within}}$, which gives participants equal weight, this coefficient weights individual windows equally; participants contributing more recordings can therefore have greater influence, despite the limit of three windows per state and recording (\cref{supp:windows}). We report the proportion of the gap accounted for by a covariate as
\[
1-\frac{\Delta_{\mathrm{adj}}}{\Delta},
\]
with 95\% intervals from 1{,}000 bootstrap replicates over person--session components.

The regressions use participants observed in both states and windows with finite scores and $\mdyn$ measurements. 
On RealTalk development and Seamless test, they additionally require a finite head speed (\cref{supp:head}) and exclude windows flagged for temporal discontinuities (\cref{supp:measure}).
A permutation control preserving participant identity and conversational state is reported in \cref{supp:gap}.

Counts of supported within-state slopes and median score-gap shares include only \emph{evaluable} detectors, with $A_{\mathrm{within}}\geq55\%$ on the analyzed windows and at least 30 participants: 15, 12, and 10 detectors on RealTalk, Seamless, and AMI, respectively.

\subsection{Decision Thresholds and Recalibration}
\label{supp:recal}

The FF95 threshold of \cref{sec:metric} is derived from real FF++ test
videos and then applied to conversational windows. To determine
whether differences between FF++ and conversational score
distributions explain the observed false-positive disparity, we
also recalibrate the threshold within each target corpus.

The procedure splits person--session components into two halves.
For each split, the 95th percentile of scores from all real
windows, speaking and non-speaking combined and weighted equally, is estimated on one half. On the other half we compute state-specific false-positive rates for each participant observed in both states, then average participants equally. 
The two halves exchange roles, and their rates are averaged. The split is fixed; each of 1{,}000 bootstrap replicates resamples components and re-estimates both thresholds, giving 95\% intervals.
For SBI on AMI, this analysis scores each window by the mean of the per-frame logits rather than the mean per-frame probability used in \cref{tab:supp-ci}; the two aggregations order windows differently.

Across detector--corpus combinations, the median overall FPR is
5.1\% (range 3.7--6.3\%). \Cref{tab:supp-recal} reports the
state-specific false-positive rates after recalibration, while
\cref{tab:supp-recal-ci} presents the corresponding non-speaking
minus speaking differences and their confidence intervals.

After corpus-specific recalibration, non-speaking windows retain higher false-positive rates for 16 of 16 detectors on RealTalk and 15 of 16 on both Seamless and AMI (\cref{tab:supp-recal}).
The differences are statistically supported for 16, 8, and 11 detectors, respectively (\cref{tab:supp-recal-ci}).
On AMI, this corresponds to 8 of 13 detectors after excluding the three I3D cells, computed in the same way but descriptive because their face-crop coverage is below 90\% (\cref{supp:ci}).
Thus, threshold transfer from FF++ does not fully account for the non-speaking penalty.

\begin{table}[h]
\centering
\scriptsize
\setlength{\tabcolsep}{3pt}
\caption{\textbf{False-positive rates after corpus-specific threshold calibration.}
False-positive rates (\%) for non-speaking (NS) and speaking (S)
windows at a target overall FPR of 5\%, cross-fitted over person--session
components. Bold indicates a positive NS--S difference with a 95\%
confidence interval excluding zero (\cref{tab:supp-recal-ci}).}
\label{tab:supp-recal}
\begin{tabular}{@{}lccc@{}}
\toprule
Detector & RealTalk & Seamless & AMI\\
\midrule
RealForensics  & \textbf{8.5 / 1.4} & \textbf{8.2 / 1.3} & \textbf{6.7 / 2.9} \\
LipForensics   & \textbf{7.9 / 2.4} & \textbf{9.9 / 2.1} & \textbf{8.6 / 2.4} \\
I3D-DF40-FR    & \textbf{7.8 / 2.0} & \textbf{9.2 / 1.4} & \textbf{8.9 / 1.8} \\
I3D-DF40-FS    & \textbf{7.5 / 2.7} & 6.5 / 4.0 & \textbf{5.3 / 3.0} \\
I3D-DF40-FSFR  & \textbf{6.3 / 3.2} & 5.3 / 2.9 & \textbf{7.0 / 2.1} \\
FTCN           & \textbf{5.7 / 4.5} & \textbf{6.8 / 2.5} & \textbf{7.1 / 3.4} \\
TALL           & \textbf{8.7 / 1.4} & \textbf{8.4 / 3.5} & \textbf{8.1 / 3.0} \\
FakeSTormer    & \textbf{6.3 / 3.9} & \textbf{6.8 / 2.7} & 4.8 / 4.6 \\
TFCU           & \textbf{7.2 / 2.9} & 6.0 / 4.6 & \textbf{7.1 / 4.0} \\
DFD-FCG        & \textbf{8.9 / 1.0} & \textbf{7.8 / 0.9} & \textbf{7.2 / 2.6} \\
DFD-HR         & \textbf{6.9 / 3.2} & 6.5 / 4.6 & 5.4 / 4.1 \\
VLAForge       & \textbf{6.7 / 3.2} & 7.0 / 6.0 & 5.8 / 4.5 \\
GenD-CLIP      & \textbf{7.3 / 2.8} & 4.1 / 4.7 & 4.8 / 4.9 \\
GenD-PE        & \textbf{7.1 / 3.1} & 6.3 / 6.0 & \textbf{6.8 / 4.1} \\
GenD-DINOv3    & \textbf{7.7 / 2.8} & 8.0 / 6.3 & 6.0 / 3.8 \\
SBI            & \textbf{7.9 / 2.3} & \textbf{9.0 / 5.1} & \textbf{6.7 / 2.6} \\
\midrule
Median & \textbf{7.4/2.8} & \textbf{6.9/3.8} & \textbf{6.7/3.2} \\
\textbf{Supported penalty} & \textbf{16/16} & \textbf{8/16} & \textbf{11/16} \\
\bottomrule
\end{tabular}
\end{table}

\begin{table}[h]
\centering
\scriptsize
\setlength{\tabcolsep}{1.2pt}
\caption{\textbf{Non-speaking minus speaking false-positive
rate after corpus-specific calibration.}
$\mathrm{FPR}_{\mathrm{NS}}-\mathrm{FPR}_{\mathrm{S}}$
(percentage points) for the pairs of \cref{tab:supp-recal},
with 95\% intervals from 1{,}000 bootstrap replicates of the
whole procedure over person--session components.
Bold: interval entirely above zero.
Median: median over detectors, with its interval from the
same replicates.
Two lower bounds close to zero are given to two decimals.}
\label{tab:supp-recal-ci}
\begin{tabular}{@{}lccc@{}}
\toprule
Detector & RealTalk & Seamless & AMI\\
\midrule
RealForensics  & \textbf{+7.1} [6.2, 8.3] & \textbf{+6.9} [4.6, 10.5] & \textbf{+3.8} [1.8, 5.8] \\
LipForensics   & \textbf{+5.5} [4.4, 6.7] & \textbf{+7.8} [5.5, 10.7] & \textbf{+6.2} [3.9, 8.2] \\
I3D-DF40-FR    & \textbf{+5.8} [4.7, 7.0] & \textbf{+7.8} [4.0, 11.3] & \textbf{+7.1} [4.8, 9.8] \\
I3D-DF40-FS    & \textbf{+4.8} [3.6, 6.0] & +2.5 [$-$0.1, 8.2] & \textbf{+2.3} [0.8, 5.1] \\
I3D-DF40-FSFR  & \textbf{+3.1} [2.0, 4.4] & +2.4 [$-$0.8, 8.7] & \textbf{+4.8} [3.1, 7.7] \\
FTCN           & \textbf{+1.2} [0.04, 2.2] & \textbf{+4.3} [1.9, 7.2] & \textbf{+3.7} [1.9, 8.1] \\
TALL           & \textbf{+7.2} [6.1, 7.9] & \textbf{+4.9} [2.2, 8.8] & \textbf{+5.1} [3.0, 8.4] \\
FakeSTormer    & \textbf{+2.4} [1.4, 3.7] & \textbf{+4.1} [1.8, 7.1] & +0.2 [$-$1.9, 2.6] \\
TFCU           & \textbf{+4.2} [2.8, 5.0] & +1.4 [$-$0.8, 3.7] & \textbf{+3.1} [1.0, 5.2] \\
DFD-FCG        & \textbf{+7.9} [6.9, 8.9] & \textbf{+6.9} [4.9, 12.1] & \textbf{+4.7} [3.0, 6.5] \\
DFD-HR         & \textbf{+3.7} [2.3, 4.9] & +1.8 [$-$0.9, 4.7] & +1.3 [$-$0.05, 2.8] \\
VLAForge       & \textbf{+3.5} [2.6, 5.0] & +1.0 [$-$0.9, 3.1] & +1.4 [$-$1.0, 3.6] \\
GenD-CLIP      & \textbf{+4.5} [3.4, 5.6] & $-$0.6 [$-$2.0, 1.9] & $-$0.1 [$-$1.9, 1.7] \\
GenD-PE        & \textbf{+4.0} [2.9, 5.2] & +0.2 [$-$3.4, 4.7] & \textbf{+2.6} [0.8, 4.3] \\
GenD-DINOv3    & \textbf{+4.9} [3.8, 5.9] & +1.8 [$-$0.8, 4.3] & +2.2 [$-$0.8, 4.4] \\
SBI            & \textbf{+5.6} [4.7, 6.7] & \textbf{+3.9} [0.3, 7.8] & \textbf{+4.1} [2.3, 5.5] \\
\midrule
Median & +4.7 [4.0, 5.2] & +3.2 [2.2, 5.2] & +3.4 [2.3, 4.4] \\
\bottomrule
\end{tabular}
\end{table}
\section{The Non-Speaking Penalty Across Detectors}
\label{supp:widespread}

\Cref{sec:widespread} reports a widespread non-speaking penalty and a larger
penalty in detectors that integrate information across frames. Here,
we provide uncertainty estimates for the detector comparison, assess
whether incomplete face preprocessing affects its interpretation,
and examine the role of temporal features and mouth opening.

\subsection{Confidence Intervals and Detector Coverage}
\label{supp:ci}

\Cref{tab:supp-ci} reports 95\% confidence intervals for the
$A_{\mathrm{within}}$ values in \cref{tab:detectors}, estimated with
2{,}000 bootstrap replicates over connected person--session
components (54 on Seamless, 363 on RealTalk, and 31 meeting series
on AMI). The underlying scores may be unavailable for some windows
when a detector's face preprocessing fails.

\begin{table}[htbp]
\centering
\scriptsize
\setlength{\tabcolsep}{2pt}
\caption{\textbf{$A_{\mathrm{within}}$ (\%) in \cref{tab:detectors}
with 95\% confidence intervals.} Intervals are bootstrapped over
person--session components. I3D on Seamless and AMI, and TALL on
Seamless, have face-crop coverage below 90\%; these cells are descriptive.}
\label{tab:supp-ci}
\resizebox{\linewidth}{!}{%

\begin{tabular}{@{}lccc@{}}
\toprule
Detector & Seamless test & RealTalk & AMI \\
\midrule
RealForensics & 87.9 [83.6, 91.8] & 86.2 [84.8, 87.5] & 77.2 [73.7, 80.6] \\
LipForensics & 69.5 [65.0, 74.2] & 67.5 [65.5, 69.4] & 70.2 [66.9, 73.5] \\
I3D-DF40-FR & 88.4 [84.2, 92.1] & 83.4 [81.9, 84.8] & 75.1 [71.8, 78.4] \\
I3D-DF40-FS & 83.9 [78.7, 88.5] & 82.7 [81.1, 84.2] & 68.9 [65.7, 72.1] \\
I3D-DF40-FSFR & 82.5 [77.7, 86.8] & 77.4 [75.6, 79.1] & 68.3 [65.3, 71.4] \\
FTCN & 58.8 [53.6, 64.3] & 52.3 [50.2, 54.4] & 60.4 [55.5, 65.0] \\
TALL & 75.6 [70.4, 80.9] & 77.8 [76.0, 79.5] & 71.8 [68.3, 75.4] \\
FakeSTormer & 66.4 [61.9, 70.9] & 63.2 [61.5, 64.9] & 45.4 [41.8, 49.2] \\
TFCU & 52.9 [48.0, 57.8] & 58.4 [56.6, 60.1] & 58.0 [54.6, 61.4] \\
DFD-FCG & 74.6 [70.8, 78.1] & 76.3 [74.7, 77.7] & 62.5 [58.8, 65.8] \\
DFD-HR & 63.6 [58.2, 69.0] & 66.9 [65.1, 68.8] & 55.2 [52.3, 58.0] \\
VLAForge & 53.8 [49.0, 58.5] & 60.0 [57.9, 62.0] & 55.8 [53.1, 58.5] \\
GenD-CLIP & 58.1 [53.1, 62.7] & 65.0 [62.9, 66.9] & 53.5 [50.03, 57.0] \\
GenD-PE & 58.2 [52.8, 63.4] & 63.0 [61.0, 64.9] & 61.5 [59.0, 64.1] \\
GenD-DINOv3 & 57.0 [51.5, 62.0] & 65.6 [63.7, 67.6] & 49.9 [47.0, 53.1] \\
SBI & 64.7 [60.0, 69.3] & 66.3 [64.4, 68.1] & 60.9 [58.1, 64.3] \\
\midrule
Participants & 97--99 (I3D 88, TALL 89) & 929--937 & 124 \\
\bottomrule
\end{tabular}%
}

\end{table}

\paragraph{Coverage and sensitivity.}
We use a 90\% valid-face-crop threshold, the coverage rule applied throughout our evaluation, to distinguish adequately covered detector--corpus cells from descriptive estimates; counts at 80\% and 95\% are given below. The three I3D models fall below 90\% on Seamless (83.3\%) and AMI (83.4\%), as does TALL on Seamless (87.6\%). These estimates describe the successfully processed windows, not necessarily the full population.

The 14/16 counts reported for Seamless and AMI in \cref{sec:widespread} include
these low-coverage cells. Retaining only cells with at least 90\% coverage yields supported penalties in 10/12 detectors on Seamless and 11/13 on AMI. All cells excluded by this restriction had a supported penalty, so the qualitative finding remains widespread. Lowering the cutoff to 80\% restores the main-text counts, whereas increasing it to 95\% gives 10/12 on both corpora (TALL on AMI, with 91.6\% coverage, is then excluded).

Coverage is also state-dependent for I3D on AMI: 76.7\% of
non-speaking windows and 90.7\% of speaking windows are processed,
although all 124 participants retain both states. Thus, selection of
successfully cropped windows may bias the comparison. Restricting
the other detectors to the same successfully cropped windows shifts
their $A_{\mathrm{within}}$ by $-3.4$ to $+1.9$ percentage points.
The corpus-specific recalibration on AMI includes the three
I3D models (\cref{tab:supp-recal,tab:supp-recal-ci}).
Given their 83.4\% face-crop coverage, these estimates
remain descriptive.

\subsection{Temporal Features and Mouth Opening}

\paragraph{Temporal and spatial heads of DFD-FCG.}
DFD-FCG combines spatial, temporal, and joint classification heads by averaging their probabilities. On 97 Seamless test participants, this combined final output yields $A_{\mathrm{within}}=74.6$ [70.8, 78.1], matching the DFD-FCG value reported in \cref{tab:detectors}.
To isolate the contribution of temporal processing, we additionally evaluate the component heads on the same participants. The temporal head yields 75.8 [72.2, 79.3], whereas the spatial head yields 53.4 [47.9, 58.9], with a paired temporal--spatial difference of 22.4 [16.0, 28.5] points.

The temporal head also has the larger penalty on RealTalk (73.9 [72.4, 75.4] versus 63.7 [61.8, 65.6]; difference 10.2 [8.0, 12.4], 937 participants) and AMI (64.8 [61.5, 68.1] versus 48.8 [45.4, 52.2]; difference 16.0 [12.6, 19.2], 124 participants). The spatial head is unsupported on Seamless and AMI, but also shows a smaller supported penalty on RealTalk (the joint head reaches 78.2 [76.6, 79.6] there). Thus, temporal features contribute more strongly across all three corpora, without being the sole contributor on RealTalk.

\paragraph{Mouth dynamics versus static opening.}
The two conversational states differ more in the amount of mouth movement than in average mouth opening. In the descriptor-comparison samples of \cref{tab:supp-compare} (RealTalk development windows further restricted by TalkNet score as in \cref{supp:windows}, and Seamless test), $\mdyn$ ranks speaking above non-speaking windows in 98.8\% and 97.8\% of within-participant pairs, respectively, compared with 87.5\% and 82.6\% for the window mean MediaPipe jaw-opening blendshape. These samples differ from the main evaluation populations, where $\mdyn$ separates the states in 96.2\% on RealTalk and 99.0\% on Seamless (\cref{supp:compare}).

On AMI, which was not used to select $\mdyn$, the same comparison on 2{,}394 windows from 123 participants gives 95.2\% [93.2, 96.9] for $\mdyn$ and 83.4\% [79.5, 86.5] for mean jaw opening; their paired difference is 11.8 [8.5, 15.2] percentage points, with 1{,}000 bootstrap replicates over meeting series. The interventions of \cref{supp:liveportrait} further distinguish motion from opening: holding the mouth still at an opening reached under doubled motion preserves detection, whereas increased lip motion reduces it. Static opening alone therefore cannot account for the observed sensitivity to mouth dynamics. 
\section{The Penalty Tracks Mouth Dynamics}
\label{supp:follows}

\Cref{sec:follows} links the non-speaking penalty to
measured mouth dynamics, showing that this association
extends beyond differences between conversational states.
Here, we assess the strength and specificity of this evidence,
examining whether it persists under alternative explanations,
measurement choices, and potential confounding factors.
Shared procedures are described in
\cref{supp:measure,supp:protocol}.

\begin{table*}[!t]
\centering\footnotesize
\setlength{\tabcolsep}{4pt}
\renewcommand{\arraystretch}{1.05}
\caption{\textbf{Incremental contribution of mouth features
to the non-speaking false-positive gap.}
Changes in the share of $\Delta$FPR accounted for by feature groups (percentage points) at FF95.
+ Mouth: $100(\rho_{\mathrm{all}}-\rho_{\mathrm{nonmouth}})$;
+ Others: $100(\rho_{\mathrm{all}}-\rho_{\mathrm{mouth}})$, where $\rho_G$ is the share accounted for by feature set $G$ (\cref{supp:follows-details}).
Brackets report paired 95\% bootstrap confidence intervals with out-of-fold predictions held fixed; bold indicates intervals excluding zero.
FPR: non-speaking / speaking false-positive rates (\%) on the analyzed real windows; FP: number of false positives.}
\label{tab:supp-regions}
\begin{tabular}{@{}lrrcc@{}}
\toprule
Detector & FPR (NS / S) & FP & + Mouth [95\% CI] & + Others [95\% CI] \\
\midrule
\multicolumn{5}{@{}l}{\textit{RealTalk development}}\\
\addlinespace[2pt]
DFD-FCG & 43.4 / 12.8 & 1037 & $\mathbf{+29.1}\,[18.3,39.9]$ & $-2.5\,[-8.0,3.4]$ \\
TALL & 53.3 / 22.4 & 1389 & $\mathbf{+59.5}\,[46.2,72.4]$ & $\mathbf{-5.5}\,[-11.8,-0.03]$ \\
RealForensics & 22.6 / 4.3 & 496 & $\mathbf{+45.7}\,[33.8,59.5]$ & $-4.2\,[-11.6,3.0]$ \\
LipForensics & 10.5 / 3.5 & 258 & $\mathbf{+54.1}\,[24.4,90.5]$ & $-4.6\,[-22.8,12.8]$ \\
I3D-DF40-FSFR & 65.4 / 33.6 & 1798 & $\mathbf{+17.7}\,[4.7,30.6]$ & $+0.2\,[-5.5,6.3]$ \\
FakeSTormer & 7.5 / 4.4 & 219 & $+37.7\,[-14.1,115.9]$ & $-0.3\,[-30.4,27.3]$ \\
TFCU & 17.9 / 12.1 & 555 & $+10.6\,[-29.1,52.1]$ & $+0.8\,[-26.8,27.2]$ \\
DFD-HR & 28.2 / 15.5 & 806 & $\mathbf{+30.2}\,[12.9,49.9]$ & $-4.9\,[-15.7,6.2]$ \\
GenD-CLIP & 26.8 / 16.1 & 792 & $\mathbf{+28.1}\,[5.9,51.8]$ & $-11.8\,[-27.9,2.5]$ \\
GenD-PE & 27.8 / 18.2 & 849 & $\mathbf{+64.3}\,[35.1,98.0]$ & $-9.3\,[-25.8,5.8]$ \\
GenD-DINOv3 & 46.7 / 32.9 & 1471 & $\mathbf{+57.3}\,[33.0,82.8]$ & $-9.1\,[-21.2,0.9]$ \\
SBI & 65.6 / 51.8 & 2169 & $\mathbf{+47.5}\,[26.4,71.2]$ & $-10.5\,[-25.0,1.7]$ \\
\midrule
\multicolumn{5}{@{}l}{\textit{Seamless test}}\\
\addlinespace[2pt]
DFD-FCG & 28.9 / 4.4 & 170 & $+25.5\,[-7.6,67.6]$ & $-0.8\,[-17.8,17.3]$ \\
RealForensics & 54.2 / 13.8 & 350 & $\mathbf{+31.9}\,[15.6,51.8]$ & $-6.7\,[-17.7,4.6]$ \\
I3D-DF40-FSFR & 45.0 / 19.8 & 298 & $+33.1\,[-4.0,96.7]$ & $+10.3\,[-6.9,30.4]$ \\
I3D-DF40-FR & 65.5 / 14.0 & 362 & $+13.4\,[-7.6,45.0]$ & $\mathbf{+14.6}\,[5.7,26.2]$ \\
\bottomrule
\end{tabular}
\end{table*}

\begin{table*}[h]
\centering
\footnotesize
\setlength{\tabcolsep}{4pt}
\renewcommand{\arraystretch}{1.05}
\caption{\textbf{Validation of the mouth-dynamics descriptor.}
(a) Within-participant speaking--non-speaking separation, absolute correlation with head speed, and recovery of the predictive $R^2$ of the ten mouth features. (b) Fraction of edited--sham pairs in which the descriptor increases, and split-half reliability.
RT, SD, and ST denote RealTalk development, Seamless development, and Seamless test, respectively. 
The comparison uses a per-frame variant of $\mdyn$ without the frame-rate factor in \cref{eq:mdyn}; since frame rate is constant within each recording, within-participant comparisons are unaffected.
Benchmark samples and speech labels differ from the main evaluation (\cref{supp:follows-details}).
Bold indicates the best available result per dataset and metric; NE denotes unavailable measurements. Head speed serves as a reference control.}
\label{tab:supp-compare}

\begin{minipage}[t]{0.56\textwidth}
\centering
\resizebox{\linewidth}{!}{%
\begin{tabular}{@{}lccc@{}}
\toprule
\multicolumn{4}{@{}l}{\textit{(a) Natural-window comparison}}\\
\addlinespace[2pt]
Descriptor & Separation (\%) $\uparrow$ & $|\rho_{\text{head}}|\downarrow$ & Recovery $\uparrow$ \\
 & RT / SD / ST & RT / SD / ST & RT / ST \\
\midrule
$\mdyn$ & \textbf{98.8} / \textbf{98.6} / 97.8 & 0.52 / 0.42 / 0.40 & \textbf{0.82} / \textbf{0.92} \\
$\plip$ (\cref{supp:plip}) & 98.0 / 97.9 / \textbf{98.6} & 0.49 / 0.39 / 0.34 & 0.78 / 0.88 \\
Tracked lip points~\cite{karaev2024cotracker3} & 98.4 / 98.3 / 98.0 & 0.57 / 0.55 / 0.53 & 0.70 / 0.84 \\
Phase-based lip motion~\cite{wadhwa2013phase} & 94.9 / 95.1 / 96.3 & 0.62 / 0.62 / 0.58 & 0.65 / 0.85 \\
Aperture variation, FAN~\cite{bulat2017fan} & 96.4 / 94.3 / 93.6 & 0.33 / \textbf{0.18} / 0.27 & 0.68 / 0.70 \\
Lip-separation AU25~\cite{chang2024libreface} & 94.9 / 95.1 / NE & 0.39 / 0.35 / NE & 0.73 / NE \\
3D jaw angle~\cite{retsinas2024smirk} & 97.9 / 95.6 / NE & 0.60 / 0.50 / NE & 0.67 / NE \\
Static mouth opening (level) & 87.5 / NE / 82.6 & \textbf{0.22} / NE / \textbf{0.12} & 0.76 / 0.58 \\
Pixel energy of the mouth & 84.0 / NE / 79.7 & 0.56 / NE / 0.41 & 0.61 / 0.74 \\
\midrule
Head speed (control) & 82.3 / 81.1 / 80.1 & reference & 0.32 / 0.66 \\
\bottomrule
\end{tabular}%
}
\end{minipage}\hfill
\begin{minipage}[t]{0.42\textwidth}
\centering
\resizebox{\linewidth}{!}{%
\begin{tabular}{@{}lcc@{}}
\toprule
\multicolumn{3}{@{}l}{\textit{(b) Controlled-motion response and split-half reliability}}\\
\addlinespace[2pt]
Descriptor & Controlled motion $\uparrow$ & Reliability $\uparrow$ \\
 & 0.01 IOD & RT / SD / ST \\
\midrule
$\mdyn$ & 0.76 & 0.67 / \textbf{0.75} / \textbf{0.73} \\
$\plip$ (\cref{supp:plip}) & 0.76 & 0.58 / 0.65 / 0.62 \\
Tracked lip points~\cite{karaev2024cotracker3} & \textbf{0.89} & 0.53 / 0.55 / 0.60 \\
Phase-based lip motion~\cite{wadhwa2013phase} & 0.88 & 0.46 / 0.51 / 0.56 \\
Aperture variation, FAN~\cite{bulat2017fan} & NE & 0.27 / 0.29 / 0.37 \\
Lip-separation AU25~\cite{chang2024libreface} & 0.68 & \textbf{0.68} / 0.72 / NE \\
3D jaw angle~\cite{retsinas2024smirk} & 0.67 & 0.39 / 0.51 / NE \\
\bottomrule
\end{tabular}%
}
\end{minipage}
\end{table*}

\subsection{Mouth Features and the False-Positive Gap}
\label{supp:regions}

To examine the contribution of mouth features to the non-speaking false-positive gap, each window is characterized by 43 features in six groups: mouth (10 features), head, eyes and gaze, upper-face expression, other facial
regions, and image quality (33 features combined). We compare the share of the within-participant false-positive gap at FF95 accounted for by mouth features alone, the other groups, and all features together.

The two comparisons reveal a clear asymmetry.
Adding mouth features to the other five feature groups increases the share of the false-positive gap accounted for in all 16 eligible detector--dataset combinations (\cref{supp:follows-details}).
This additional contribution is statistically supported in 11 combinations, ranging from 17.7 to 64.3 percentage points (\cref{tab:supp-regions}).

In contrast, once mouth features are included, adding the other five groups increases this share by at most one percentage point in 14 of 16 combinations. 
Only I3D-DF40-FR on Seamless shows a statistically supported positive increment.

This asymmetry indicates that mouth features capture information about the non-speaking false-positive gap that is largely absent from other facial features.
Because the mouth group includes both appearance and motion, we next examine the contribution of mouth dynamics specifically.

\subsection{Validating the Mouth-Dynamics Descriptor}
\label{supp:compare}

We evaluate whether $\mdyn$ provides a useful, interpretable summary of mouth dynamics.
On the main evaluation windows, it ranks speaking above non-speaking within participants in 96.2\% [95.3, 97.1] of pairs on RealTalk, 99.0\% [97.8, 99.8] on Seamless, and 95.2\% on AMI (\cref{sec:follows}).
The Seamless and AMI results use audio- or transcript-based speech labels, independently of the lip-based labels used on RealTalk.

\Cref{tab:supp-compare} compares $\mdyn$ with alternative mouth-motion descriptors and a head-speed control on separate benchmark samples (\cref{supp:follows-details}).
Five descriptors apply the band power of $\plip$ (\cref{supp:plip}) to different per-frame mouth signals.
$\plip$ uses the inner-lip area $a_t$. Tracked lip points use the vertical separation between 18 lower-lip and 18 upper-lip points in canonical coordinates, tracked by CoTracker3~\cite{karaev2024cotracker3} from MediaPipe initializations; this descriptor is  therefore not fully independent of landmark extraction.
Phase-based lip motion~\cite{wadhwa2013phase} uses the vertical displacement of the lower relative to the upper vermilion, accumulated from the phase of a complex Gabor filter ($\lambda=32$\,px), which cancels common rigid motion.
Lip-separation AU25 uses the action-unit intensity
estimated by LibreFace~\cite{chang2024libreface},
from a separate full-frame extraction on RealTalk.
The 3D jaw angle uses the opening component of the
FLAME jaw pose estimated by SMIRK~\cite{retsinas2024smirk}.
Static mouth opening is the window mean of the
MediaPipe jaw-opening blendshape.
FAN aperture variation and mouth pixel energy are
defined with the regional features in
\cref{supp:follows-details}, and head speed in
\cref{supp:head}.
Despite using a single measure, $\mdyn$ recovers
82\% and 92\% of the predictive $R^2$ of the
ten mouth features for false positives on
RealTalk development and Seamless test, respectively.
This measures predictive information rather than
the fraction of the non-speaking false-positive
gap accounted for in \cref{supp:regions}.

Controlled-motion interventions and split-half
reliability provide complementary validation:
$\mdyn$ responds to imposed lip motion and shows
relatively stable measurements across window halves.
However, tracked-point and phase-based descriptors
respond more consistently to imposed motion,
AU25 is slightly more reliable on RealTalk,
and FAN aperture variation correlates less with
head speed.

Overall, $\mdyn$ provides a compact summary of
mouth dynamics, but is not an isolated measure
of lip movement; its sensitivity to head motion
and landmark instability is examined in
\cref{supp:head}.

\begin{table*}[!t]
\centering
\footnotesize
\setlength{\tabcolsep}{4pt}
\renewcommand{\arraystretch}{1.18}
\definecolor{tabgray}{gray}{0.95}
\newcommand{\slopeCI}[2]{%
  \shortstack[c]{$#1$\\[-1pt]{\scriptsize$[#2]$}}%
}
\caption{\textbf{Within-state associations in non-speaking windows before and after adjustment.}
Participant fixed-effects slopes of rank-normalized detector scores on rank-normalized $\mdyn$ in non-speaking windows. Adjusted estimates additionally control for
landmark instability, head speed, sharpness, and encoding; raw and
adjusted fits use identical complete-case windows.
Brackets show 95\% person--session bootstrap confidence intervals.
Bold indicates supported negative slopes (interval entirely below zero).
$^{\dagger}$: not evaluable ($A_{\mathrm{within}}<55\%$); $^{\S}$: not part of the main-text summaries (\cref{supp:measure}).
Marked cells are excluded from counts; as in the main text, counts include cells with face-crop coverage below 90\% (\cref{supp:ci}).
Retained: initially supported negative slopes remaining supported after adjustment.
Bounds near zero retain extra precision (\cref{supp:follows-details,supp:ci}).}
\label{tab:supp-slopes-controls}
\begin{tabular}{@{}l rr@{\hspace{7pt}} rr@{\hspace{7pt}} rr@{}}
\toprule
 & \multicolumn{2}{c}{RealTalk} & \multicolumn{2}{c}{Seamless test} & \multicolumn{2}{c}{AMI} \\
\cmidrule(lr){2-3}\cmidrule(lr){4-5}\cmidrule(lr){6-7}
Detector & Raw & Adjusted & Raw & Adjusted & Raw & Adjusted \\
\midrule
RealForensics & \slopeCI{\mathbf{-0.18}}{-0.25,-0.12} & \slopeCI{\mathbf{-0.14}}{-0.23,-0.04} & \slopeCI{-0.10}{-0.22,0.01} & \slopeCI{+0.25}{0.02,0.45} & \slopeCI{+0.08}{0.01,0.14} & \slopeCI{+0.15}{-0.03,0.30} \\
\rowcolor{tabgray}
LipForensics & \slopeCI{+0.00}{-0.08,0.08} & \slopeCI{-0.04}{-0.17,0.09} & \slopeCI{+0.12}{-0.05,0.29} & \slopeCI{-0.03}{-0.27,0.20} & \slopeCI{-0.07}{-0.16,0.04} & \slopeCI{-0.18}{-0.35,0.01} \\
I3D-DF40-FR & \slopeCI{\mathbf{-0.58}}{-0.69,-0.48} & \slopeCI{\mathbf{-0.31}}{-0.47,-0.14} & \slopeCI{\mathbf{-0.41}}{-0.54,-0.26} & \slopeCI{-0.02}{-0.18,0.17} & \slopeCI{-0.44^{\S}}{-0.54,-0.35} & \slopeCI{+0.04^{\S}}{-0.07,0.17} \\
\rowcolor{tabgray}
I3D-DF40-FS & \slopeCI{\mathbf{-0.37}}{-0.42,-0.31} & \slopeCI{\mathbf{-0.28}}{-0.39,-0.18} & \slopeCI{\mathbf{-0.34}}{-0.46,-0.21} & \slopeCI{\mathbf{-0.24}}{-0.42,-0.04} & \slopeCI{-0.09}{-0.18,0.004} & \slopeCI{-0.04}{-0.19,0.14} \\
I3D-DF40-FSFR & \slopeCI{\mathbf{-0.44}}{-0.50,-0.39} & \slopeCI{\mathbf{-0.10}}{-0.20,-0.01} & \slopeCI{\mathbf{-0.38}}{-0.53,-0.24} & \slopeCI{-0.03}{-0.21,0.15} & \slopeCI{\mathbf{-0.36}}{-0.45,-0.27} & \slopeCI{+0.06}{-0.07,0.20} \\
\rowcolor{tabgray}
FTCN & \slopeCI{+0.08^{\dagger}}{-0.05,0.21} & \slopeCI{-0.13^{\dagger}}{-0.38,0.12} & \slopeCI{-0.06}{-0.16,0.06} & \slopeCI{\mathbf{-0.25}}{-0.44,-0.04} & \slopeCI{\mathbf{-0.14}}{-0.20,-0.08} & \slopeCI{\mathbf{-0.15}}{-0.26,-0.03} \\
TALL & \slopeCI{\mathbf{-0.39}}{-0.47,-0.32} & \slopeCI{\mathbf{-0.13}}{-0.24,-0.03} & \slopeCI{\mathbf{-0.44}}{-0.55,-0.33} & \slopeCI{-0.07}{-0.21,0.09} & \slopeCI{\mathbf{-0.21}}{-0.34,-0.08} & \slopeCI{-0.07}{-0.28,0.13} \\
\rowcolor{tabgray}
FakeSTormer & \slopeCI{+0.01}{-0.05,0.07} & \slopeCI{\mathbf{-0.11}}{-0.22,-0.002} & \slopeCI{-0.00}{-0.14,0.14} & \slopeCI{+0.11}{-0.17,0.36} & \slopeCI{+0.08^{\dagger}}{0.0004,0.16} & \slopeCI{-0.06^{\dagger}}{-0.26,0.11} \\
TFCU & \slopeCI{+0.00}{-0.06,0.06} & \slopeCI{-0.10}{-0.21,0.005} & \slopeCI{-0.01^{\dagger}}{-0.14,0.12} & \slopeCI{-0.09^{\dagger}}{-0.34,0.15} & \slopeCI{+0.02}{-0.10,0.12} & \slopeCI{-0.08}{-0.28,0.11} \\
\rowcolor{tabgray}
DFD-FCG & \slopeCI{\mathbf{-0.56}}{-0.64,-0.49} & \slopeCI{\mathbf{-0.55}}{-0.70,-0.42} & \slopeCI{\mathbf{-0.44}}{-0.56,-0.31} & \slopeCI{\mathbf{-0.51}}{-0.85,-0.19} & \slopeCI{\mathbf{-0.16}}{-0.25,-0.07} & \slopeCI{\mathbf{-0.21}}{-0.41,-0.01} \\
DFD-HR & \slopeCI{\mathbf{-0.13}}{-0.19,-0.07} & \slopeCI{-0.08}{-0.19,0.04} & \slopeCI{\mathbf{-0.17}}{-0.26,-0.07} & \slopeCI{-0.05}{-0.18,0.10} & \slopeCI{-0.02^{\dagger}}{-0.11,0.07} & \slopeCI{-0.07^{\dagger}}{-0.24,0.10} \\
\rowcolor{tabgray}
VLAForge & \slopeCI{\mathbf{-0.18}}{-0.26,-0.11} & \slopeCI{-0.03}{-0.16,0.10} & \slopeCI{-0.25^{\dagger}}{-0.38,-0.12} & \slopeCI{-0.17^{\dagger}}{-0.33,0.01} & \slopeCI{-0.16^{\dagger}}{-0.25,-0.06} & \slopeCI{-0.07^{\dagger}}{-0.21,0.06} \\
GenD-CLIP & \slopeCI{\mathbf{-0.12}}{-0.19,-0.06} & \slopeCI{\mathbf{-0.12}}{-0.25,-0.01} & \slopeCI{-0.08}{-0.22,0.08} & \slopeCI{\mathbf{-0.21}}{-0.41,-0.02} & \slopeCI{-0.06^{\dagger}}{-0.15,0.03} & \slopeCI{-0.00^{\dagger}}{-0.16,0.17} \\
\rowcolor{tabgray}
GenD-PE & \slopeCI{\mathbf{-0.08}}{-0.15,-0.02} & \slopeCI{-0.08}{-0.20,0.03} & \slopeCI{-0.07^{\dagger}}{-0.21,0.10} & \slopeCI{-0.17^{\dagger}}{-0.35,0.02} & \slopeCI{-0.10}{-0.23,0.02} & \slopeCI{-0.09}{-0.28,0.10} \\
GenD-DINOv3 & \slopeCI{\mathbf{-0.18}}{-0.24,-0.13} & \slopeCI{-0.03}{-0.14,0.07} & \slopeCI{-0.13^{\dagger}}{-0.26,0.01} & \slopeCI{-0.01^{\dagger}}{-0.17,0.17} & \slopeCI{+0.05^{\dagger}}{-0.04,0.13} & \slopeCI{+0.02^{\dagger}}{-0.10,0.15} \\
\rowcolor{tabgray}
SBI & \slopeCI{-0.04}{-0.11,0.02} & \slopeCI{\mathbf{-0.18}}{-0.30,-0.06} & \slopeCI{+0.01}{-0.11,0.14} & \slopeCI{+0.07}{-0.16,0.30} & \slopeCI{\mathbf{-0.17}}{-0.27,-0.08} & \slopeCI{\mathbf{-0.20}}{-0.38,-0.04} \\
\midrule
Negative (95\% CI $<0$) & 11/15 & 9/15 & 6/12 & 4/12 & 5/10 & 3/10 \\
Retained & \multicolumn{2}{c}{7/11} & \multicolumn{2}{c}{2/6} & \multicolumn{2}{c}{3/5} \\
\bottomrule
\end{tabular}
\end{table*}

\begin{table*}[!t]
\centering\footnotesize
\setlength{\tabcolsep}{4pt}
\renewcommand{\arraystretch}{1.05}

\caption{\textbf{Mouth dynamics contribute to score-gap
attenuation beyond head motion.}
Incremental attenuation of the within-participant
score gap after adding $\mdyn$ to head speed
(+ Mouth) or head speed to $\mdyn$ (+ Head).
Values are in normal-score units;
brackets report paired 95\% person--session
bootstrap confidence intervals.
Bold indicates intervals entirely above zero.
The main-text counts exclude I3D-DF40-FR
and FTCN on RealTalk development.}
\label{tab:supp-head-gap}
\begin{tabular}{@{}l cccc@{}}
\toprule
 & \multicolumn{2}{c}{RealTalk development} & \multicolumn{2}{c}{Seamless test}\\
\cmidrule(lr){2-3}\cmidrule(l){4-5}
Detector & + Mouth [95\% CI] & + Head [95\% CI] & + Mouth [95\% CI] & + Head [95\% CI]\\
\midrule
RealForensics & $\mathbf{+0.58}\,[+0.52,+0.65]$ & $-0.006\,[-0.024,+0.010]$ & $\mathbf{+0.53}\,[+0.38,+0.67]$ & $-0.008\,[-0.028,+0.010]$ \\
LipForensics & $\mathbf{+0.26}\,[+0.19,+0.32]$ & $-0.007\,[-0.026,+0.012]$ & $\mathbf{+0.25}\,[+0.09,+0.40]$ & $-0.016\,[-0.052,+0.017]$ \\
I3D-DF40-FR & $\mathbf{+0.24}\,[+0.20,+0.29]$ & $+0.029\,[-0.019,+0.078]$ & $\mathbf{+0.30}\,[+0.17,+0.42]$ & $+0.063\,[-0.044,+0.167]$ \\
I3D-DF40-FS & $\mathbf{+0.51}\,[+0.44,+0.58]$ & $+0.002\,[-0.001,+0.008]$ & $\mathbf{+0.59}\,[+0.47,+0.72]$ & $-0.000\,[-0.012,+0.013]$ \\
I3D-DF40-FSFR & $\mathbf{+0.24}\,[+0.18,+0.30]$ & $+0.021\,[-0.014,+0.058]$ & $\mathbf{+0.23}\,[+0.10,+0.37]$ & $+0.047\,[-0.034,+0.126]$ \\
FTCN & $\mathbf{+0.22}\,[+0.16,+0.28]$ & $-0.012\,[-0.046,+0.021]$ & $\mathbf{+0.19}\,[+0.07,+0.31]$ & $-0.026\,[-0.072,+0.023]$ \\
TALL & $\mathbf{+0.48}\,[+0.42,+0.54]$ & $+0.000\,[-0.002,+0.003]$ & $\mathbf{+0.52}\,[+0.41,+0.64]$ & $+0.003\,[-0.006,+0.021]$ \\
FakeSTormer & $\mathbf{+0.14}\,[+0.08,+0.20]$ & $-0.001\,[-0.008,+0.003]$ & $+0.08\,[-0.02,+0.17]$ & $-0.005\,[-0.020,+0.007]$ \\
TFCU & $\mathbf{+0.17}\,[+0.12,+0.23]$ & $-0.005\,[-0.019,+0.008]$ & $\mathbf{+0.19}\,[+0.05,+0.32]$ & $-0.027\,[-0.081,+0.032]$ \\
DFD-FCG & $\mathbf{+0.49}\,[+0.43,+0.55]$ & $+0.001\,[-0.002,+0.006]$ & $\mathbf{+0.43}\,[+0.30,+0.55]$ & $-0.005\,[-0.022,+0.008]$ \\
DFD-HR & $\mathbf{+0.25}\,[+0.21,+0.31]$ & $-0.004\,[-0.015,+0.006]$ & $\mathbf{+0.24}\,[+0.16,+0.34]$ & $-0.018\,[-0.056,+0.021]$ \\
VLAForge & $\mathbf{+0.20}\,[+0.15,+0.25]$ & $-0.001\,[-0.005,+0.002]$ & $\mathbf{+0.28}\,[+0.17,+0.40]$ & $-0.013\,[-0.043,+0.015]$ \\
GenD-CLIP & $\mathbf{+0.25}\,[+0.20,+0.31]$ & $-0.004\,[-0.016,+0.007]$ & $\mathbf{+0.18}\,[+0.06,+0.30]$ & $-0.026\,[-0.078,+0.032]$ \\
GenD-PE & $\mathbf{+0.19}\,[+0.13,+0.24]$ & $-0.004\,[-0.015,+0.006]$ & $\mathbf{+0.26}\,[+0.11,+0.42]$ & $-0.026\,[-0.079,+0.030]$ \\
GenD-DINOv3 & $\mathbf{+0.26}\,[+0.21,+0.31]$ & $-0.001\,[-0.007,+0.003]$ & $\mathbf{+0.22}\,[+0.13,+0.32]$ & $-0.013\,[-0.040,+0.016]$ \\
SBI & $\mathbf{+0.11}\,[+0.06,+0.15]$ & $+0.002\,[-0.003,+0.008]$ & $+0.08\,[-0.004,+0.16]$ & $-0.014\,[-0.046,+0.016]$ \\
\bottomrule
\end{tabular}
\end{table*}

\begin{table*}[!t]
\centering
\footnotesize
\setlength{\tabcolsep}{4pt}
\renewcommand{\arraystretch}{1.08}
\caption{\textbf{Score-gap sensitivity to adjustment models.}
Participant fixed-effects coefficients of the non-speaking indicator on
rank-normalized detector scores. Raw: no mouth-dynamics adjustment;
Linear, Cubic, Decile, and State-specific: alternative adjustments for
$\mdyn$. Raw and Linear report 95\% person--session bootstrap
confidence intervals in brackets; other columns report point estimates.
The state-specific gap is evaluated midway between the speaking and
non-speaking medians of rank-normalized $\mdyn$.
$^{\dagger}$: not evaluable, excluded from summaries
(\cref{supp:protocol}); $^{\S}$: not part of the main-text summaries (\cref{supp:measure}).}
\label{tab:supp-nonlinear}
\begin{tabular}{@{}lccccc@{}}
\toprule
Detector & Raw [95\% CI] & Linear [95\% CI] & Cubic & Decile & State-specific \\
\midrule
\multicolumn{6}{@{}l}{\textit{RealTalk development}} \\
\addlinespace[2pt]
RealForensics & $1.07\, [1.01,1.12]$ & $0.48\, [0.41,0.54]$ & $0.31$ & $0.27$ & $0.46$ \\
LipForensics & $0.46\, [0.40,0.51]$ & $0.30\, [0.23,0.38]$ & $0.20$ & $0.15$ & $0.30$ \\
I3D-DF40-FR & $0.92\, [0.83,1.02]$ & $0.11\, [-0.01,0.25]$ & $0.05$ & $0.04$ & $0.11$ \\
I3D-DF40-FS & $0.92\, [0.87,0.98]$ & $0.21\, [0.15,0.27]$ & $0.11$ & $0.11$ & $0.21$ \\
I3D-DF40-FSFR & $0.71\, [0.66,0.76]$ & $0.05\, [-0.02,0.11]$ & $-0.01$ & $-0.02$ & $0.04$ \\
FTCN & $0.05^{\dagger}\, [-0.07,0.15]$ & $0.14\, [-0.02,0.29]$ & $0.12$ & $0.12$ & $0.14$ \\
TALL & $0.69\, [0.64,0.74]$ & $0.07\, [0.003,0.14]$ & $-0.06$ & $-0.09$ & $0.07$ \\
FakeSTormer & $0.30\, [0.26,0.34]$ & $0.15\, [0.09,0.21]$ & $0.12$ & $0.11$ & $0.15$ \\
TFCU & $0.16\, [0.12,0.20]$ & $0.10\, [0.04,0.16]$ & $0.07$ & $0.06$ & $0.10$ \\
DFD-FCG & $0.78\, [0.73,0.83]$ & $0.12\, [0.05,0.20]$ & $0.01$ & $0.00$ & $0.13$ \\
DFD-HR & $0.35\, [0.31,0.39]$ & $0.14\, [0.08,0.21]$ & $0.08$ & $0.06$ & $0.14$ \\
VLAForge & $0.21\, [0.16,0.25]$ & $-0.02\, [-0.08,0.05]$ & $-0.05$ & $-0.06$ & $-0.01$ \\
GenD-CLIP & $0.31\, [0.26,0.35]$ & $0.12\, [0.06,0.18]$ & $0.07$ & $0.06$ & $0.11$ \\
GenD-PE & $0.21\, [0.17,0.25]$ & $0.07\, [0.01,0.13]$ & $0.03$ & $0.02$ & $0.07$ \\
GenD-DINOv3 & $0.32\, [0.28,0.36]$ & $0.05\, [-0.01,0.11]$ & $-0.03$ & $-0.04$ & $0.05$ \\
SBI & $0.37\, [0.33,0.41]$ & $0.18\, [0.12,0.24]$ & $0.12$ & $0.11$ & $0.18$ \\
\midrule
\multicolumn{6}{@{}l}{\textit{Seamless test}} \\
\addlinespace[2pt]
RealForensics & $1.07\, [0.92,1.21]$ & $0.49\, [0.33,0.68]$ & $0.27$ & $0.14$ & $0.45$ \\
LipForensics & $0.40\, [0.23,0.57]$ & $0.23\, [0.08,0.41]$ & $0.23$ & $0.12$ & $0.20$ \\
I3D-DF40-FR & $1.11\, [0.99,1.25]$ & $0.36\, [0.16,0.54]$ & $0.37$ & $0.16$ & $0.35$ \\
I3D-DF40-FS & $0.94\, [0.75,1.13]$ & $0.23\, [0.04,0.46]$ & $0.19$ & $0.17$ & $0.22$ \\
I3D-DF40-FSFR & $0.81\, [0.68,0.94]$ & $0.23\, [0.003,0.41]$ & $0.16$ & $0.05$ & $0.22$ \\
FTCN & $0.21\, [0.09,0.34]$ & $0.16\, [-0.0001,0.33]$ & $0.18$ & $-0.00$ & $0.17$ \\
TALL & $0.47\, [0.33,0.63]$ & $-0.19\, [-0.35,0.02]$ & $-0.25$ & $-0.50$ & $-0.19$ \\
FakeSTormer & $0.24\, [0.16,0.31]$ & $0.19\, [0.07,0.31]$ & $0.21$ & $0.15$ & $0.19$ \\
TFCU & $-0.08^{\dagger}\, [-0.19,0.02]$ & $-0.08\, [-0.24,0.13]$ & $-0.12$ & $-0.04$ & $-0.08$ \\
DFD-FCG & $0.70\, [0.58,0.82]$ & $0.22\, [0.07,0.38]$ & $0.24$ & $0.26$ & $0.24$ \\
DFD-HR & $0.06\, [-0.02,0.14]$ & $-0.09\, [-0.21,0.05]$ & $-0.10$ & $-0.04$ & $-0.07$ \\
VLAForge & $-0.04^{\dagger}\, [-0.11,0.04]$ & $-0.26\, [-0.39,-0.13]$ & $-0.21$ & $-0.21$ & $-0.25$ \\
GenD-CLIP & $0.03\, [-0.07,0.14]$ & $0.02\, [-0.16,0.21]$ & $-0.01$ & $0.06$ & $0.02$ \\
GenD-PE & $0.06^{\dagger}\, [-0.07,0.19]$ & $-0.03\, [-0.28,0.22]$ & $-0.10$ & $0.06$ & $-0.04$ \\
GenD-DINOv3 & $0.04^{\dagger}\, [-0.06,0.15]$ & $-0.12\, [-0.29,0.07]$ & $-0.10$ & $0.07$ & $-0.12$ \\
SBI & $0.19\, [0.12,0.28]$ & $0.22\, [0.09,0.34]$ & $0.17$ & $0.16$ & $0.22$ \\
\midrule
\multicolumn{6}{@{}l}{\textit{AMI}} \\
\addlinespace[2pt]
RealForensics & $0.88\, [0.73,1.02]$ & $0.53\, [0.42,0.66]$ & $0.46$ & $0.48$ & $0.47$ \\
LipForensics & $0.60\, [0.47,0.73]$ & $0.41\, [0.27,0.55]$ & $0.34$ & $0.33$ & $0.41$ \\
I3D-DF40-FR & $0.84^{\S}\, [0.72,0.98]$ & $0.25^{\S}\, [0.12,0.40]$ & $0.24^{\S}$ & $0.20^{\S}$ & $0.26^{\S}$ \\
I3D-DF40-FS & $0.59\, [0.45,0.73]$ & $0.23\, [0.10,0.37]$ & $0.20$ & $0.26$ & $0.22$ \\
I3D-DF40-FSFR & $0.60\, [0.51,0.70]$ & $0.13\, [-0.02,0.28]$ & $0.17$ & $0.14$ & $0.13$ \\
FTCN & $0.22\, [0.14,0.30]$ & $0.13\, [0.02,0.23]$ & $0.11$ & $0.12$ & $0.14$ \\
TALL & $0.61\, [0.52,0.69]$ & $0.28\, [0.15,0.45]$ & $0.20$ & $0.18$ & $0.28$ \\
FakeSTormer & $-0.17^{\dagger}\, [-0.26,-0.09]$ & $-0.12\, [-0.22,-0.03]$ & $-0.12$ & $-0.11$ & $-0.13$ \\
TFCU & $0.22\, [0.12,0.31]$ & $0.18\, [0.09,0.28]$ & $0.14$ & $0.13$ & $0.18$ \\
DFD-FCG & $0.43\, [0.31,0.54]$ & $0.24\, [0.09,0.37]$ & $0.16$ & $0.16$ & $0.24$ \\
DFD-HR & $0.10^{\dagger}\, [0.03,0.18]$ & $0.03\, [-0.07,0.13]$ & $-0.00$ & $0.03$ & $0.03$ \\
VLAForge & $0.10^{\dagger}\, [0.03,0.17]$ & $-0.05\, [-0.17,0.05]$ & $-0.05$ & $0.00$ & $-0.05$ \\
GenD-CLIP & $0.09^{\dagger}\, [-0.01,0.19]$ & $-0.02\, [-0.14,0.09]$ & $-0.04$ & $-0.02$ & $-0.02$ \\
GenD-PE & $0.19\, [0.13,0.26]$ & $0.09\, [-0.03,0.21]$ & $0.08$ & $0.10$ & $0.09$ \\
GenD-DINOv3 & $0.04^{\dagger}\, [-0.03,0.11]$ & $0.12\, [0.02,0.22]$ & $0.08$ & $0.09$ & $0.12$ \\
SBI & $0.23\, [0.16,0.31]$ & $0.05\, [-0.06,0.16]$ & $0.01$ & $0.05$ & $0.06$ \\
\bottomrule
\end{tabular}
\end{table*}

\begin{table*}[!t]
\centering
\footnotesize
\setlength{\tabcolsep}{8pt}
\renewcommand{\arraystretch}{1.10}
\caption{\textbf{Score-gap estimates under limited overlap in mouth dynamics.}
Unadjusted (Raw) and $\mdyn$-adjusted (Adj.)
participant fixed-effects coefficients on rank-normalized
detector scores (point estimates).
For each detector--dataset pair, the overlap interval
extends from the speaking 5th percentile to the
non-speaking 95th percentile of $\mdyn$.
$n$ denotes the number of participants retaining both states.
Estimates require $n\geq30$ and more than 50 windows;
--- indicates insufficient support.
No Seamless pair meets these requirements.}
\label{tab:supp-common-support}
\begin{tabular}{@{}lrrrrrr@{}}
\toprule
 & \multicolumn{3}{c}{RealTalk}
 & \multicolumn{3}{c}{AMI} \\
\cmidrule(lr){2-4}
\cmidrule(l){5-7}
Detector & Raw & Adj. & $n$ & Raw & Adj. & $n$ \\
\midrule
RealForensics & 0.36 & 0.28 & 124 & 0.33 & 0.25 & 46 \\
LipForensics & 0.29 & 0.20 & 124 & 0.38 & 0.40 & 46 \\
I3D-DF40-FR & $-0.12$ & $-0.23$ & 39 & \multicolumn{2}{c}{---} & 25 \\
I3D-DF40-FS & 0.11 & $-0.03$ & 113 & \multicolumn{2}{c}{---} & 25 \\
I3D-DF40-FSFR & $-0.03$ & $-0.12$ & 113 & \multicolumn{2}{c}{---} & 25 \\
FTCN & 0.18 & 0.27 & 42 & 0.35 & 0.33 & 46 \\
TALL & $-0.09$ & $-0.18$ & 122 & 0.28 & 0.29 & 32 \\
FakeSTormer & 0.11 & 0.09 & 124 & 0.05 & 0.03 & 46 \\
TFCU & 0.09 & 0.14 & 124 & 0.11 & 0.07 & 46 \\
DFD-FCG & $-0.02$ & $-0.15$ & 124 & 0.26 & 0.31 & 46 \\
DFD-HR & 0.23 & 0.17 & 124 & $-0.05$ & $-0.02$ & 46 \\
VLAForge & 0.00 & $-0.05$ & 124 & $-0.06$ & $-0.06$ & 46 \\
GenD-CLIP & 0.03 & $-0.03$ & 124 & 0.02 & 0.02 & 46 \\
GenD-PE & 0.15 & 0.13 & 124 & 0.25 & 0.32 & 46 \\
GenD-DINOv3 & 0.04 & 0.02 & 124 & 0.07 & 0.08 & 46 \\
SBI & 0.09 & 0.07 & 124 & 0.12 & 0.12 & 46 \\
\bottomrule
\end{tabular}
\end{table*}

\subsection{Within-State Associations and Their Specificity}
\label{supp:within}

We examine whether detector scores track mouth dynamics
even within the same conversational state.
We regress scores on $\mdyn$ separately within speaking
and non-speaking windows, with participant fixed effects.
Negative non-speaking slopes are statistically supported
for 11/15 detectors on RealTalk, 6/12 on Seamless,
and 5/10 on AMI; the corresponding speaking counts
are 15/15, 6/12, and 6/10.
The RealTalk pattern persists under alternative
speech-label definitions (\cref{supp:windows}).

DFD-FCG, TALL, and I3D-DF40-FSFR show negative
unadjusted associations across all three datasets,
but differ in their sensitivity to potential confounders.
Head-speed adjustment preserves this pattern for
DFD-FCG and TALL, while substantially attenuating
the associations for the I3D models.
After jointly adjusting for head speed, landmark
instability, sharpness, and encoding, only DFD-FCG
retains a statistically supported negative association
on all three datasets (\cref{tab:supp-slopes-controls}).
TALL retains support only on RealTalk after
full adjustment and is particularly sensitive
to landmark instability.

Further controls reinforce this distinction.
For DFD-FCG, negative associations persist on
RealTalk development and Seamless test when using
tracked-point or phase-based lip-motion descriptors
while controlling for lip-landmark instability.
They also persist when adjusting for the log motion
and log instability of rigid eye/nose points:
the $\mdyn$ slopes are $-0.63$ [$-0.73$, $-0.52$]
and $-0.46$ [$-0.61$, $-0.32$], respectively.
Yaw/pitch and static-appearance adjustments preserve
the negative association on pooled RealTalk and
Seamless test, although the static-adjusted AMI
estimate is not statistically supported
($-0.106$ [$-0.227$, $0.014$]).

For TALL, the negative association persists under
rigid-point and yaw/pitch adjustments, but is not
statistically supported with tracked-point or
phase-based lip-motion descriptors.
Static-appearance adjustment yields a positive
association on RealTalk and an unsupported
estimate on Seamless.
For the I3D models, rigid-point motion or landmark
instability alone accounts for at least half the
magnitude of the unadjusted $\mdyn$ slope, limiting
the separation of mouth and head motion.

These controls provide the strongest evidence for
a mouth-dynamics-specific association in DFD-FCG.
For some detectors, the association between mouth dynamics and fake scores weakens after controlling for head motion and landmark instability. However, these factors are correlated, preventing a clear attribution of the observed effect to any single factor.
The controlled lip-motion interventions in
\cref{supp:liveportrait} provide complementary
evidence by directly modifying mouth dynamics
for DFD-FCG and TALL.

\subsection{Statistical Attenuation of the Score Gap}
\label{supp:gap}

We examine how much of the within-participant
score gap between speaking and non-speaking windows
is statistically associated with $\mdyn$.
Adding $\mdyn$ to the score-gap model
(\cref{sec:metric,supp:protocol}) attenuates
the non-speaking coefficient by a median of
67\% on RealTalk, 61\% on Seamless, and 49\% on AMI.
When $\mdyn$ is permuted within participant and state,
preserving the between-state contrast but disrupting
its window-level association with scores, median
attenuation falls to 28\%, 22\%, and 10\%,
respectively.
Thus, the observed attenuation is not attributable
solely to differences in mouth dynamics between states.

The contribution of $\mdyn$ also extends beyond
head motion.
Adding it after head speed yields further statistically
supported attenuation in 28 of 30 main-text
comparisons, and 30 of 32 when including two
additional RealTalk-development fits.
Conversely, adding head speed after $\mdyn$
produces no statistically supported positive
increment (\cref{tab:supp-head-gap}).
The two unsupported mouth increments occur for
FakeSTormer and SBI on Seamless.

The remaining score gap depends on the adjustment
model. Cubic and decile-indicator models often yield
smaller signed coefficients than the linear model
(\cref{tab:supp-nonlinear}). However, a numerically
smaller coefficient does not always mean a gap
closer to zero: the estimated gap can even reverse
direction, as for TALL on Seamless.

Interpretation is further limited by the small overlap
in $\mdyn$ between states.
The common range contains only 19\%, 2\%, and 15\%
of windows on RealTalk, Seamless, and AMI, respectively.
Within this range, adjusted gaps remain positive for
RealForensics and LipForensics, whereas DFD-FCG and
TALL show negative gaps on RealTalk and positive
gaps on AMI (\cref{tab:supp-common-support}).

Together, these results establish a substantial
statistical association between mouth dynamics
and the non-speaking score gap, but not a causal
contribution of the same magnitude.
The attenuation shares are model-dependent summaries,
not measured reductions in false positives.
The controlled interventions in
\cref{supp:liveportrait} complement these analyses
by directly modifying lip motion.

\subsection{Analysis Details}
\label{supp:follows-details}

\paragraph{Regional comparison.}
Let $Y_i$ indicate a false positive at FF95,
$L_i$ indicate non-speaking, and $\mathbf{x}_i^G$
denote the features belonging to groups $G$.
Features are rank-normalized within each dataset,
and variables are centered within participants:
$\tilde z_i=z_i-\bar z_{p(i)}$.
We use five-fold cross-fitting, keeping entire
person--session components within the same fold.
For each feature subset $G$, least squares without
an intercept predicts $\tilde L$ from
$\tilde{\mathbf{x}}^G$ using four folds and
produces out-of-fold predictions $\widehat L_i^G$
on the held-out fold.
We estimate the adjusted non-speaking coefficient
and the share of the original gap accounted for as
\begin{equation}
\widehat\beta_G=
\frac{\sum_i(\tilde L_i-\widehat L_i^G)\tilde Y_i}
     {\sum_i(\tilde L_i-\widehat L_i^G)\tilde L_i},
\qquad
\rho_G=1-\frac{\widehat\beta_G}{\widehat\beta_0},
\end{equation}
where
$\widehat\beta_0=
\sum_i\tilde L_i\tilde Y_i/\sum_i\tilde L_i^2$.
Only the non-speaking indicator is residualized.
We obtain confidence intervals from 1{,}000
person--session component bootstrap resamples,
holding the out-of-fold predictions fixed and
using paired differences for incremental contributions.
Shares and increments are not constrained to
$[0,100]\%$: adjustment can increase the residual
coefficient or reverse its sign.

The analysis includes 3{,}696 RealTalk-development
windows from 646 participants and 1{,}055
Seamless-test windows from 96 participants,
before detector-specific missing scores.
FTCN and I3D-DF40-FR are unavailable in the
RealTalk feature table, while AMI lacks the
complete feature set.
Of 30 candidate detector--dataset combinations,
16 satisfy the eligibility criteria: at least
30 false positives, 30 non-false-positives,
and a strictly positive lower 95\% confidence
bound for the full model's out-of-fold predictive
$R^2$.
Changing the minimum count to 20 or 50
selects the same combinations.
FF95 calibration and evaluation use consistent
score fields, including TFCU's released probability.

\paragraph{Predictive contribution of feature groups.}
In a separate analysis, least squares predicts
$\tilde Y$ from the centered features using
the same cross-validation folds.
Mouth features achieve the highest individual-group
out-of-fold $R^2$ in 12 of 16 eligible combinations.
However, their incremental contribution exceeds
that of the other five groups combined with a
paired confidence interval above zero in only
4 of 16 combinations.
Thus, the mouth's contribution to the non-speaking
false-positive gap does not imply that mouth
features dominate the prediction of false positives
more generally.

\paragraph{Regional features.}
We extract 43 features in six predefined groups.
The mouth group (10) includes $\plip$ (\cref{supp:plip}),
FAN aperture variation~\cite{bulat2017fan},
the mean and standard deviation of jaw opening,
mean mouth closure and smile, three pixel-based
motion energies, and one learned-feature energy.
The head group (7) comprises the means and standard
deviations of yaw, pitch, and roll, together with
head speed.
Eyes and gaze (10) include the means and standard
deviations of horizontal and vertical gaze,
mean blink and squint, blink rate, two pixel
energies, and one learned-feature energy.
Upper-face expression (3) includes inner-brow
raise, brow lowering, and cheek squint.
Other facial regions (9) contribute two pixel
energies and one learned-feature energy each
for the nose, chin, and remaining face.
The chin is assigned to this group rather than
to the mouth group.
Image quality (4) includes face size, sharpness,
luminance, and blockiness.

\emph{Facial geometry and expression.}
MediaPipe FaceLandmarker~\cite{kartynnik2019facemesh}
in video mode provides pose, gaze, and blendshapes
on a fixed square crop.
Horizontal and vertical gaze are computed as
\[
g^h=\tfrac12(o_L+i_R-i_L-o_R),\qquad
g^v=\tfrac12(u_L+u_R-d_L-d_R),
\]
where $o$, $i$, $u$, and $d$ denote look-out,
look-in, look-up, and look-down blendshapes.
Blink rate counts upward crossings per second
of the mean blink blendshape above its
window median plus 0.2.
FAN aperture variation is defined as
\[
\operatorname{median}_t
\left(\frac{|\ell_{t+8}-\ell_t|}{d_t}\right),
\]
where $\ell_t$ is the inner-lip distance
(FAN landmarks 62 and 66) and $d_t$ is the
interocular distance (landmarks 36 and 45).

\emph{Regional motion energies.}
Pixel energies are computed from the ten aligned
$224\times224$ luminance frames used by DFD-FCG:
\[
P_k(\mathbf u)=\frac1{10}
\left|
\sum_{t=0}^{9}
Y_t(\mathbf u)e^{-2\pi i kt/10}
\right|^2.
\]
Let $\bar P_{k,R}$ denote the spatial mean
of $P_k$ over region $R$.
We compute
$\log\sum_{k=1}^{2}\bar P_{k,R}$ and
$\log\sum_{k=3}^{5}\bar P_{k,R}$
for each region, with the additional
$\log\sum_{k=1}^{5}\bar P_{k,R}$
for the mouth.
Regions are defined on a $16\times16$ grid
of $14\times14$-pixel patches:
rows/columns 10--12/5--10 for the mouth,
5--7/3--12 for the eyes,
8--9/5--10 for the nose,
13--14/4--11 for the chin,
and the remaining patches for the rest of the face.

Learned-feature energy is computed from
$\ell_2$-normalized layer-24 DINOv2-large
tokens extracted from the same ten frames:
\[
V_R=
\operatorname{mean}_{p\in R}
\left[
1-
\frac{
\sum_{t\ne s}
\mathbf h_{t,p}^{\top}\mathbf h_{s,p}
}{
T(T-1)
}
\right],
\qquad T=10.
\]

\emph{Head motion and image quality.}
Head speed is defined in \cref{supp:head}.
Image-quality features are sampled every fourth
frame and comprise the log median face-box side,
log median face-crop Laplacian variance,
mean luminance, and the ratio of mean absolute
luminance differences across versus within
8-pixel block boundaries.

\paragraph{Descriptor comparison.}

All descriptors are evaluated on the same 3{,}685 RealTalk-development, 5{,}240 Seamless-development, and 1{,}078 Seamless-test windows, including editing cuts.
Unlike the main evaluation, RealTalk windows are further restricted by TalkNet score as in  \cref{supp:windows}.
This difference in samples and labels explains why the separation results differ from those reported in the main evaluation.

We evaluate speaking--non-speaking separation using participant-balanced within-participant AUC. 
Split-half reliability is the Spearman correlation between descriptor values computed on the two halves of non-speaking windows.
Head-motion dependence is measured by the Spearman correlation with head speed on the same windows; this captures genuine co-movement as well as potential measurement artifacts.

To test sensitivity to controlled lip motion, we impose a 4\,Hz lower-lip oscillation with peak amplitude 0.01\,IOD on 150 non-speaking Seamless-development clips.
Each edited clip is compared with an identically re-encoded zero-amplitude sham, yielding 149 valid
pairs for $\mdyn$.
FAN aperture variation lacks a compatible controlled-motion extraction, while static mouth opening and pixel energy lack comparable estimates for these validation checks.

Finally, predictive recovery is defined as the ratio of the single-descriptor predictive $R^2$ to that of the full ten-feature mouth group.
We report medians across 10 RealTalk-development and 4 Seamless-test detector comparisons with mouth-group $R^2\geq0.02$, using 3{,}681 and 1{,}055 complete-feature windows, respectively.

All entries are invariant to monotone transformations of a descriptor, such as the
logarithm: separation, reliability, and head-motion dependence are rank statistics, the controlled-motion test counts increases, and recovery uses descriptors rank-normalized and centered within participants, as in the regional comparison.

\paragraph{Alternative lip-area summaries.}
\label{supp:choice}
We further compare $\mdyn$ with alternative summaries of lip-aperture variation: the 5th--95th percentile range, standard deviation, and spectral power in the 1--8, 1--10, and 2--7\,Hz bands.
Range and standard deviation account for more within-state score variation on RealTalk development.
However, on the 290-participant RealTalk evaluation split, the median within-state $R^2$ of $\mdyn$ across 14 detectors is only 0.010 [95\% CI 0.003, 0.036], and the comparisons do not establish a consistently superior descriptor.
Here, $R^2$ is the squared correlation between rank-normalized descriptors and untransformed detector scores, both centered within participant and conversational state using the benchmark labels.

\paragraph{Within-state and score-gap analyses.}
We follow the participant fixed-effects models, rank normalization, and component bootstrap described in \cref{supp:protocol}, with editing cuts handled as in \cref{supp:measure}.
The analyses include up to 917, 96, and 123 participants on RealTalk, Seamless, and AMI, respectively.
I3D-DF40-FR and FTCN are evaluated only on the RealTalk evaluation split (285 of 290 participants).
Summary counts and median gap-attenuation shares include the 15, 12, and 10 evaluable detectors of \cref{supp:protocol}.
Tables retain all available estimates. As in the main text, counts include cells with face-crop coverage below 90\% (\cref{supp:ci}) but not I3D-DF40-FR on AMI, whose effect on the summaries is reported in \cref{supp:measure}.

Raw and quality-adjusted slopes are estimated on identical complete-case windows.
The adjustment includes landmark instability measured from three augmented crops, framewise head speed, face and mouth sharpness, clip size in bytes, and quantization parameter (QP).
It additionally includes source-video size on RealTalk and Seamless, and source-video QP on Seamless.
The incremental comparison with head motion instead uses the video-mode head speed (\cref{supp:head}), on RealTalk development and Seamless test.

Additional controls examine the specificity of the association with mouth dynamics. Static-appearance adjustment includes mean jaw opening, mouth closure, smile, and inner-lip area, together with head speed; yaw/pitch adjustment
uses their within-window standard deviations.
Both analyses use pooled RealTalk, Seamless test,
and AMI.
Non-lip controls include log motion and log
instability of rigid eye/nose points and are
evaluated, together with alternative lip-motion
descriptors, on RealTalk development and
Seamless test.

\section{Changing the Mouth Changes the Score}
\label{supp:liveportrait}

\Cref{sec:changing-mouth} shows that increasing
lip motion in manipulated videos can substantially
reduce detection, whereas static mouth opening
alone does not reproduce this effect.
Here, we isolate the contribution of mouth dynamics
from rendering, static appearance, and encoding
effects, examine how detector responses differ,
and test whether the effect persists across
animation methods.

\subsection{Detection under Controlled Lip Animations}

\paragraph{Lip motion lowers detection.}
Using LivePortrait~\cite{guo2024liveportrait}, we generate still-mouth and doubled-motion renders for 5,227 of the 5,240 Celeb-DF++ test fakes~\cite{li2025celebdfpp}.
Still mouths improve detection relative to the original videos, whereas doubled motion reduces detection relative to still mouths across all manipulation families (\cref{tab:supp-allfakes}).
Under otherwise identical rendering conditions, detection drops by 43.2 [37.9, 48.3] percentage points for DFD-FCG and 43.3 [40.0, 47.1] for TALL when moving from still mouths to doubled motion.
The doubled-motion condition remains within the observed range of real speech: among 5,216 clips with measured $\mdyn$, 95\% fall below the 99th percentile of RealTalk speaking windows, and all fall below its maximum.

\paragraph{Construction and evaluation.}
LivePortrait re-renders the lip region of a 3\,s window from each fake while preserving head pose, eye motion, and mean lip shape.
The driving video is a different real video of the same identity, excluding the source videos used to construct the fake.
Eligible driver windows must exhibit at least the median motion observed in real videos.
Among these, we select the window whose log motion is closest to a target inner-lip aperture change of 0.045 interocular distances, measured as the median absolute change over eight frames at 30\,fps (real-video median: 0.034).
In absolute mode, the driver's lip-keypoint deviations from their temporal mean are scaled by $\alpha$ and added to the fake's mean lip shape.
Setting $\alpha=0,1,2$ produces still-mouth, natural-amplitude, and doubled-amplitude renders, respectively, while keeping the driver, rendering pipeline, and x264 CRF~18 encoding unchanged.
Detection uses FF95 (\cref{supp:recal}).
We compute 95\% confidence intervals for the full-set and 400-fake comparisons using 2,000 bootstrap resamples clustered by the 59 identities, preserving the pairing across conditions.
To assess rendering variability, we regenerate six still-mouth clips with identical settings.
The median absolute logit differences are 0.088 for DFD-FCG and 0.183 for TALL, compared with median still-to-doubled logit shifts of $-0.62$ and $-2.11$, respectively, on the full complete-case sets.

\begin{table}[t]
\centering\footnotesize
\setlength{\tabcolsep}{3pt}
\renewcommand{\arraystretch}{1.05}
\caption{\textbf{Lip-motion edits on Celeb-DF++ fakes.} Detection at FF95 (\%) on clips scored in all three arms: 5,193 for DFD-FCG and 5,222 for TALL, out of 5,227 rendered. Motion: doubled minus still, in percentage points, with 95\% identity-bootstrap intervals.}
\label{tab:supp-allfakes}
\begin{tabular}{@{}llrrrrr@{}}\toprule
& Fakes & $n$ & Orig. & Still & $\times 2$ & Motion [95\% CI] \\\midrule
\multirow{4}{*}{\rotatebox{90}{DFD-FCG}}
& All & 5,193 & 77.8 & 93.5 & 50.3 & $-43.2$ [$-48.3$, $-37.9$] \\
& Face swap & 1,708 & 73.7 & 91.6 & 45.1 & $-46.5$ [$-52.5$, $-40.1$] \\
& Reenact. & 1,385 & 76.6 & 92.4 & 45.6 & $-46.8$ [$-54.4$, $-38.8$] \\
& Talking face & 2,100 & 82.0 & 95.9 & 57.7 & $-38.1$ [$-43.1$, $-33.3$] \\\midrule
\multirow{4}{*}{\rotatebox{90}{TALL}}
& All & 5,222 & 49.8 & 72.3 & 29.0 & $-43.3$ [$-47.1$, $-40.0$] \\
& Face swap & 1,740 & 50.6 & 72.1 & 31.4 & $-40.7$ [$-45.3$, $-36.1$] \\
& Reenact. & 1,382 & 53.0 & 72.6 & 25.8 & $-46.8$ [$-52.3$, $-41.1$] \\
& Talking face & 2,100 & 47.1 & 72.3 & 29.1 & $-43.2$ [$-47.6$, $-38.9$] \\\bottomrule
\end{tabular}\end{table}

\subsection{Rendering, Opening, and Encoding Controls}

Comparisons between original and rendered videos may also reflect effects introduced by the rendering process.
To disentangle these effects from changes in mean mouth opening and encoding, we analyze a subset of 400 fakes with complete scores for both detectors, sampled across manipulation families and generation batches from all 59 identities.
\Cref{tab:supp-gen-controls} reports the results for all control conditions on this common subset.

\begin{table}[t]
\centering
\footnotesize
\setlength{\tabcolsep}{4pt}
\renewcommand{\arraystretch}{1.08}
\caption{\textbf{Effects of rendering, mouth opening, and encoding.}
FF95 detection (\%) on the same 400 Celeb-DF++
fakes from 59 identities, with 95\% identity-bootstrap
confidence intervals.
All LivePortrait edits use CRF~18 unless marked
lossless (CRF~0); original videos are unedited.}
\label{tab:supp-gen-controls}
\begin{tabular}{@{}lrr@{}}
\toprule
Arm & DFD-FCG & TALL \\
\midrule
\multicolumn{3}{@{}l}{\textit{Original and rendered edits}} \\
\addlinespace[2pt]
Original
& 79.0 [73.3, 84.4]
& 47.0 [41.1, 52.8] \\
Still ($\alpha=0$)
& 94.5 [92.0, 96.8]
& 71.5 [66.2, 76.5] \\
Natural ($\alpha=1$)
& 82.5 [77.1, 87.6]
& 52.0 [45.6, 58.3] \\
Still-open
& 92.2 [89.0, 95.0]
& 65.0 [60.1, 70.0] \\
Doubled ($\alpha=2$)
& 52.5 [44.8, 59.9]
& 28.0 [22.5, 33.5] \\
\midrule
\multicolumn{3}{@{}l}{\textit{Lossless encoding control}} \\
\addlinespace[2pt]
Still, lossless
& 94.5 [92.1, 96.6]
& 69.5 [64.2, 74.5] \\
Doubled, lossless
& 63.2 [56.2, 70.0]
& 24.8 [19.6, 30.3] \\
\bottomrule
\end{tabular}
\end{table}

\paragraph{Natural amplitude.}
Natural-amplitude renders exhibit similar median $\mdyn$ to the original videos ($-1.95$ versus $-1.92$), with lip motion inherited from the driver.
Relative to the originals, detection rates change by $+3.5$ [$-2.1$, $8.6$] percentage points for DFD-FCG and $+5.0$ [$0.0$, $9.7$] for TALL.
These differences are substantially smaller than the still-to-doubled drops of 42.0 and 43.5 percentage points, respectively, measured on the same clips.

\paragraph{Static opening.}
The still-open condition freezes the lip shape at the frame of the doubled-motion render whose opening is closest to the clip's temporal mean.
The median time-averaged inner-lip area across clips, normalized by squared interocular distance, is 0.0332 for still-open, 0.0326 for doubled motion, and 0.0261 for still mouths.
Despite the closely matched mean opening, the still-open condition reproduces only 5\% [0, 11] of the still-to-doubled detection drop for DFD-FCG and 15\% [4, 24] for TALL.
Thus, mean mouth opening explains only a small fraction of the effect, although this control does not isolate all changes in mouth appearance.

\paragraph{Encoding.}
Increased lip motion reduces detection under both lossless and CRF~18 encoding.
For DFD-FCG, detection drops by 31.2 [24.7, 38.1] percentage points under lossless encoding, compared with 42.0 [34.5, 49.9] under CRF~18, yielding a ratio of 0.74 [0.65, 0.83].
For TALL, the corresponding drops are 44.7 [38.7, 50.8] and 43.5 [37.2, 49.6] percentage points, with a ratio of 1.03 [0.97, 1.10].
Compression therefore amplifies the effect for DFD-FCG, while the TALL contrast remains similar across encoding conditions.

\subsection{Responses across Detectors}

We evaluate the still, still-open, and doubled-motion conditions using twelve additional detectors, comparing changes in mean $P(\mathrm{fake})$ across all 14 detectors.
Among temporal detectors, DFD-FCG, TALL, LipForensics, and I3D-DF40-FS show substantial decreases under the dynamic contrast (13.7--38.9 percentage points), with much smaller changes under static opening (at most 3.5 points).
FTCN responds in the opposite direction, whereas FakeSTormer and TFCU show comparatively small changes.
RealForensics has a mean $P(\mathrm{fake})$ of 99.8\% in the still condition, limiting the sensitivity of this probability-scale comparison.
I3D-DF40-FR and I3D-DF40-FSFR are excluded from this evaluation.

All six frame-based detectors show changes of at most 3.6 percentage points under either contrast.
Because these detectors process individual frames, their responses can arise only from changes in frame appearance rather than temporal motion itself.

Despite responding weakly to these edits, several
frame-based detectors exhibit a non-speaking penalty
on real conversational videos. Thus, observing the
penalty does not imply that every detector responds
strongly to controlled changes in lip motion.

\subsection{Authentic False Positives}

On the Seamless test set, we identify non-speaking false positives from DFD-FCG, TALL, and LipForensics at FF95 using the original 4\,s windows.
For each false positive, we select a driving window from the same participant and session, with speaking motion close to the corpus median.
All conditions are shortened to 3\,s and re-encoded, with 4K inputs resized to a maximum dimension of 1,920 pixels.
Recovery rates are reported descriptively, and the unedited 3\,s control provides a reference for recovery associated with preprocessing alone.

Under doubled motion, 72.8\% of DFD-FCG's 169 false positives (49 participants) and 31.8\% of TALL's 280 false positives (58 participants) fall below the detection threshold.
For DFD-FCG and TALL, respectively, recovery rates are 24.3\% and 2.5\% in the unedited 3\,s controls, 3.0\% and 2.5\% with still mouths, and 34.9\% and 7.9\% at natural amplitude.
The effect also extends to DFD-FCG's speaking false positives: among 24 windows from 13 participants, 79\% recover under doubled motion, compared with 33\% in the unedited control and none with still mouths.
For LipForensics (131 false positives, 47 participants), recovery reaches 20.6\% under doubled motion, compared with 18.3\% in the unedited control, 0.8\% with still mouths, and 2.3\% at natural amplitude.

\subsection{Replication with Other Animation Methods}

We test whether the direction of the response generalizes beyond LivePortrait using two additional animation methods.
MuseTalk~\cite{musetalk} modifies 188 face-swap and reenactment fakes from 35 identities at 30\,fps, selected for low lip motion with FAN aperture variation at most 0.02392 (\cref{supp:follows-details}).
Across motion amplitudes, DFD-FCG's logit decreases by 0.176 [0.143, 0.216] per unit increase in $\mdyn$, compared with 0.304 [0.271, 0.341] for LivePortrait on the same 188 clips.

As a non-neural alternative, we use a piecewise-affine warp that rescales each clip's own lip motion.
After removing head motion with a similarity transform fitted to rigid facial points, the warp moves 49 lip and jaw landmarks from their position $p_t$ in frame $t$ to $m+\alpha(p_t-m)$, where $m$ is their median position in the clip.
The rest of the face contour, the nose, and the eye corners stay fixed.
Warping alone raises DFD-FCG and TALL logits, even without added motion, so we compare $\alpha=2$ with $\alpha=0$ rather than with the unwarped clip ($\alpha=1$).
These two arms deform the same region, and in every frame they move each landmark by the same distance, $\lVert p_t-m\rVert$, in opposite directions; only the resulting lip motion differs.
This contrast is applied to 295 clips from 22 generators and the same 59 identities, with no overlap with the MuseTalk sample.
For the available scores (293 clips for DFD-FCG and 295 for TALL), the logit slopes are $-0.177$ [$-0.204$, $-0.150$] and $-0.796$ [$-0.886$, $-0.711$] per unit increase in $\mdyn$, respectively.
For each method, slopes are estimated by regressing within-clip logit changes against changes in $\mdyn$ relative to the still-mouth condition, with the regression constrained through the origin.
We compute 95\% confidence intervals using 2,000 identity-level bootstrap resamples.

DFD-FCG consistently shows lower logits with increased lip motion across all three methods, although effect sizes vary by method.
These experiments are limited to synthetic lip motion on Celeb-DF++ identities and visual detectors: the edits may open the lips without reconstructing the mouth interior and do not evaluate audio-visual consistency.

\section{Mouth-Dynamics Regularities in FF++}
\label{supp:ffpp}

In FF++, manipulated videos exhibit reduced measured mouth dynamics, while authentic low-motion behavior is underrepresented in the training data (\cref{sec:ff_regularities}). Whether this imbalance provides a potential shortcut for detection depends in part on its robustness to measurement choices and recording conditions.

\subsection{Manipulation Labels and Low-Motion Real Clips}

\paragraph{Sampling and uncertainty.}
We measure the first three non-overlapping 3\,s clips from each FF++ training video at its native frame rate, retaining clips with valid facial landmarks in at least 90\% of frames (\cref{supp:measure}).
The resulting sample contains 10,709 clips: 2,142 real and 8,567 fake.
For real--fake comparisons, we compute 95\% confidence intervals by resampling the 360 connected identity components, preserving dependencies among source, target, and manipulated videos.

\begin{figure}[t]
\centering
\includegraphics[width=\linewidth]{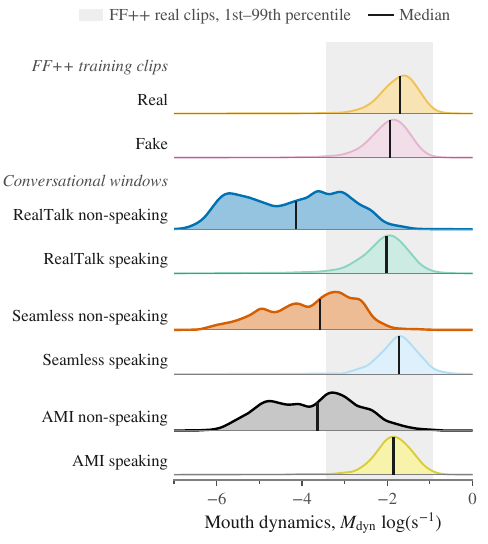}
\caption{\textbf{Non-speaking medians lie below the 1st percentile of FF++ real training clips.}
Each row is a kernel density of $\mdyn$ (bandwidth 0.15 log units) scaled to its own peak, so row heights do not reflect group sizes.
FF++: 3\,s training clips, 2{,}142 real and 8{,}567 fake (DF, F2F, NT, and FS pooled).
Conversational corpora: 4\,s windows with valid $\mdyn$, non-speaking and speaking, respectively: RealTalk (development and held-out pooled), 2{,}395 and 2{,}337; Seamless test, 491 and 531; AMI, 1{,}211 and 1{,}187.
RealTalk development and Seamless exclude windows flagged for temporal discontinuities (\cref{supp:measure}).}
\label{fig:supp-ffpp}
\end{figure}

\paragraph{Reduced dynamics in manipulated clips.}
The median $\mdyn$ is 0.234 [0.215, 0.252] lower in fake than in real clips (\cref{fig:supp-ffpp}), corresponding to an approximately 0.79 ratio in the underlying aperture-change rate.
Both the difference between distribution medians and paired real--target comparisons indicate larger reductions for DF, F2F, and NT than for FS (\cref{tab:supp-ffpp-motion}).
This ordering persists at 25 and 30\,fps, which together account for 97\% of clips.
For FS, the small difference excludes zero at 25\,fps (0.034 [0.005, 0.064]) but not at 30\,fps (0.018 [$-0.011$, 0.042]).

\begin{table}[t]
\centering\footnotesize
\setlength{\tabcolsep}{4pt}
\renewcommand{\arraystretch}{1.05}
\caption{\textbf{Reduced mouth dynamics by manipulation method.} Real-minus-fake differences in $\mdyn$, reported as differences between distribution medians or as median paired differences for matched real--target temporal blocks. Paired differences are expressed in units of the real-clip standard deviation (0.492, fixed during resampling). Brackets indicate 95\% confidence intervals from resampling the 360 identity components.}
\label{tab:supp-ffpp-motion}
\begin{tabular}{@{}lcc@{}}\toprule
Method & Median difference & Paired difference (SD) \\\midrule
DF & 0.394 [0.361, 0.424] & 0.70 [0.65, 0.75] \\
F2F & 0.256 [0.226, 0.279] & 0.47 [0.43, 0.51] \\
NT & 0.284 [0.261, 0.308] & 0.49 [0.46, 0.53] \\
FS & 0.021 [0.001, 0.041] & 0.06 [0.04, 0.08] \\\bottomrule
\end{tabular}
\end{table}

Because FF++ contains more fake than real training
clips, we assign equal total weight to the two
classes. With this weighting, fakes account for
67\% [59, 78] of the clips with the lowest 2\%
of $\mdyn$ values, 76\% [67, 85] between the
2nd and 5th percentiles, and 21\% [17, 25]
in the highest 2\%.

The bins are defined by pooled clip quantiles and held fixed during resampling.

\paragraph{Authentic low mouth dynamics are rare in FF++.}
The median non-speaking window in pooled RealTalk ($\mdyn=-4.14$) falls at the 0.47th percentile [0.09, 0.93] of the FF++ real training distribution, compared with the 23.8th percentile for the median speaking window ($-2.02$).
The non-speaking medians of Seamless test and AMI fall at the 0.79th [0.28, 1.54] and 0.65th percentiles, respectively.
Only 10 real and 70 fake FF++ training clips have $\mdyn$ at or below the RealTalk non-speaking median.
\Cref{fig:supp-ffpp} shows the full distributions.

Confidence intervals for percentiles resample both FF++ and conversational components.
These percentiles characterize coverage of the measured descriptor rather than physical stillness, whose interpretation is limited by the landmark floor (\cref{supp:head}).

\subsection{Agreement across Descriptors and Temporal Transfer}

\paragraph{Alternative descriptors.}
With equal class weights, logistic regressions of the fake label on each rank-normalized descriptor yield negative slopes for DF, F2F, and NT across all four descriptors (\cref{tab:supp-ffpp}).
The alternatives to $\mdyn$ are syllabic-band power $\plip$ (\cref{supp:plip}), SMIRK's 3D jaw angle~\cite{retsinas2024smirk}, and LibreFace's lip-separation action unit AU25~\cite{chang2024libreface}; the latter two do not rely on inner-lip area.
For FS, $\plip$ and jaw angle show negative associations, whereas $\mdyn$ and AU25 do not provide comparable evidence.
All descriptors use the same face crops. However, stabilizing the crops over 1\,s for 80 real clips leaves the jaw-angle--$\plip$ rank correlation nearly unchanged (0.889 versus 0.884).

\begin{table*}[t]
\centering
\footnotesize
\setlength{\tabcolsep}{5pt}
\renewcommand{\arraystretch}{1.12}
\caption{\textbf{Association between mouth dynamics and manipulation labels in FaceForensics++.}
Logistic-regression coefficients for the fake label as a function of each
rank-normalized mouth-motion descriptor, with equal class weighting.
Negative coefficients indicate that higher mouth dynamics are associated
with lower odds of a fake label.
Results are reported across all manipulation methods (All) and separately
for DeepFakes, Face2Face, NeuralTextures, and FaceSwap.
The analysis includes 10,709 training clips (10,594 for AU25).
Brackets indicate 95\% confidence intervals from resampling the 360 identity components.}
\label{tab:supp-ffpp}
\begin{tabular}{@{}lccccc@{}}
\toprule
Descriptor
& All
& DeepFakes
& Face2Face
& NeuralTextures
& FaceSwap \\
\midrule
$\mdyn$
& $-0.54$ [$-0.57$, $-0.51$]
& $-0.90$ [$-0.96$, $-0.84$]
& $-0.65$ [$-0.70$, $-0.60$]
& $-0.68$ [$-0.73$, $-0.64$]
& $-0.03$ [$-0.07$, $0.00$] \\
$\plip$
& $-0.71$ [$-0.74$, $-0.68$]
& $-0.98$ [$-1.05$, $-0.92$]
& $-0.89$ [$-0.94$, $-0.84$]
& $-0.71$ [$-0.75$, $-0.66$]
& $-0.35$ [$-0.39$, $-0.32$] \\
3D jaw angle~\cite{retsinas2024smirk}
& $-0.65$ [$-0.68$, $-0.62$]
& $-0.87$ [$-0.93$, $-0.82$]
& $-0.95$ [$-1.00$, $-0.89$]
& $-0.68$ [$-0.72$, $-0.64$]
& $-0.21$ [$-0.24$, $-0.18$] \\
AU25~\cite{chang2024libreface}
& $-0.41$ [$-0.44$, $-0.38$]
& $-0.63$ [$-0.69$, $-0.58$]
& $-0.59$ [$-0.64$, $-0.54$]
& $-0.48$ [$-0.53$, $-0.44$]
& $+0.01$ [$-0.02$, $0.05$] \\
\bottomrule
\end{tabular}
\end{table*}

Within the lowest 30\% of clips by $\plip$, and after stratification by $\plip$, the estimated probability that a DF, F2F, or NT fake has greater landmark noise than a real clip is 0.55 [0.51, 0.58] (DF: 0.59; FS: 0.60).
This result does not support landmark smoothing as an explanation for the observed attenuation.
Additional jitter may instead offset reduced motion in $\mdyn$; therefore, the small FS difference does not establish that mouth motion is preserved.

\begingroup
\clubpenalties 3 10000 10000 0
\paragraph{Temporal transfer from source and target.}
To examine which reference video supplies the transferred motion, we measure lip aperture over the first 300 frames of each video and aggregate cross-spectra across approximately 710 fake--real pairs per method.
The aperture trajectory is defined as the inner-lip midpoint distance normalized by the outer eye-corner distance (MediaPipe in image mode), rather than the area-based measure used in $\mdyn$.
F2F and NT track their expression source, with coherences of 0.58 and 0.64 and gains of 0.53 [0.51, 0.54] and 0.59 [0.57, 0.61], respectively, at 0.016--0.06 cycles per frame; their coherence with the target is below 0.002.
DF tracks the target, with coherence 0.55 and gain 0.49 [0.47, 0.52], while FS exhibits target-relative gains of 0.65--0.78.
Coherence between the two real videos in each pair is below 0.001.
DF also compresses the aperture range: the 10th percentile increases by 0.017 [0.015, 0.019], while the 90th percentile decreases by 0.032 [0.030, 0.034] normalized units.
Because reference landmark noise can attenuate estimated gains, their absolute values should not be interpreted as direct measures of physical motion preservation.
\par
\endgroup

\subsection{Pose, Frame Rate, Encoding and Window Length}
\label{supp:ffpp-robust}

\paragraph{Head rotation.}
We quantify head rotation as the larger of the yaw and pitch standard deviations.
The median is 3.1\textdegree{} for FF++ real clips; 1.3\textdegree{}, 2.1\textdegree{}, and 2.9\textdegree{} for non-speaking RealTalk, Seamless test, and AMI windows, respectively; and 3.2\textdegree{} for speaking RealTalk windows.
Within FF++ real clips, its correlation with $\mdyn$ is weak (Spearman 0.10 [0.04, 0.16]).
Restricting the reference distribution to real clips with below-median rotation places the three non-speaking medians at the 0.47th [0.00, 1.06], 0.66th [0.09, 1.47], and 0.66th [0.09, 1.39] percentiles.
In the same pose-available data without this restriction, they fall at the 0.47th, 0.75th, and 0.65th percentiles.
Applying the same threshold, derived from real clips, to both classes also preserves the real--fake difference.
The unpaired probability that a real clip has higher $\mdyn$ is 0.647 [0.633, 0.663] after restriction, compared with 0.652 [0.644, 0.660] across all 10,646 clips with pose estimates (2,115 real; 8,531 fake).

\paragraph{Matched frame rate.}
Matching the FF++ real reference distribution to each corpus's frame rate places the non-speaking medians at the 0.86th percentile [0.19, 1.73] for RealTalk, the 0.58th [0.00, 1.38] for Seamless test, and the 0.96th [0.19, 2.00] for AMI.
The corresponding speaking medians fall at the 27.7th, 38.1st, and 43.2nd percentiles.
The Seamless sensitivity analysis uses 491 unflagged non-speaking windows; the main comparison additionally includes recoverable within-shot measurements, yielding 514 windows (\cref{supp:measure}).

\paragraph{Encoding, face size and duration.}
Re-encoding at CRF~23, shortening to 3\,s, and matching FF++ face size change median non-speaking $\mdyn$ by $-0.04$ [$-0.15$, $0.03$] for RealTalk (60 windows) and $-0.02$ [$-0.07$, $0.08$] for Seamless test (40 windows), relative to an FF++ real-clip standard deviation of 0.49.
This paired check retains flagged windows and measures each window over its full retained duration.
Across CRF~23 and 35, half resolution, and 2--4\,s durations for RealTalk and Seamless, as well as duration variants for AMI, the non-speaking medians remain below the 1.3rd percentile of FF++ real clips, including with frame-rate-matched reference distributions; the largest upper confidence limit is 2.3\%.
Speaking medians remain between the 21st and 53rd percentiles.

\section{Retraining DFD-FCG}
\label{supp:retraining}

Changing the composition of FF++ training fakes
alters DFD-FCG's non-speaking penalty
(\cref{sec:ffpp-training}). To understand this
dependence, we assess how the penalty varies with
the strength of the mouth-motion cue, while accounting
for variability across training seeds and possible
trade-offs in manipulation detection.

\subsection{Training Design and Evaluation}

We retrain DFD-FCG on FF++ c23 using the official
architecture and augmentations, with a frozen CLIP
backbone.
Each fake clip is paired with its corresponding
real-target clip at the same temporal offset.
The configured real pool contains 717 videos;
the selected fake samples determine which targets
enter valid training pairs and how frequently
they are sampled.

We evaluate two sets of training conditions.
First, we compare all four manipulation methods
with models trained on each method individually.
All models use 2{,}160 optimizer updates per seed,
corresponding to 30 epochs for all methods and
120 epochs for each single-method condition.
Second, we compare faithful and attenuated
mouth-dynamics arms with matched fake counts,
manipulation-method proportions, and optimizer
settings.
Both arms use 30 epochs (750 updates), with
selected video lists fixed across seeds
(\cref{tab:supp-retrain-conditions}).

Training uses AdamW with a learning rate of
$10^{-4}$, weight decay of $10^{-3}$, and
gradient accumulation over four micro-batches
of 30 pairs each.
We report results for seeds 1019, 2, and 3,
using the final checkpoint for each run.

Conversational evaluation uses the final fused
score from the first detector clip of each window.
We evaluate 647 RealTalk-development and
97 Seamless-test participants with valid scores
in both conversational states.
Participant- and seed-level uncertainty is
described in \cref{supp:retrain-uncertainty}.
Separately, we assess reproduction of the
published training recipe using the checkpoint
with the highest FF++ validation AUC and
the original video-level benchmark protocol.

\begin{table}[t]
\centering\footnotesize
\setlength{\tabcolsep}{3pt}
\renewcommand{\arraystretch}{1.05}
\caption{\textbf{Training conditions and mouth-dynamics bias.}
Training composition, optimizer updates per seed,
and mean $A_{\mathrm{within}}$ (\%) across three seeds
(1019, 2, 3).
Fake counts refer to selected video-list entries;
real targets count distinct videos represented
among valid training pairs.
RT and ST denote RealTalk development (647 participants)
and Seamless test (97 participants).
Conversational results use the final checkpoint
and first-clip detector scores.}
\label{tab:supp-retrain-conditions}
\begin{tabular}{@{}lccccc@{}}
\toprule
Condition & Fake list & Real targets & Steps/seed & RT & ST \\
\midrule
All methods & 2{,}868 & 717 & 2{,}160 & 75.9 & 70.7 \\
DF only & 716 & 716 & 2{,}160 & 78.3 & 80.5 \\
F2F only & 717 & 717 & 2{,}160 & 78.4 & 79.2 \\
NT only & 717 & 716 & 2{,}160 & 76.6 & 74.7 \\
FS only & 718 & 717 & 2{,}160 & 51.6 & 49.8 \\
\addlinespace[2pt]
Faithful arm & 1{,}000 & 566 & 750 & 70.2 & 66.7 \\
Attenuated arm & 1{,}000 & 576 & 750 & 79.5 & 81.6 \\
\bottomrule
\end{tabular}
\end{table}

\subsection{Recipe Reproduction and Seed Variability}

We assess reproduction of the original training
recipe using the published video-level evaluation
protocol.
The first three retrainings achieve Celeb-DF v2
AUCs of 93.9, 92.9, and 94.8, respectively,
compared with 95.0 for the released checkpoint
evaluated through the same pipeline
(\cref{tab:supp-recipe}).
These runs belong to a set of five all-method
retrainings (seeds 1019, 2, 3, 4, and 5),
each trained for 30 epochs (2{,}160 optimizer
updates).
The recipe-reproduction results use the checkpoint
with the highest FF++ validation AUC.

To quantify seed variability, we compare the
last checkpoints of seeds 2--5 against seed 1019
on individual FF++ test clips.
Across the four comparisons, the absolute AUC
difference never exceeds 0.115 percentage points.
This provides the reference for the seed-variability
analysis in \cref{supp:heads}.

\subsection{Training on Individual Manipulation Methods}

Training on FS alone produces a substantially smaller
non-speaking penalty than training on DF alone.
The difference in $A_{\mathrm{within}}$ is
26.7 [21.6, 31.8] percentage points on RealTalk
development and 30.6 [20.5, 40.8] on Seamless test,
with the same direction across all three seeds.
F2F-only and NT-only training also yield larger
penalties than FS-only training.

However, the smaller penalty under FS-only training
comes with weaker generalization across manipulation
methods.
On FF++ test (140 real and 560 fake videos, using
the first clip per video), FS-only achieves
99.8--99.9\% AUC on FS fakes, while DF-only
achieves 99.1--99.3\% on DF fakes.
Across all four manipulation methods, mean AUC
falls to 83.0\% for FS-only, compared with
93.3\% for DF-only, with particularly poor
performance on NT fakes (52.0--56.6\% AUC).

\subsection{Varying Mouth Dynamics within Methods}

\paragraph{Selection at fixed method proportions.}
To vary mouth dynamics while holding
manipulation-method proportions fixed,
we construct two training sets.
Within each method, we regress the mean $\mdyn$
of each fake video, averaged over its training clips,
on the corresponding video-level means of its
real target and source.
The residual captures variation in fake mouth dynamics
not accounted for by those of the source and target.

We select the 250 videos with the highest residuals
per method for the \emph{faithful} arm and the
250 with the lowest for the \emph{attenuated} arm.
Each arm therefore contains 1{,}000 fake videos,
approximately 35\% of the available fakes,
with identical method proportions and
2{,}993 and 3{,}000 training clips, respectively.

The resulting real-target sets contain 566 and
576 distinct videos, of which 438 are shared.
To characterize the mouth-dynamics cue available
during training, we compare video-mean $\mdyn$
between fake videos and their real targets,
counting each target once per associated fake
to reflect the training-pair sampling rule.
Under this weighting, mean real-target $\mdyn$
is similar across arms ($-1.78$ and $-1.77$).

Despite similar real-target mouth dynamics,
the two training arms differ substantially in
the motion of their fake videos. Comparing each
fake with the real targets in the same arm,
including matched and unmatched combinations,
the probability that a real target has higher
video-mean $\mdyn$ than a fake is 0.47 in the
faithful arm and 0.82 in the attenuated arm.
The corresponding probability is 0.66 across
all 2,854 eligible fakes with complete source,
target, and fake measurements.

These are the probabilities reported in \cref{sec:ffpp-training}; they are based on video-level means rather than individual clips.
At clip level, the corresponding value across all real and fake training clips is 0.65 (\cref{supp:ffpp-robust}).
Thus, residual-based selection substantially changes the strength of the mouth-dynamics cue associated with the real/fake label.

\begin{table}[t]
\centering
\footnotesize
\setlength{\tabcolsep}{5pt}
\renewcommand{\arraystretch}{1.10}
\caption{\textbf{Reproducing the DFD-FCG training recipe.}
Video-level AUC (\%).
Retrained: mean $\pm$ standard deviation across three seeds (1019, 2, 3), using the best FF++ validation-AUC checkpoint.
Released: authors' checkpoint evaluated using our protocol; Published: results from the original paper. --- indicates an unavailable published result.}
\label{tab:supp-recipe}
\begin{tabular}{@{}lccc@{}}
\toprule
Dataset & Retrained & Released & Published \\
\midrule
FF++   & $99.45 \pm 0.04$ & 99.4 & --- \\
Celeb-DF v2       & $93.9 \pm 1.0$   & 95.0 & 95.0 \\
FaceShifter       & $97.7 \pm 0.4$   & 98.0 & 98.1 \\
DeeperForensics   & $99.4 \pm 0.1$   & 99.5 & 99.6 \\
WildDeepfake      & $83.5 \pm 0.7$   & 85.2 & 87.2 \\
\bottomrule
\end{tabular}
\end{table}

\paragraph{Conversational effect.}
Training on attenuated-mouth-dynamics fakes increases $A_{\mathrm{within}}$ relative to the faithful arm in all three corresponding seeds on both datasets.
The mean increase is 9.3 [3.6, 15.0] percentage points on RealTalk development (79.5 versus 70.2) and 14.9 [9.8, 19.9] on Seamless test (81.6 versus 66.7).

\paragraph{Manipulation detection and interpretation.}
The two training arms achieve similar mean AUC on FF++ test (98.8 for faithful versus 98.5 for attenuated) and on the Celeb-DF v2 subset of Celeb-DF++ (88.6 versus 88.9; 335 fakes and 174 reals).
On the full scored Celeb-DF++ test set (5{,}187 fakes and 174 reals), however, the faithful arm achieves higher AUC (92.0 versus 89.2).
These last-checkpoint AUCs are computed from up to three clips per video, weighting each clip by $1/n_{\mathrm{clip}}$ so that videos contribute equally.

The contrast shows that selecting training fakes according to their mouth dynamics changes the non-speaking penalty while holding fake counts, manipulation-method proportions, and training budgets fixed.
However, the selected sets also differ in rendering properties, real-target coverage, and sampling frequencies. The observed change therefore cannot be attributed exclusively to mouth dynamics.
The full-data reference provides context at a larger training size and optimization budget.

\subsection{Uncertainty across Participants and Seeds}
\label{supp:retrain-uncertainty}

We report mean $A_{\mathrm{within}}$ across three training seeds, accounting for both participant-sampling uncertainty and between-seed variability.
For each training condition, we estimate
\[
\mathrm{SE}^2=V_{\mathrm{boot}}+\frac{s^2}{3},
\]
where $V_{\mathrm{boot}}$ is obtained from 2{,}000 bootstrap resamples of person--session components and $s^2$ is the sample variance across seeds.
We construct 95\% confidence intervals using a $t$ distribution with two degrees of freedom.

For a contrast between training conditions $A$ and $B$, we combine the paired bootstrap variance with the seed-specific contributions: \begin{equation}
\mathrm{SE}_{\Delta}^{2}
=V_{\mathrm{boot},\Delta}+a+b,
\qquad
a=\frac{s_A^2}{3},\quad
b=\frac{s_B^2}{3}.
\end{equation}
The bootstrap is paired across training conditions,
while between-seed variances are estimated separately.
We use Welch-type degrees of freedom,
\[
\nu=
\frac{(V_{\mathrm{boot},\Delta}+a+b)^2}
{a^2/2+b^2/2}.
\]
Holm correction is applied to the DF--FS
hypothesis tests across the two datasets;
reported confidence intervals remain pointwise.

\section{Heads on Frozen Features}
\label{supp:heads}

\Cref{sec:ffpp-selection} keeps the learned
representations fixed and asks whether FF++ can
distinguish classification heads with different
penalties. Here we detail the
label-informed reference $s_1$, the FF++ feature
geometry, and the 285 FF++-trained candidate heads.

\subsection{Detectors and the Label-Informed Reference}
\label{supp:heads-detectors}

\paragraph{Detector selection.}
We consider eight detectors trained using FF++ data and providing deterministic features suitable for the head comparison (\cref{tab:supp-heads}).
Their training sets overlap but are not identical:
SBI and FakeSTormer use self-blended fakes from FF++ real videos, and GenD-PE represents the shared GenD training recipe.
VLAForge's two-probability ensemble and FTCN's non-deterministic original features are not included.
TFCU lacks a statistically supported non-speaking penalty on Seamless; an additional RealTalk-development check yields $A_{\mathrm{within}}=63.2\%$ for $s_0$, 49.0\% for $s_1$, and 66.6\% after validation-AUC selection.

For DFD-FCG, we use our seed-1019 retraining at the last checkpoint (\cref{supp:retraining}),
with $A_{\mathrm{within}}=76.0\%$.
Unlike the released checkpoint, this retraining has four same-recipe siblings (seeds 2--5), which show how much two trainings of the same recipe differ on FF++; this is the reference against which the changes from $s_0$ to $s_1$ are compared.
Both panels of \cref{tab:supp-heads} use this checkpoint, so in \cref{sec:ffpp-selection} the
DFD-FCG score $s_0$ comes from this retraining rather than from the released model.
Its original score is
\[
s_0=
\log\frac{\bar p_{\mathrm{fake}}+10^{-4}}
         {\bar p_{\mathrm{real}}+10^{-4}},
\]
where $\bar p$ averages the spatial, temporal, and joint heads' probabilities.
The frozen features concatenate layer-normalized spatial and temporal representations
($p=1{,}280$).
The score correction in \cref{supp:mdyn-edit} instead uses the released checkpoint.

Across four additional all-method retrainings (seeds 2--5), the original penalty ranges from
73.1\% to 82.1\%, compared with 48.9--53.0\% for $s_1$ and 76.5--82.6\% after validation-AUC selection.

\paragraph{Construction of $s_1$.}
\label{supp:s1}
Let $x\in\mathbb{R}^p$ denote the frozen features
and $s_0$ the original detector score.
We apply an additive correction directly to $s_0$,
including for DFD-FCG and SBI.
Let $\mu_{\mathrm{FF}}$ and $\Sigma_{\mathrm{FF}}$
be the pooled mean and covariance of FF++ training
features, drawn from 717 real and 2,868 fake videos
before feature-validity exclusions.
For DFD-FCG, these statistics use 10,745 clips,
weighted equally.
Using 323 RealTalk calibration participants,
we define $d$ and $g$ as the participant-averaged
non-speaking minus speaking differences in features
and scores, respectively, with equal participant weights.
We construct
\begin{equation*}
\begin{split}
s_1(x) &= s_0(x)-w^\top(x-\mu_{\mathrm{FF}}),\\
w &= \frac{g(\Sigma_{\mathrm{FF}}+\lambda I)^{-1}d}
{d^\top(\Sigma_{\mathrm{FF}}+\lambda I)^{-1}d}.
\end{split}
\end{equation*}
Since $w^\top d=g$, the correction cancels the mean
within-participant score gap on the calibration
participants by construction.
Thus, $s_1$ is a label-informed reference:
its correction direction and magnitude depend
on conversational-state labels.
It is distinct from the cross-fitted
$s_{1,\mathrm{cf}}$ and descriptor-based $s_w$
in \cref{supp:mdyn-edit}.

We set $\lambda=r\,\mathrm{tr}(\Sigma_{\mathrm{FF}})/p$,
with $r\in\{10^{-6},10^{-5},\ldots,1\}$.
For each candidate, we estimate the covariance on four fifths of FF++ training identities, in a single fixed identity split, and select $r$ by minimizing the RMS score change on the remaining fifth.
We then recompute the covariance and correction using all training identities.
The resulting score is evaluated on 324 disjoint RealTalk development participants, with 1,000 bootstrap resamples of person--session components.

We classify a head as \emph{in-band} if its $A_{\mathrm{within}}$ point estimate lies in $[45,55]$ and its confidence interval is contained in $[40,60]$, and as \emph{out-of-band} if its interval lies entirely above 55 or below 45.
All other cases are indeterminate.
These criteria are diagnostic, not a formal equivalence test.

FF++ AUC uses equally weighted evaluation records:
2,085 test clips for DFD-FCG and 700 test-video
records for each other detector.
Before--after comparisons use identical records.

\begin{table*}[t]
\centering
\footnotesize
\setlength{\tabcolsep}{6pt}
\renewcommand{\arraystretch}{1.12}
\caption{\textbf{Label-informed score correction and FF++ head selection.}
(a) $A_{\mathrm{within}}$ (\%) on 324 held-out RealTalk
development participants for the original score ($s_0$),
the label-informed correction ($s_1$), and its restriction to $U_k$ (\cref{supp:subspace}). $\Delta$AUC is the change in FF++ test AUC
(percentage points) relative to $s_0$.
(b) $U_k$ retains 99\% of FF++ feature variance;
$k/p$ gives the retained PCA dimensions over the
original feature dimensionality.
Linear gap is the percentage of the fitted linear-score
gap in $U_k$; correction outside is the percentage
of the correction direction's squared norm outside $U_k$.
Validation and Test report $A_{\mathrm{within}}$ for
the highest-AUC head among 285 FF++-trained candidates,
selected by validation and test AUC, respectively.
Head counts classify these candidates as in-band,
out-of-band, or indeterminate (\cref{supp:s1}).
Brackets denote 95\% confidence intervals.}
\label{tab:supp-heads}

\begin{tabular}{@{}l c cc cc@{}}
\toprule
\multicolumn{6}{@{}l}{\textit{(a) Label-informed score correction}} \\
\addlinespace[3pt]
 & & \multicolumn{2}{c}{Full correction $s_1$}
   & \multicolumn{2}{c}{Correction in $U_k$} \\
\cmidrule(lr){3-4}
\cmidrule(l){5-6}
Detector
& $A_{\mathrm{within}}(s_0)$
& $A_{\mathrm{within}}$
& $\Delta$AUC
& $A_{\mathrm{within}}$
& $\Delta$AUC \\
\midrule
DFD-FCG
& 76.0 & 49.8 [47.1, 52.6] & $-0.03$
& 49.2 [46.2, 52.1] & $-0.02$ \\
RealForensics
& 86.2 & 47.1 [43.3, 51.2] & $-0.01$
& 47.2 [43.2, 51.2] & $-0.03$ \\
LipForensics
& 68.9 & 53.1 [49.2, 57.2] & $-0.08$
& 52.5 [48.9, 56.1] & $-0.05$ \\
TALL
& 79.0 & 51.2 [47.7, 54.9] & $-0.12$
& 46.8 [43.5, 50.4] & $-0.33$ \\
FakeSTormer
& 62.7 & 57.3 [54.6, 59.8] & $-0.48$
& 51.2 [48.8, 53.7] & $-5.83$ \\
DFD-HR
& 69.5 & 51.7 [48.4, 54.8] & $-0.09$
& 51.5 [48.4, 54.8] & $-0.12$ \\
GenD-PE
& 64.5 & 46.5 [42.8, 50.0] & $-0.04$
& 45.3 [41.5, 48.8] & $-0.04$ \\
SBI
& 67.0 & 50.9 [47.1, 54.4] & $-0.03$
& 50.0 [46.2, 53.5] & $-0.10$ \\
\bottomrule
\end{tabular}

\par\vspace{8pt}

\begin{tabular}{@{}l c cc cc c@{}}
\toprule
\multicolumn{7}{@{}l}{\textit{(b) Feature geometry and FF++ head selection}} \\
\addlinespace[3pt]
 & & \multicolumn{2}{c}{Feature geometry}
   & \multicolumn{2}{c}{AUC-selected head}
   & \\
\cmidrule(lr){3-4}
\cmidrule(lr){5-6}
Detector
& $k/p$
& \shortstack{Linear gap\\in $U_k$ (\%)}
& \shortstack{Correction\\outside $U_k$ (\%)}
& Validation
& Test
& \shortstack{Head counts\\In / Out / Indet.}\\
\midrule
DFD-FCG
& 861/1280 & 98.7 & 69.9
& 79.0 [76.4, 81.4] & 76.7 [74.3, 79.3]
& 0 / 285 / 0 \\
RealForensics
& 26/2048 & 101.1 & 99.4
& 90.8 [88.7, 92.7] & 79.5 [76.7, 82.2]
& 23 / 235 / 27 \\
LipForensics
& 8/768 & 97.2 & 100.0
& 63.1 [60.0, 65.9] & 73.6 [70.5, 76.6]
& 44 / 224 / 17 \\
TALL
& 261/1024 & 99.8 & 99.5
& 75.0 [71.7, 78.3] & 78.1 [75.0, 81.1]
& 0 / 285 / 0 \\
FakeSTormer
& 42/768 & 99.8 & 99.6
& 63.2 [60.3, 65.9] & 62.2 [59.2, 64.9]
& 0 / 268 / 17 \\
DFD-HR
& 442/1536 & 98.6 & 100.0
& 65.9 [63.0, 68.9] & 70.1 [67.4, 72.6]
& 0 / 285 / 0 \\
GenD-PE
& 211/1024 & 100.4 & 99.6
& 58.7 [55.7, 61.4] & 63.4 [60.3, 66.1]
& 5 / 277 / 3 \\
SBI
& 954/1792 & 106.8 & 64.6
& 72.5 [69.6, 75.3] & 70.6 [67.6, 73.4]
& 0 / 275 / 10 \\
\bottomrule
\end{tabular}
\end{table*}

\subsection{Benchmark Performance and Conversational Penalties}

The label-informed reference $s_1$ brings seven
of eight detectors into the diagnostic band on
held-out RealTalk development participants
(\cref{tab:supp-heads}a), while leaving FF++ benchmark
performance nearly unchanged.
Relative to $s_0$, FF++ test AUC changes by at most
0.12 percentage points for seven detectors and
0.48 points for FakeSTormer.
For comparison, the same-recipe DFD-FCG retrainings
in \cref{supp:retraining} differ by up to 0.115
AUC points.

The similarity extends beyond AUC.
On FF++ training examples, the change in mean
log-loss lies between $-0.003$ and $+0.003$ nats
for seven detectors.
Using the original score's 95th-percentile
threshold on real training examples for both
$s_0$ and $s_1$, only 0.03--0.92\% of decisions
change.
FakeSTormer is the exception, with a log-loss
increase of 0.060 nats and 3.2\% changed decisions.

For seven detectors, substantial differences in
conversational penalties coexist with minimal changes
in FF++ AUC, log-loss, and binary decisions,
consistent with underspecification
\cite{damour2022underspecification}.

\subsection{The FF++ Feature Subspace}
\label{supp:subspace}

Let $U_k$ contain the smallest set of principal
directions explaining 99\% of FF++ feature variance.
To examine the geometry of the original score gap,
we fit a linear surrogate $w_0$ to each detector's
scores on the combined FF++ training and RealTalk
calibration features.
The surrogate achieves $R^2=0.998$ for DFD-FCG
and $0.987$ for SBI on these data.
We measure the contribution of $U_k$ to the
reconstructed score gap as
\begin{equation*}
\frac{w_0^\top U_kU_k^\top d}{w_0^\top d}.
\end{equation*}

Across detectors, the principal FF++ feature
directions account for 97--107\% of the
reconstructed score gap. Values above 100\%
arise when directions outside $U_k$ contribute
with the opposite sign.

However, 65--100\% of the correction direction's
squared norm lies outside $U_k$
(\cref{tab:supp-heads}). These percentages
describe different quantities: the contribution
to the original score gap and the magnitude
of the correction weights.

Restricting the correction to $U_k$ still brings
$A_{\mathrm{within}}$ point estimates into
$[45,55]$ for all eight detectors.
The FF++ test AUC decreases by at most 0.12
percentage points for six detectors, by 0.33
for TALL, and by 5.83 for FakeSTormer
(\cref{tab:supp-heads}a).
At a 99.9\% variance threshold, FakeSTormer
remains indeterminate under the diagnostic
band criterion.

Finally, we compare $s_1$ with 1,000 random
correction directions.
We divide the eigenvalues of $\Sigma_{\mathrm{FF}}$ into ten bands of equal width on a logarithmic scale. In the eigenbasis, each random direction has the same norm as the $s_1$ direction within each band, with a random Gaussian orientation inside the band; this matches the spectral profile.
Its amplitude gives the same RMS score change as $s_1$ over the FF++ training examples, in units of the standard deviation of $s_0$ on real training examples, and its sign is chosen to reduce the calibration gap. Unlike the random directions of \cref{supp:mdyn-edit}, it is not rescaled to cancel that gap.
None achieves $A_{\mathrm{within}}\leq55\%$
for any detector.
This indicates that the reduction in the non-speaking penalty is not reproduced by arbitrary corrections of comparable magnitude and spectral structure.

\subsection{Training and Selecting 285 Heads per Detector}
\label{supp:ffpp-heads}

For each detector, we train 285 candidate heads
using 80\% of FF++ training identities for fitting
and the remaining 20\% for validation.
The pool comprises 78 logistic regressions,
39 squared-hinge classifiers, 39 weighted
least-squares classifiers, 13 label-smoothed
logistic regressions, 20 one-hidden-layer MLPs,
95 SGD/Adam runs, and one score ensemble.

The logistic, squared-hinge, and least-squares models use three feature normalizations: global scaling, per-dimension standardization, and PCA whitening retaining 99.9\% of variance, each estimated on the fitting identities.
We consider 13 regularization strengths, $\lambda=\kappa\lambda_{\max}$, with $\kappa\in\{10^{-9},10^{-8},\ldots,10^3\}$, where $\lambda_{\max}$ is the largest eigenvalue of the fitting covariance.
Logistic regression additionally varies natural versus balanced class weights.
The label-smoothed logistic models use global scaling, smoothing factor 0.1, and the same regularization grid.
The MLPs combine 64 or 256 hidden units, weight decay $10^{-4}$ or $10^{-2}$, and five random seeds.

The SGD/Adam block consists of 50 standard SGD runs, 20 Adam runs, 20 SGD runs with tenfold initialization scale, and five zero-initialization SGD runs, together with their mean-score ensemble.
These models use global scaling computed from all FF++ training features, fitting and validation identities together, and batches of 256.
Training lasts at most 300 epochs, with early stopping after ten epochs without validation-loss improvement.

\paragraph{Selection criteria.}
We rank the 285 heads by FF++ validation AUC
and, separately, by test AUC, breaking ties
using validation log-loss.
Validation AUC reaches 99.9--100\% for six
detectors, 99.4\% for SBI, and 93.3\%
for FakeSTormer.
Despite these high benchmark scores, both
selection criteria choose heads with
$A_{\mathrm{within}}>55\%$ for all eight
detectors (\cref{tab:supp-heads}b).
Tie-breaking log-loss is computed after Platt
scaling across two identity-disjoint folds.

\paragraph{Candidate generation versus selection.}
The candidate pool contains in-band heads for
RealForensics, LipForensics, and GenD-PE, yet
neither FF++ AUC ranking selects them.
For DFD-FCG, TALL, and DFD-HR, all 285 candidates
are out of band; FakeSTormer and SBI have no
in-band candidates but include indeterminate ones
(\cref{tab:supp-heads}b).
Thus, FF++ AUC selection fails to identify
available lower-penalty heads for three detectors,
whereas for the remaining five, the candidate
pool itself contains no heads meeting the
diagnostic in-band criterion.

\paragraph{Adding a mouth descriptor on FF++ alone.}
A separate rule regresses features on video-averaged $\mdyn$ using 717 FF++ real training videos (716 for TALL).
It uses the pooled FF++ covariance, selects ridge strength by the correlation between out-of-fold projections and $\mdyn$, and cross-fits amplitude over five identity folds.
Its success criterion requires $A_{\mathrm{within}}\in[45,55]$ with a joint 95\% interval inside $[40,60]$, standardized-gap reduction $\rho\in[0.8,1.25]$ with lower bound at least 0.5, and a ratio of within-participant, within-state score standard deviations in $[1/1.5,1.5]$ relative to $s_0$.
Here $\rho=1-G_1/G_0$, where $G_h$ is the mean participant score gap divided by the standard deviation of score $h$ on real FF++ test videos.
The $A_{\mathrm{within}}$ interval jointly resamples FF++ identities and RealTalk components, refitting direction and amplitude with the covariance and ridge fixed.
The interval for $\rho$ resamples RealTalk components only, holding the fitted correction and FF++ standard deviations fixed.
An interval wholly above 55 or below 45 is classified as a miss; other cases are indeterminate.

The rule succeeds in none of the eight detectors, misses in five (DFD-FCG, GenD-PE, SBI, TALL and DFD-HR; 8--25\% of the standardized gap removed), and is indeterminate in three: RealForensics, 53.5 [33.4, 78.1]; LipForensics, 49.2 [35.2, 65.0]; and FakeSTormer, 63.3 [54.7, 65.3].
Most of the wide uncertainty for RealForensics and LipForensics comes from resampling FF++ identities.
All five DFD-FCG retraining seeds are misses.
Replacing $\mdyn$ with $\plip$ (\cref{supp:plip}) gives the same detector-level classifications.

\section{Mouth-Dynamics-Aware Score Correction}
\label{supp:mdyn-edit}

Reducing the non-speaking ranking penalty does not
necessarily produce balanced decisions or preserve
manipulation detection performance.
The score correction in \cref{sec:score-edit}
illustrates this limitation: it substantially
reduces the penalty on unseen RealTalk participants,
but transfers inconsistently across datasets and
weakens detection of certain manipulations.
We detail the correction procedure and examine
these trade-offs below.

\subsection{Estimator and Calibration}

\paragraph{Within-person moments.}
With $x$ and $s_0$ as in \cref{supp:s1},
let $z=\mdyn$. For each calibration participant $j$ with
$n_j\geq2$ windows, we compute
\begin{equation*}
\begin{split}
C_j &=
\frac{1}{n_j}\sum_i
(x_i-\bar x_j)(z_i-\bar z_j),\\
\gamma_j &=
\frac{1}{n_j}\sum_i
(s_{0,i}-\bar s_{0,j})(z_i-\bar z_j).
\end{split}
\end{equation*}
We split calibration participants into two halves,
$A$ and $B$, and average these moments with equal
participant weights within each half.
For $h\in\{A,B\}$, let $\bar h$ denote the
opposite half. We construct
\begin{equation}
\begin{gathered}
v_h=(\Sigma_{\mathrm{FF}}+\lambda_h I)^{-1}C_h,
\qquad
\kappa_h=
\frac{\gamma_{\bar h}}{C_{\bar h}^{\top}v_h},\\
w=\frac{1}{2}\sum_{h\in\{A,B\}}\kappa_h v_h,\\
s_w(x)=s_0(x)-w^\top(x-\mu_{\mathrm{FF}}).
\end{gathered}
\label{eq:mouth-edit-w}
\end{equation}
Here $\mu_{\mathrm{FF}}$ and $\Sigma_{\mathrm{FF}}$
are estimated from all available FF++ real and fake
training features, using the population covariance
normalization.
Each half determines a correction direction,
whose amplitude is calibrated on the opposite half
to cancel its mean within-person score--mouth-dynamics
covariance, provided the denominator is nonzero.
The two corrections are then averaged and need
not cancel the covariance exactly.

\paragraph{Regularization and geometry.}
We set $\lambda_h=r_h\,\mathrm{tr}(\Sigma_{\mathrm{FF}})/p$, selecting $r_h\in\{10^{-6},10^{-5},\ldots,1\}$
separately for each half.
For each candidate, we estimate the covariance on four fifths of FF++ training identities and
evaluate the RMS score change on the remaining fifth, selecting the first minimum.
For each candidate, RealTalk calibration moments provide the covariance direction and the
cross-fitted amplitude constraint.
We then recompute the covariance, ridge scale, direction, and amplitude using all FF++ training
identities.

Up to a multiplicative constant, $v_h$ minimizes $u^\top(\Sigma_{\mathrm{FF}}+\lambda_h I)u$
subject to a fixed $C_h^\top u$.
The resulting direction therefore limits score variation under the FF++ covariance
metric while retaining sensitivity to the RealTalk within-person covariance.
The construction draws on LEACE \cite{belrose2023leace}, but uses RealTalk
within-person cross-covariances, an FF++ feature metric, and cross-fitted correction amplitudes.

\paragraph{Calibration sample.}
Requiring finite mouth-dynamics and head-motion estimates leaves 1,848 of 1,904 authentic windows
from 323 RealTalk participants, equally divided between speaking and non-speaking states.
Detector-specific coverage reduces this sample to 1,817 windows for I3D and 1,838 for TALL.
Participants are assigned to two halves of 162 and 161 by alternating sorted, hashed
participant identifiers, ensuring that all windows from the same participant remain together.

The additive $\log\mathrm{fps}$ term in $\mdyn$ cancels under within-participant centering,
as frame rate is constant for each participant.
Speaking annotations are used for balanced sampling and descriptor comparisons on RealTalk
development and Seamless, including its test split, but do not enter the score-correction estimator.
Real/fake labels are likewise not used to fit the correction.

\paragraph{Application.}
Features are aggregated over the same clips used
to compute detector scores.
For DFD-FCG, we use the released checkpoint,
the fused log-probability margin defined in
\cref{supp:heads-detectors}, and concatenated
layer-normalized spatial and temporal features
($p=1{,}280$).

At inference, the correction requires only
the frozen features $x$, the original score $s_0$,
and the calibrated parameters $w$ and
$\mu_{\mathrm{FF}}$; it requires no mouth-dynamics
measurements, speaking labels, or participant
identifiers.
For a linear readout $s_0(x)=\beta^\top x+b$,
the correction is equivalent to replacing
$\beta$ with $\beta-w$ and $b$ with
$b+\mu_{\mathrm{FF}}^\top w$.
For nonlinear or fused scores, we retain
the original readout and add
$-w^\top(x-\mu_{\mathrm{FF}})$.

The frozen features remain unchanged, so any
fixed mouth-dynamics decoder operating on $x$
produces identical predictions before and
after correction.

\begin{table*}[t]
\centering
\footnotesize
\setlength{\tabcolsep}{5pt}
\renewcommand{\arraystretch}{1.10}
\caption{\textbf{Within-person ranking after score correction.}
$A_{\mathrm{within}}$ (\%) with 95\% person--session bootstrap confidence intervals, evaluated on the populations of \cref{tab:mouth-edit}.
The descriptor-based correction $s_w$ uses the main-table scores, whereas the cross-fitted label-informed reference $s_{1,\mathrm{cf}}$ uses feature-pipeline scores and is evaluated on RealTalk and AMI.
The final row counts detectors whose entire confidence interval lies within $[45,55]$.
For TFCU, the reference baseline is 59.3\% versus 56.8\% for $s_w$; FakeSTormer uses 290 participants for the reference and 289 for $s_w$.
Coverage and descriptive results follow
\cref{supp:ci}.}
\label{tab:supp-mdyn-edit-ci}
\begin{tabular}{@{}lccccc@{}}
\toprule
 & \multicolumn{2}{c}{RealTalk (unseen)}
 & Seamless test & \multicolumn{2}{c}{AMI} \\
\cmidrule(lr){2-3}
\cmidrule(lr){4-4}
\cmidrule(l){5-6}
Detector
& $s_w$
& $s_{1,\mathrm{cf}}$
& $s_w$
& $s_w$
& $s_{1,\mathrm{cf}}$ \\
\midrule
RealForensics
& 51.6 [48.3, 54.8]
& 44.6 [41.4, 47.9]
& 65.7 [60.1, 70.8]
& 47.0 [42.6, 51.4]
& 43.9 [39.7, 48.3] \\
LipForensics
& 53.3 [49.8, 56.7]
& 51.0 [47.4, 54.5]
& 56.1 [50.9, 61.3]
& 61.6 [58.0, 65.3]
& 60.1 [56.4, 63.9] \\
I3D-DF40-FR
& 54.3 [50.0, 58.5]
& 52.9 [48.6, 57.1]
& 78.1 [71.6, 83.7]
& 52.3 [49.4, 55.0]
& 50.4 [47.5, 53.0] \\
I3D-DF40-FS
& 40.5 [36.7, 44.5]
& 39.5 [35.8, 43.4]
& 56.6 [48.0, 64.8]
& 29.5 [26.5, 32.6]
& 28.9 [26.0, 31.9] \\
I3D-DF40-FSFR
& 49.5 [45.4, 53.7]
& 49.5 [45.5, 53.6]
& 66.8 [59.3, 74.2]
& 36.4 [33.1, 39.8]
& 36.4 [33.2, 39.9] \\
TALL
& 51.0 [47.1, 54.9]
& 47.1 [43.1, 50.9]
& 51.0 [44.5, 57.7]
& 60.9 [57.6, 64.2]
& 60.9 [57.3, 64.3] \\
FakeSTormer
& 53.5 [50.3, 56.6]
& 50.6 [47.3, 53.8]
& 65.2 [60.8, 69.5]
& 44.6 [41.2, 48.1]
& 45.3 [42.0, 48.7] \\
TFCU
& 43.9 [40.6, 47.1]
& 45.5 [42.2, 48.7]
& 37.6 [33.0, 42.2]
& 51.9 [48.4, 55.2]
& 52.4 [49.0, 55.7] \\
DFD-FCG
& 49.0 [45.3, 52.4]
& 49.0 [45.3, 52.3]
& 49.4 [44.7, 53.7]
& 43.0 [39.9, 46.1]
& 43.2 [40.0, 46.4] \\
DFD-HR
& 47.0 [43.3, 50.7]
& 45.9 [42.1, 49.6]
& 45.7 [40.7, 50.8]
& 48.1 [45.7, 50.7]
& 47.6 [45.1, 50.2] \\
GenD-CLIP
& 48.4 [44.9, 52.1]
& 47.5 [44.1, 51.2]
& 42.3 [37.1, 47.0]
& 46.8 [43.4, 50.5]
& 47.3 [43.9, 50.9] \\
GenD-PE
& 45.2 [41.2, 49.1]
& 43.8 [39.9, 47.8]
& 39.9 [34.9, 45.2]
& 53.8 [51.1, 56.4]
& 55.7 [53.2, 58.2] \\
GenD-DINOv3
& 48.9 [45.3, 52.6]
& 47.8 [44.1, 51.5]
& 31.7 [26.6, 37.0]
& 37.9 [35.0, 41.7]
& 38.6 [35.8, 42.3] \\
SBI
& 50.3 [46.8, 53.9]
& 46.8 [43.4, 50.2]
& 58.1 [53.5, 62.8]
& 56.9 [53.7, 60.0]
& 56.2 [52.9, 59.4] \\
\midrule
Median
& 49.3 & 47.3 & 53.5 & 47.5 & 47.4 \\
95\% CI in $[45,55]$
& 6/14 & 4/14 & 0/14 & 2/14 & 2/14 \\
\bottomrule
\end{tabular}
\end{table*}

\subsection{Evaluation and Speaking-Label Reference}

\paragraph{Populations and score pairing.}
We evaluate the calibrated corrections without
refitting on 290 unseen RealTalk participants,
whose conversations do not overlap with those
of calibration or development participants,
as well as on Seamless test and AMI.
Detector-specific coverage follows \cref{supp:ci},
and each original--corrected comparison uses
identical windows.

For TFCU on all three datasets and DFD-FCG and
FakeSTormer on Seamless, feature-pipeline scores
differ from those used in the main tables.
We therefore apply the fixed per-window correction
to the main-table scores, using the logit scale
for TFCU.
For SBI on AMI, we use the logit of the mean
predicted probability.
Since evaluation does not require test-window
$\mdyn$, AMI includes 124 participants rather
than the 123 available for the mouth-adjusted
analysis in \cref{supp:gap}.

We evaluate 14 detectors.
FTCN is excluded because deterministic features
are unavailable across the full evaluation,
and the pipeline does not support VLAForge's
two-head ensemble.
These are implementation limitations, not
restrictions of the additive correction,
which also applies to nonlinear scores.

\paragraph{Speaking-label reference.}
The cross-fitted speaking-label reference
$s_{1,\mathrm{cf}}$ follows \cref{eq:mouth-edit-w},
replacing $C_h$ and $\gamma_h$ with the
participant-averaged non-speaking minus speaking
differences in features and scores, respectively.
It also includes calibration windows without valid
$\mdyn$, provided both speaking states are available
for the participant.
Unlike equal weighting of participant-level gaps,
binary-state covariance weights each gap by
$\pi_j(1-\pi_j)$, where $\pi_j$ is the participant's
non-speaking fraction.
The reference $s_1$ in \cref{supp:s1} instead
estimates its direction and amplitude using
all calibration participants without cross-fitting.

\paragraph{Ranking outcomes.}
On unseen RealTalk participants, the median
$A_{\mathrm{within}}$ decreases from 65.7\% to
49.3\% after mouth-dynamics-aware correction
(\cref{tab:supp-mdyn-edit-ci}).
Twelve of 14 detectors have point estimates in
$[45,55]$, although only six have their entire
95\% confidence intervals within this range.
The speaking-label reference yields 11 point
estimates and four intervals within the same band.

Cross-dataset results are less consistent.
On Seamless test and AMI, corrected median $A_{\mathrm{within}}$ values are 53.5\% and 47.5\%, respectively, but only three and six point estimates fall within $[45,55]$.
No detector on Seamless and only two on AMI have confidence intervals entirely within the band.
The speaking-label reference yields five in-band point estimates and two intervals within the band on AMI.
These intervals describe proximity to chance, not statistical equivalence.

\subsection{Specificity, Dependence and Decisions}

\paragraph{Specificity controls.}
In these controls, unlike the head classification of \cref{supp:s1}, a detector is in-band when its $A_{\mathrm{within}}$ point estimate lies in $[45,55]$.
On 324 RealTalk development participants held out from calibration, shuffling $\mdyn$ within participants yields a median of three in-band detectors out of 14 across 20 shuffles, while replacing $\mdyn$ with head motion residualized on $\mdyn$ yields only one.
Random feature directions matched in correction norm leave rankings largely unchanged in the 13-detector comparison excluding TALL.
When scaled to cancel the same covariance, random directions can bring rankings near chance, but require approximately 20 times greater RMS score change on FF++ test and reduce Celeb-DF v2 AUC by about 16 percentage points.

The FF++ covariance metric also matters.
Replacing it with within-person calibration covariance, pooled calibration covariance, or the identity yields 7, 10, and 11 in-band detectors out of 13, respectively, compared with 12 using the FF++ metric.
Under the ridge-selection validation objective, the corresponding RMS score changes are 6.2, 2.5, and 18.5 times larger.

\paragraph{Dependence within each state.}
Reducing the between-state ranking penalty does not
necessarily remove score dependence on mouth dynamics
within each state.
On unseen RealTalk participants, median
participant-fixed-effect slopes change from
$-0.16$ to $+0.15$ in non-speaking windows
and from $-0.58$ to $-0.32$ in speaking windows.
After correction, all 14 non-speaking slopes
are positive, with confidence intervals above
zero for nine detectors, whereas speaking slopes
remain significantly negative for 13.

These slopes are standardized using within-participant
standard deviations of the original score and
$\mdyn$ estimated on calibration participants,
rather than the rank normalization used in
\cref{supp:within}.
For DFD-FCG, the non-speaking slope changes
from $-0.46$ to $+0.06$ [$-0.07$, $0.18$],
while the speaking slope changes from $-0.60$
to $-0.20$.
Because calibration combines within-state variation
with the strong between-state mouth-dynamics contrast,
reducing the overall covariance need not eliminate
within-state associations and can reverse their sign.
Landmark jitter may also affect associations
at low motion levels (\cref{supp:head}).

\paragraph{False-positive gaps.}
We recompute the FF95 threshold for each score
using 140 real FF++ test videos, weighting
DFD-FCG clips equally within each video.
Original and corrected decisions are compared
on matched windows using feature-pipeline scores.
In 2,000 person--session bootstrap replicates,
the estimated FF95 thresholds remain fixed.

On unseen RealTalk participants, the median
non-speaking-minus-speaking false-positive rate
gap decreases from $+10.6$ to $+3.9$ percentage
points, but eight of 14 detectors retain
confidence intervals entirely above zero.
For TALL, the gap decreases from $+26.9$ to
$+5.3$ points [1.3, 9.1].
On AMI, the median gap changes from $+9.8$
to $-1.4$ points across 13 detectors, with
five retaining intervals above zero.
TFCU is excluded from the AMI comparison
because its aligned scores are not on the
calibrated FF95 scale.

Even when speaking and non-speaking windows
have similar within-person rankings, non-speaking
windows may still produce more scores above
the detection threshold. Therefore, reducing
the ranking penalty does not guarantee equal
false-positive rates between the two states.

The reported intervals condition on the
estimated FF95 thresholds.

\subsection{Manipulation Detection}

\paragraph{Benchmarks and mouth animation.}
The correction produces median AUC changes of
$-0.05$ percentage points on FF++ test and
$-0.20$ on the Celeb-DF v2 subset of Celeb-DF++.
However, its impact varies across detectors:
FakeSTormer's Celeb-DF++ AUC decreases from
81.48 to 75.21 ($-6.27$ points).

We also test whether the correction mitigates
the detection loss caused by increased mouth
motion. On 4,580 Celeb-DF++ fakes from 59
identities outside the low-motion subset,
doubling mouth motion reduces DFD-FCG detection
by 26.2 percentage points before correction
and 26.5 points afterward.
The difference between these effects is
$-0.33$ points [$-1.61$, $0.88$], providing
no evidence that the correction mitigates
this vulnerability.
The original-to-rendered comparison includes
rendering differences; the matched
still-to-doubled comparison is reported in
\cref{supp:liveportrait}.

\paragraph{Motion-transfer fakes.}
We use \mbox{TPSMM}~\cite{zhao2022tpsmm} and
\mbox{FOMM}~\cite{siarohin2019fomm} to animate
all 324 RealTalk development participants with
speaking and non-speaking clips from their
conversation partners.
For each participant, both conditions use the
same source frame, chosen for minimum mouth
opening from a non-speaking clip.
Sham controls undergo the same cropping,
pasting, and encoding operations without
generation.
Blind inspection identifies gross visual
failures in three of 20 non-speaking-driven
clips per generator; the results include
these outputs.

Before correction, median TPR across 14
detectors is 75.8\% for non-speaking-driven
versus 81.8\% for speaking-driven TPSMM
fakes, and 85.4\% versus 90.0\% for FOMM.
After correction and FF95 recalibration,
TPR decreases on non-speaking-driven fakes
for all 14 detectors, with median losses
of 8.9 percentage points for TPSMM and
7.7 for FOMM.
Speaking-driven fakes also show median
losses of 6.1 and 4.0 points, respectively.

\end{document}